# Discrete approach to machine learning[1]

**Dmitriy Kashitsyn**[2]
dk@daiylabs.com

**Dmitriy Shabanov**[3]
ds@daiylabs.com

## Abstract

The article explores an encoding and structural information processing approach using sparse bit vectors and fixed-length linear vectors.

The following are presented:

- A discrete method of speculative stochastic dimensionality reduction of multidimensional code and linear spaces with linear asymptotic complexity;
- A geometric method for obtaining discrete embeddings of an organised code space that reflect the internal structure of a given modality.

The structure and properties of a code space are investigated using three modalities as examples: morphology of Russian and English languages, and immunohistochemical markers.

Parallels are drawn between the resulting map of the code space layout and so-called *pinwheels* appearing on the mammalian neocortex. A cautious assumption is made about similarities between neocortex organisation and processes happening in our models.

## Contents

[1] Russian edition of the article as well as supplementary materials are available at https://daiylabs.com/papers/daml/

[2] Дмитрий Каши́цын.

[3] Дмитрий Шаба́нов.

# 1. Introduction

Recently, models based on neural network transformer architecture have shown impressive results.

This makes it more challenging to offer something new. Not only it is necessary to demonstrate the feasibility of the approach but also its effectiveness and advantages over the mainstream.

In this work, we take the first step towards this goal.

Although traditional neural network models are based on a discrete representation of real numbers, conceptually, the models operate in a continuous space. Moreover, modern methods of training of artificial neural networks impose differentiability requirements on all functions involved in the process.

On the contrary, our approach is based on the discrete representation of discrete concepts. This occurs at all levels of the hierarchy: from primary encoding to the representation of knowledge and operations with it.

The model's facts and experience are represented as sparse bit vectors, usually of fixed length.

Notably, only the group of bits is semantically significant. This is reminiscent of population coding in neural networks [1].

By gathering the code representations of many concepts together, it is possible to reduce the dimensionality (do a *layout*) of the two-dimensional projection of this *code space*. This is done using an algorithm similar to UMAP [2] and reminiscent of the process of topological organisation of the mammalian neocortex [3], [4].

The layout not only allows one to visualise the code-space topology, but at the same time, to construct discrete embeddings that reflect the *structural similarity* of code elements and, in turn, the original concepts.

This paper focuses on the structure of the resulting code-space and operations with it. We show that in order to obtain proper layout, it is important to choose the right code system and solve the problem of primary code density.

## 1.1. Related work

The idea of population coding is not new and was considered long before the era of machine learning.

Georgopoulos et al. (1986) demonstrated that a population of neurons in the motor cortex encodes the direction of hand movement in primates.

Bonhoeffer & Grinvald (1991) studied the visual cortex of mammals and showed that the orientation sensitivity map of minicolumns in visual cortex is organised in the form of regular structures resembling pinwheels.

Pouget et al. (2003) developed Bayesian models of population coding that explain how the brain integrates information from different neural ensembles.

Boerlin (2013) demonstrated how population coding can efficiently represent information in spike networks.

Dimension reduction algorithms were investigated in the following works by Kohonen (1982), Maaten & Hinton (2008), McInnes et al. (2020).

Artificial neural networks with per-layer training were studied in the works Rauber et al. (2002), Hinton & Salakhutdinov (2006), Bengio et al. (2006), Hinton et al. (2006), Salakhutdinov & Hinton (2012).

Najafian et al. (2022) formulated a theory of cortical formation based on the density of thalamic afferents and the dimensionality of stimuli. Their method of modelling cortical organisation is based on sorting afferents by local point replacements, which is very similar to our layout algorithm.

Our research is based mainly on the work[4] of A. Redozubov, who described [18] the acquisition and use of sparse bit vectors with Gray code properties for the primary encoding of stimuli, proposed a possible biologically motivated mechanism for the implementation of holographic associative memory, and offered his view on the role of the hippocampus in the process of memory consolidation [19]. Based on the code hypothesis of neocortex organisation, Redozubov showed [20] the possibility of organising a multidimensional space of contexts (orientation, shifts along X and Y, eye dominance).

## 1.2. Our contribution

### 1.2.1. Encoding system

Based on Redozubov's ideas, we developed a primary coding system and identified hyperparameters that work for our layout algorithms.

Much work has been done on the practical study of primary coding methods for different modalities. General principles for effective primary coding have been formulated.

We have developed the "colour"-aware algorithm for code merging and applied it to the initial stimuli coding and the hierarchy of detectors. We formulated the proximity-sensitivity-density problem and demonstrated the effectiveness of "colour" codes for solving it.

### 1.2.2. Layout

We implemented[5] two layout algorithms that, working in tandem, can achieve long- and short-range order. We fine-tuned hyperparameters and demonstrated the effectiveness of clipping (parameter $\lambda$) for successful layout.

Redozubov viewed the layout [20, p. 3.2], [21, p. 2] as a method of trophic and logistical optimisation, as well as for visualisation, similar to other dimension reduction algorithms.

Najafian et al. (2022) described the processes of topology organisation and its dependence on afferent density and stimulus dimensionality, but, to our knowledge, they do not speculate on the reasons for such organisation, limiting themselves to considerations of efficiency [4, pp. 3, 13] and hypothesise that the orderliness of cortical minicolumns arises naturally as a consequence of the topical organisation of afferents.

On the contrary, our experiments have shown that solving the NP-complete layout problem is valuable and critical for obtaining a locally ordered topology. It can then be used to construct a detector space and encode its activity as discrete structural embeddings. This reduces the dimensionality of the codes and their transformation from the stimulus domain to a structural domain specific to a given modality.

In other words, this type of cortex organisation is not simply a natural consequence, but a necessity.

### 1.2.3. Detectors and activation

To localise activation points, Redozubov proposed an energy convolution algorithm. We had implemented it [21, Fig. 5], but subsequently abandoned it in favour of hierarchical detection without convolution.

Redozubov proposed using random projections to construct embedding codes. The technique

[4]Many concepts have not been properly formalised in scientific articles, so it is difficult to trace their sources. A popular description of them can be found in a series of articles [15], [16], and [17].

[5]At that time, the UMAP algorithm [2] have not yet been described. In our early experiments, we independently found similar solution.

works, but it is subject to the same sensitivity-density problem.

We have developed algorithms for constructing detector spaces that consider the code space's topology and provide near-optimal coverage, resulting in compact yet efficient codes.

We proposed using space's static energy to suppress noise and highlight cluster boundaries. The same method filters out outliers when activating the code space.

To test the theory, we solved several practical problems involving structural coding of stimuli of different nature: the morphology of the Russian and English languages, and immunohistochemical markers[6].

## 2. Sparse Bit Vectors

The entire method is based on operations with sparse bit vectors of fixed length:

$$\mathbf{v} = (v_1, v_2, ..., v_n), \quad \text{where} \quad v_i \in \{0, 1\}.$$

The vectors are called *sparse* because only some of the $v_i$ have values other than 0. In practice, bit vectors work well when no more than 25% of bits are set to 1.

In addition to bit vectors, it is often beneficial to use normalised *feature* vectors of real numbers $v_i \in \mathbb{R}$ for primary information encoding. This avoids information loss during code space layout, and helps to obtain the smoothest possible cluster structure.

#### 2.0.1. Population coding

Each vector represents a single discrete concept that encodes an entity or phenomenon from the subject domain.

Unlike $n$-hot encoding [22], individual vector bits are assigned randomly[7] and do not mean anything independently.

Only a non-random combination of several otherwise random bits is considered semantically meaningful. This approach has its advantages, including the ability to encode many concepts.

For example, with a bit vector of 128 bits and 12 bits set, it is possible to encode $\binom{128}{12} \approx 2.37 \times 10^{16}$ unique discrete concepts. This is more than enough to describe most objects in the real world.

In practice, codes with similarity properties must describe conceptually different entities; their codes must be unique and sufficiently distant from each other in terms of Hamming [23].

But even in this case, the estimate of the spherical packing boundary [24] for constant-weight codes, for distances $d \geq 5$ gives an order of $10^{11}$.

### 2.1. Codes and similarity

The code space is expected to be organised in such a way that conceptually similar entities are mapped to similar codes[8]. That way it will be possible to operate with complex concepts just by performing simple bitwise operations on their codes.

Methods for constructing such codes are described in Chapter 4.

#### 2.1.1. Formal definition of similarity

Let $D$ be a set of objects from the initial domain, and $M$ be a set representing a mathematical model that describes entities and phenomena from $D$.

Let $V$ be the set of vectors used to represent elements from $D$ in model $M$.

Let us define a mapping $f : D \to V$ that assigns each entity or phenomenon from $D$ a corresponding vector from $V$.

Let $d_D : D \times D \to \mathbb{R}$ be a metric that defines the similarity between entities and phenomena in the source domain.

Let $d_V : V \times V \to \mathbb{R}$ be a metric that defines the similarity between vectors in model $M$.

[6] We also studied the structural coding of human speech. While we obtained interesting results, due to length constraints, we decided to publish them in a separate paper.

[7] This is only true for unique concept codes. The detector codes discussed in Chapter 6 can be considered n-hot, where each active bit expresses the concept's membership in the codes that activate the corresponding detector. However, this does not make much practical sense.

[8] Gray codes work similarly [25, Section 7.2.1.1]

The coding system should be organised in such a way that for all $x, y \in D$, if $d_D(x, y)$ is small, that is, $x$ and $y$ are close in the domain, then $d_V(f(x), f(y))$ should also be small, i.e., the vectors $f(x)$ and $f(y)$ are close in the model $M$:

$$\exists \varepsilon, \delta \in \mathbb{R}^+ : \forall x, y \in D,$$
$$d_D(x, y) \leq \varepsilon \iff d_V(f(x), f(y)) \leq \delta$$

Here, $\varepsilon$ and $\delta$ are the threshold values determining the desired similarity in the domain and model, respectively.

## 2.2. Operations

Functions can be defined over a set of bit vectors, which allows individual concepts to be grouped into descriptions, and various operations can be performed on them.

### 2.2.1. Conjunction (bitwise OR)

To encode a *description* comprised of several discrete concepts, an element-wise conjunction operation can be used:

$$\mathbf{a} \vee \mathbf{b} = (a_i \vee b_i)_{i=1}^{n}$$

If the saturation (amount of set bits) of the source code is low, this is sufficient. For the cases of higher saturation, a colour merge operation (Section 3.3) was defined that allows us to construct complex descriptions and merge many codes without over-saturating the result.

### 2.2.2. Intersection (bitwise AND)

To test whether a particular concept belong to a complex description, an element-wise disjunction operation is can be used:

$$\mathbf{a} \wedge \mathbf{b} = (a_i \wedge b_i)_{i=1}^{n}$$

As with Bloom filters [26], this operation is probabilistic.

The higher the code density and the more elements in the description, the higher the probability of *collisions*. This is the main reason why vector lengths and their densities should be considered carefully.

### 2.2.3. Concatenation

In some cases, it may be necessary to combine concepts from different domains, such as an object's code and the code of its position in space. However, code lengths may vary, making conjunctions undesirable or impossible.

In this case, it makes sense to combine the vectors into a tuple (concatenate them), resulting in a longer code:

$$(\mathbf{a}, \mathbf{b}) \equiv (a_1, a_2, ..., a_m, b_1, b_2, ..., b_n).$$

Here, $m$ and $n$ are the number of elements in vectors $\mathbf{a}$ and $\mathbf{b}$, respectively.

The positional encoding methods adopted in neural network models, either mix-in the position code into the embedding (like sinusoidal codes of a classical transformer [27, p. 3.5], or trainable codes in BERT [28]) or change the embedding (like RoPE [29]).

In our case, we also have a choice: either merge the codes for the concept and position or concatenate them.

Code merging preserves the vector's original length but increases code density and the likelihood of collisions. Concatenation, on the other hand, preserves the original code intact but increases the overall length of the code.

### 2.2.4. Measures of similarity

Given two vectors, we can calculate the measure of their similarity $V \times V \to \mathbb{R}$ and thereby estimate the conceptual similarity of concepts from the domain $D$.

The Jaccard index and the discrete analogue of the cosine measure work well as similarity functions for bit vectors.

#### 2.2.4.1. Cosine similarity

Traditionally, for machine learning tasks, the cosine similarity is well-suited for continuous vectors:

$$S(\mathbf{a}, \mathbf{b}) = \frac{\sum_i a_i \cdot b_i}{\sqrt{\sum_i a_i^2} \sqrt{\sum_i b_i^2}}.$$

In some cases, when working with normalised vectors, we used a less strict variant:

$$S'(\mathbf{a}, \mathbf{b}) = \frac{\sum_i a_i \cdot b_i}{\sqrt{\sum_i a_i \cdot \sum_i b_i}}.$$

The discrete version of the cosine measure is defined as follows:

$$C(\mathbf{a}, \mathbf{b}) = \frac{|\mathbf{a} \wedge \mathbf{b}|}{\sqrt{|\mathbf{a}| \cdot |\mathbf{b}|}}$$

For the edge case where the denominator is 0, we treat the entire function as 0.

#### 2.2.4.2. Jaccard index

In general, it is defined as the ratio of the number of elements in the intersection of sets to the number of elements in their union:

$$J(A, B) = \frac{|A \cap B|}{|A \cup B|},$$

For vectors in $\mathbb{R}^n$, this will be

$$J(\mathbf{a}, \mathbf{b}) = \frac{\sum_i \min(a_i, b_i)}{\sum_i \max(a_i, b_i)}, \quad a_i \in A, \ b_i \in B.$$

Similarly, for bit vectors:

$$J(\mathbf{a}, \mathbf{b}) = \frac{|\mathbf{a} \wedge \mathbf{b}|}{|\mathbf{a} \vee \mathbf{b}|},$$

where $|\mathbf{v}|$ is the number of ones in the vector[9].

To enhance the influence of individual peaks and suppress noise, it makes sense to use a quadratic version of the Jaccard index:

$$J_2(\mathbf{a}, \mathbf{b}) = \frac{\sum_i a_i \cdot b_i}{\sum_i \max(a_i^2, b_i^2)}.$$

### 2.2.5. Fuzzy search

With a bunch of binary vectors, we can perform fuzzy search on them, just like we do with embedding vectors in modern vector databases.

Formally speaking, for a certain similarity metric $d_V : V \times V \to \mathbb{R}$ and a certain similarity threshold $\varepsilon$, it is possible to define a mapping $S : V \to \mathcal{P}(V)$, which, given a code $v \in V$, returns the set of all codes $v' \in V$ that are close to $v$ with an accuracy of $\varepsilon$:

$$S : V \to \mathcal{P}(V),$$
$$S(v) = \{v' \in V : d_V(v, v') \leq \varepsilon\}.$$

A detailed description of the algorithms is beyond the scope of this article, but we mention two methods we used:

1. Random subspaces
2. Mask hierarchy (search tree)

The first method is structurally similar to a hash table, in which each bucket corresponds to a random bit mask of a specific density, and all elements of one bucket are comparable to each other by that mask.

The second method uses a complex hierarchy of masks to construct a multi-root tree (forest) that allows for efficient fuzzy searching.

## 2.3. Application

It is possible to use sparse bit vectors for encoding of various concepts of different nature.

The main requirement is that the code space's topology be as similar to the domain's topology as possible.

### 2.3.1. Association coding

The simplest way to encode a connection between two concepts is to combine their codes, either by merging or by concatenating them.

Let $\mathbf{a}$ be a vector representing the concept of "apple" and $\mathbf{r}$ be a vector corresponding to the colour "red".

Then $\mathbf{a} \mid \mathbf{r}$ will describe[10] the concept of a "red apple". The same can be done using concatenation or tuples: $(\mathbf{a}, \mathbf{r})$.

Associative queries can be performed after storing the resulting vectors in *memory* (vector database).

For example, to find out which objects in memory are red, we perform a fuzzy search using the code mask $\mathbf{r}$ or, in the case of tuples, $(\varnothing, \mathbf{r})$. Here, $\varnothing$ denotes an empty vector consisting of zeros.

[9] This is similar to the Sørensen index, but unlike the Jaccard index, the former does not satisfy the triangle inequality. Since we use these measures for geometric code layout, this is a significant argument against using the Sørensen index.

[10] Here and below, the operator | denotes the conjunction or colour merging operation, depending on the types of codes and the tasks to be solved.

This is how associative sets can be encoded. To preserve the order of association entries, a list can be used.

### 2.3.2. Lists

Chains of associations can be used to encode ordered sequences of concepts or numbered lists.

Here is an example of a list that use code merging:

- $\top \mid \text{Richard}$
- $\text{Richard} \mid \text{Of}$
- $\text{Of} \mid \text{York}$
- …
- $\text{In} \mid \text{Vain}$

And here is the tuple variant:

- $(\top, \text{Richard})$
- $(\text{Richard}, \text{Of})$
- $(\text{Of}, \text{York})$
- …
- $(\text{In}, \text{Vain})$

In both cases, the beginning of the sequence is marked with a predefined code $\top$, and the subsequent elements are encoded in pairs. This allows us to store a list of any length in memory, but only one.

To record multiple lists, we add a unique list identifier to $\top$ and each first element in the pair:

- $\mathbf{id} \mid \top \mid \text{Richard}$
- $\mathbf{id} \mid \text{Richard} \mid \text{Of}$
- …

#### 2.3.2.1. List traversal

In order to get the heads of all lists, we need to perform a fuzzy search of $\top$. To get the contents of a specific list, we just need to search for **id**.

It is possible to reconstruct the entire sequence by going through the pairs individually, starting with $\top$.

This is equivalent to performing topological sorting [30, ch. 22.4], if the elements of the pairs are interpreted as nodes and the pairs themselves as edges of a certain directed acyclic graph (DAG).

If a reverse pass is required, we add an element containing the code $\bot$ to the memory, from which a reverse chain of associations can be constructed. For the example above, this is the pair $\text{Vain} \mid \bot$.

#### 2.3.2.2. Indexed access

In order to access any item in the list by an index, we augment the first item in the pair with the index.

The length of the list can be encoded by adding the known value $\bot$ to the index of the last pair:

- $(1 \mid \mathbf{id}, \text{Richard})$
- $(2 \mid \mathbf{id}, \text{Of})$
- …
- $(7 \mid \mathbf{id} \mid \bot, \text{Vain})$

By performing a fuzzy search of the key $\mathbf{id} \mid \bot$, the associated index 7 can be found.

If desired, indexes can be assigned only to certain elements in the list. That way it would be analogous to an indexed skip list[11] [31].

### 2.3.3. Graphs

Graphs and hypergraphs can be encoded using association chains. This is equivalent to specifying a graph using a list of edges.

While the most economical option for traversing a list requires only $O(1)$ memory, traversing a graph requires additional $O(n)$ memory to store previously visited nodes.

If the graph is undirected, it is more reasonable to use merged elements instead of tuples, should the codes density allow it.

#### 2.3.3.1. Topographic maps

Graphs can represent a map of the terrain and possible routes between points.

For example, on Figure 1, a path to work can be encoded by a chain of associations:

(🏡, 🌳), ⛲, 🏓, 🏢.

#### 2.3.3.2. Pathfinding

A path in a graph can be found by associatively extracting edges by place code and traversing edge lists. In this sense, the algorithm resembles $A^*$ [32].

[11] A data structure that combines the advantages of an array and a list.

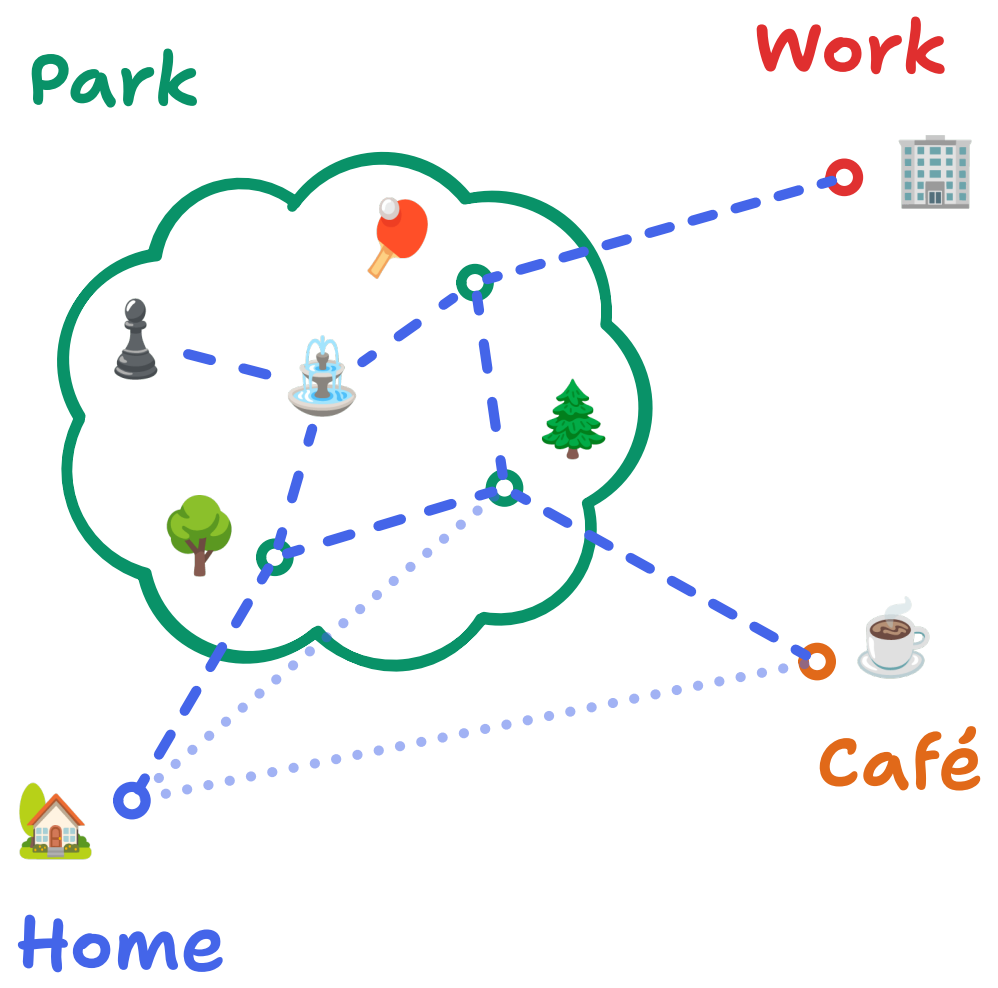

Figure 1: Terrain map with paths.

Chains of secondary associations can generate structures resembling skip-lists and contraction hierarchies[12] [33], [34].

This allows associative chains to be constructed with minimum memory accesses.

For example, the route from home to work can be reduced to 🏡, ⛲, 🏢 whereas from home to the café to 🏡, 🌲, ☕.

#### 2.3.4. Numbers

To encode natural numbers a lexical variant is suitable, in which numbers are encoded through their symbolic representation in a given number system.

In this case, each digit $a$ in each digit position $i$ is assigned its own unique code $a_i$, that is:

$$\forall a \in \{0, 1, ..., 9\}, \forall i, j \in \mathbb{N},$$
$$a_i = a_j \Leftrightarrow i = j.$$

For example, the number 42 can be represented as $4_1 \mid 2_0$, and the number 101 as $1_2 \mid 0_1 \mid 1_0$. Zeros (except for the *number* 0) can be omitted from the encoding, if necessary.

To represent negative numbers, it is sufficient to define a standard “minus” element and add it to the number code: $- \mid 1_0$.

Rational numbers can be encoded by entering separate codes for the numerator and denominator:

$$\Phi \approx \frac{21}{13} = 2_1|1_0|1_1^{-1}|3_0^{-1}.$$

To encode real numbers, their lexical representation can be used as decimal fractions:

$$\pi = 3_0|1_{-1}|4_{-2}|1_{-3}|5_{-4}|9_{-5}|...$$

This method of representing numbers allows us to evaluate the similarity of numbers lexically, by comparing their codes, but it has drawbacks. In particular, in such encoding, the codes for the numbers 123 and 1000023 may be closer to each other than 123 and 234. That is, such a mapping does not preserve the original similarity metric.

To resolve this situation, it is necessary to encode all zeros or assign more bits to the higher orders so that the code reflects the significance of individual digits.

It is essential to understand that evertyhing described here is for *primary encoding*. The goal is to obtain a description that is convenient for further processing and has a similarity property. By itself, it will not allow arithmetic operations to be performed — this is the task of the model.

Another variant of encoding numbers using wide detectors is described in Section 4.2.1.

#### 2.3.5. Characters and words

Similar to numbers, words can be encoded as combinations of positional and character codes:

$$h_0|e_1|l_2|l_3|o_4$$

Unlike numbers, indexing here is done from left to right, in the natural order of characters.

A more complex coding option that considers words’ morphological similarity is discussed in Section 7.1.

## 3. Chromodynamics

As it was shown before, encoding complex descriptions require combining the codes of individual concepts (Section 2.2.1).

[12] A method for optimising path search in a graph based on the preliminary generation of a hierarchy of virtual edges. Instead of exploring a dense graph, a search is performed on a small subset of virtual edges.

In the simplest case, a simple conjunction is sufficient for this purpose. However, as the number of elements to be combined increases, the saturation of the resulting code grows rapidly and becomes a problem.

This can be solved in different ways, such as increasing the length of the code or reducing the density of a single element. This works, but it is not always possible due to practical reasons. The main issue is that reducing the code length and density inevitably affects other essential properties of codes, described below.

A more interesting approach is to use the redundancy property of binary codes to selectively filter individual bits in the concept codes and thus, obtain a union code of a given saturation that still preserves enough information about the individual elements and their similarity.

By continuing the glorious tradition of confusing the reader, we have called such an approach *chromodynamics*, by analogy with quantum chromodynamics, which operates with the concept of *colour charge* in quarks. In both cases, "colour" has nothing to do with physical colours, but is convenient for describing the phenomenon's essence.

## 3.1. Code requirements

At first glance, our binary codes must combine several contradictory properties:

- Concept codes must have *significant overlap* with other conceptually close codes, so that the similarity function (Section 2.2.4) can run smoothly and yield values over the entire range. In addition, codes should be comparable both, to close and to relatively distant concepts.
- Concept codes should be sensitive to small changes. Ideally, a change in a signal (stimulus) at the resolution limit in the source domain should result in a change of at least one bit in its code.
- Finally, the codes should be of reasonable length and density. Otherwise, combining them without oversaturation or deteriorating properties would be nearly impossible. Codes that are too long and dense are also undesired because of the difficulties in storing, processing, and implementing fuzzy search (Section 2.2.5).

All these issues are uncompromisingly solved by adding another virtual coordinate, *colour*, to the discrete codes[13].

## 3.2. Colour encoding

The basic idea of colour coding is that for each bit in a code, a colour is assigned, that expresses relative "importance" and "priority" of the associated bit. The colour is used during the colour merge procedure.

It is important to note, that colours do not affect the information component of the code. Coloured-code contains as much semantic information as colorless code.

The specific order in which colours are assigned to bits depends on the domain and practical implications. Examples of colour coding are discussed in Chapter 4.

## 3.3. Colour merge

The main idea is to select bits based on their "importance" and available saturation "budget".

If the total code saturation is sufficient to accommodate all bits and does not exceed a threshold $t$, the result is equivalent to a simple conjunction:

$$\mathbf{a} \mid \mathbf{b} = \mathbf{a} \vee \mathbf{b}, \text{ if } |\mathbf{a} \vee \mathbf{b}| \le t.$$

Otherwise, the bits are filtered using some filter function $f$:

$$\mathbf{a} \mid \mathbf{b} = (f(a_i, b_i))_{i=1}^{n}, \text{ if } |\mathbf{a} \vee \mathbf{b}| > t.$$

### 3.3.1. Short-range and long-range order

As shown in Chapter 5, successful layout requires concept codes to have similarity over a wide range.

This means that codes must be comparable on the short-range scale of neighbouring codes and, at the same time, on the scale of the whole code space. Colour codes can help tackle this problem.

[13] At the neurophysiological level, this difference can potentially be expressed by the composition of neurotransmitters of a given code. Potentially this can also be one of the reasons for neuronal co-transmittion [35], [36] in the central nervous system.

In this sense, it is possible to compare the bits pseudo-colours with frequencies and wavelengths of separate harmonics and their medium propagation properties.

The red part of the code “spectrum” corresponds to longer wavelengths and the long-range order of code comparison, while the shorter-wavelength part corresponds to the short-range order.

In other words, long-range order gives comparability, and short-range order provides uniqueness.

#### 3.3.2. Order-aware merging

Depending on goals, bits can be filtered from one end of the “spectrum” or the other:

- To preserve more meaningful bits, one can filter the long-wavelength part while preserving the short-wavelength bits.
- If the goal is to obtain a description that makes sense as a whole, the short-wavelength bits can be discarded.
- In other cases, it may be necessary to preserve the average scale, sacrificing the long-range and short-range order.

## 4. Wide Detectors

The ultimate goal of coding is to obtain a code system that fulfils the requirements outlined in the chapter on chromodynamics (Section 3).

Such a code system can be defined in various ways, including table-, geometric- or analytical definition. Analytical definition is usually sufficient for simple code systems (linear, one-dimensional).

However, geometric methods are the most effective in our practice, especially if the space's topology is non-trivial, has circular coordinates, or includes more than two dimensions.

### 4.1. The idea of wide detectors

A detector is a discrete entity or function that maps its *receptive field* into one or more bits of its output code.

An important structural feature is that the receptive fields of detectors noticeably overlap. Therefore, any given stimulus typically activates several detectors at once.

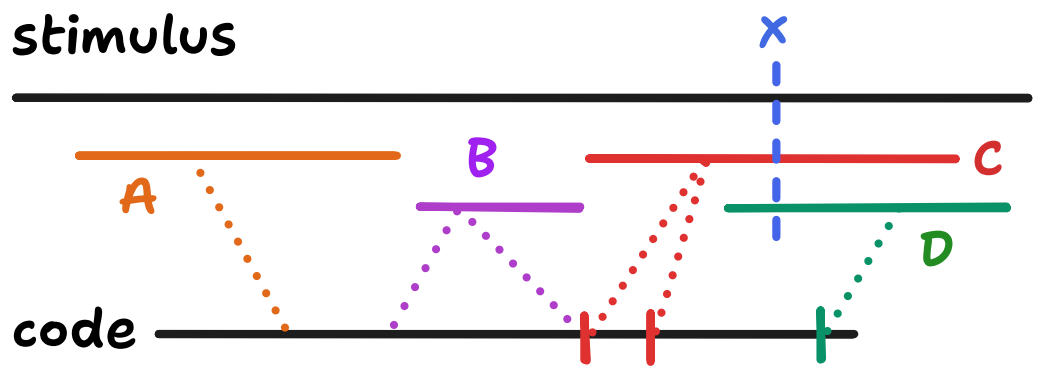

Figure 2: Example of encoding a one-dimensional stimulus. $A$, $B$ — inactive detectors; $C$, $D$ — active detectors; $x$ — stimulus.

A stimulus description is obtained by combining the codes of the active detectors (Section 3.3). On Figure 2, the stimulus value $x$ occurred at the intersection of detectors $C$ and $D$, so both detectors were activated and added their bits to the output code.

The idea of detectors did not arise by chance. In neurophysiology, many structures are known to behave in a similar way. For example, hair cells in the cochlea are mechanical receptors that convert acoustic vibrations of the basilar membrane to electrical signals [37], [38].

Another example is retinal ganglion cells, which aggregate signals from amacrine and bipolar cells in their receptive field [39], [40].

In both cases, the receptive fields of individual cells do overlap significantly [41]. Thus, the ensemble activation of several detectors can be used to estimate the localisation and type of the stimulus.

The more detectors cover the perceptual field and the more densely they overlap, the greater would be the spatial resolution of encoded values.

### 4.2. One-dimensional codes

Previously we showed an example of encoding a one-dimensional stimulus. In general, the stimulus space can be discrete or continuous and have any number of dimensions.

Continuous stimulus spaces are usually used in the case of direct mapping of values from the real world.

Discrete ones are useful when processing initially discrete data and for codes of descriptions obtained from previous level of the model hierarchy.

### 4.2.1. Encoding of integers

Numbers can be encoded in several ways.

In Section 2.3.4, a variant was described in which numbers are encoded lexically, through their symbolic representation in a given number system.

This option is generally good for encoding large or infrequently used numbers. However, it is unsuitable for their direct processing because its code space topology does not preserve the original similarity metric (Section 2.1.1).

Encoding through wide detectors is better suited for values of well-known and fixed range (e.g., instrument scales, sensor values).

In this way, it is possible to select the optimal overlap value and adapt the code space topology to the stimuli topology and actual scale (linear, logarithmic). In this case, the similarity of codes will, to a certain degree, correspond to the similarity of initial values.

The Figure 3 shows the logarithmic detector space and examples of stimulus encoding ranging from 0 to 1000.

For clarity, the detectors are shown without overlap. In reality, detectors must overlap at each level of the hierarchy.

The detector space resembles the position codes used in neural networks [27, p. 3.5], and the linear absolute encoder.

Unlike position codes at classical transformer [27, Section 3.5], we do not deal with sinusoids but random bits. Unlike the encoder, we encode position only in ones; zeros are meaningless in our codes.

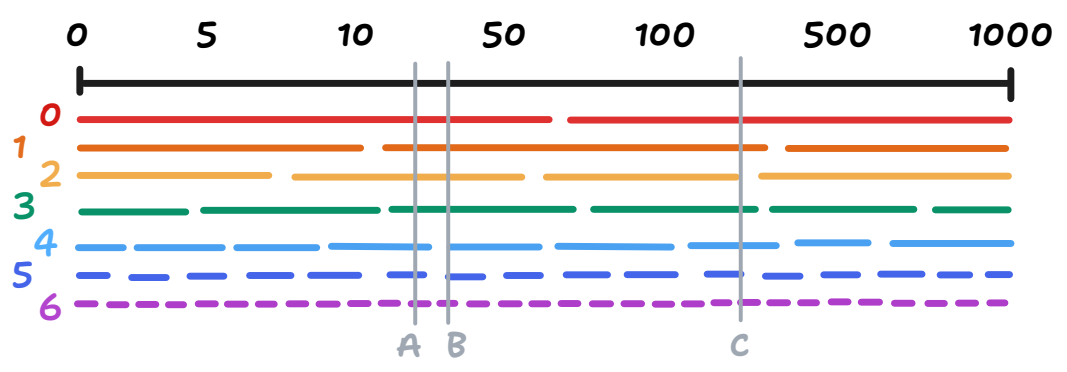

Figure 3: One-dimensional logarithmic detector space.

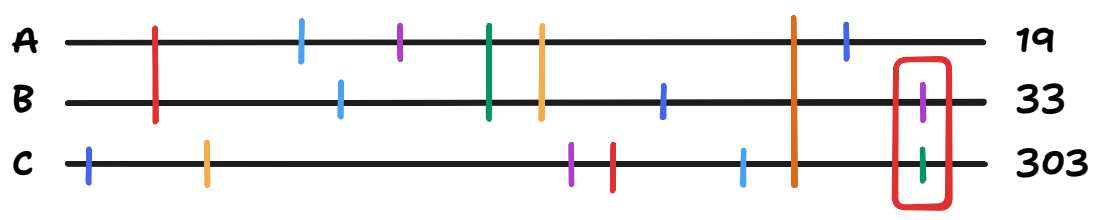

Figure 4: Common code bits and a collision.

Each detector is matched to one random bit of the output code, so that for each stimulus, there are, on average, about seven set bits in its code.

On Figure 4, the values $A$ and $B$ are close, so they have four bits in common at levels 0, 1, 2 and 3 in the codes. Code $C$ is far away, so it has only one common bit at level 1.

Codes $B$ and $C$ accidentally got one extra common bit due to *collision in detector codes* at levels 3 and 6, marked by a rectangle in the figure.

The "longer wavelength" part of the spectrum maintains the codes' comparability, while the "shorter wavelength" part ensures their uniqueness.

The number of detectors, their overlap, the size and density of the output code are determined experimentally, considering expected number of elements in a description, desired resolution, and predicted collision probability.

Collisions lead to parasitic similarity of codes and increase the noise level. It can be reduced by increasing the code density, adding extra layers of detectors, or increasing the number of bits per detector.

Spatial resolution of a code can be increased by reducing the size of receptive fields and increasing the number of detector layers. However, it is important to remember, that by shrinking receptive fields we also reduce the detector overlap, and thus, negatively affect the long-range order (Section 4.5).

### 4.2.2. Real numbers and the fractal nature of encoding

In the example on Figure 5, there was a need to encode values between 2.71 and 3.14 more accurately. At this scale, even layer 6 does not provide the necessary resolution. Therefore, additional detector layers 7, 8, and 9 were added.

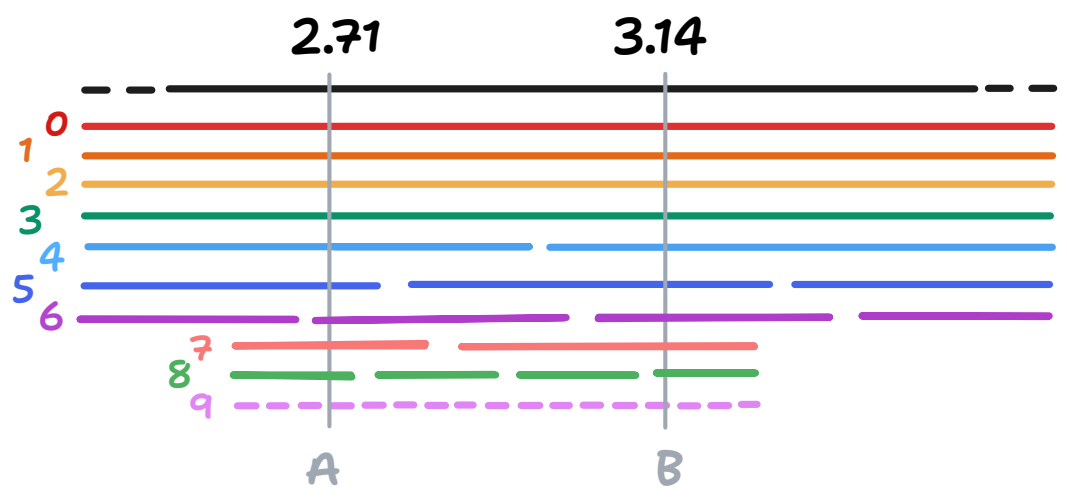

Figure 5: Additional layers of detectors.

An interesting feature is that detectors can be added dynamically and only for a specific area, if desired.

## 4.3. Two-dimensional position codes

It is possible to implement two-dimensional spatial codes similarly to one-dimensional codes. In this case, the detectors would be circles on a plane, instead of line segments.

The Figure 6 shows the active subset of detectors used to encode the position of point $A$ and its code.

For clarity, only active detectors and some inactive detectors are shown. In reality the whole space is filled with detectors of all hierarchy levels.

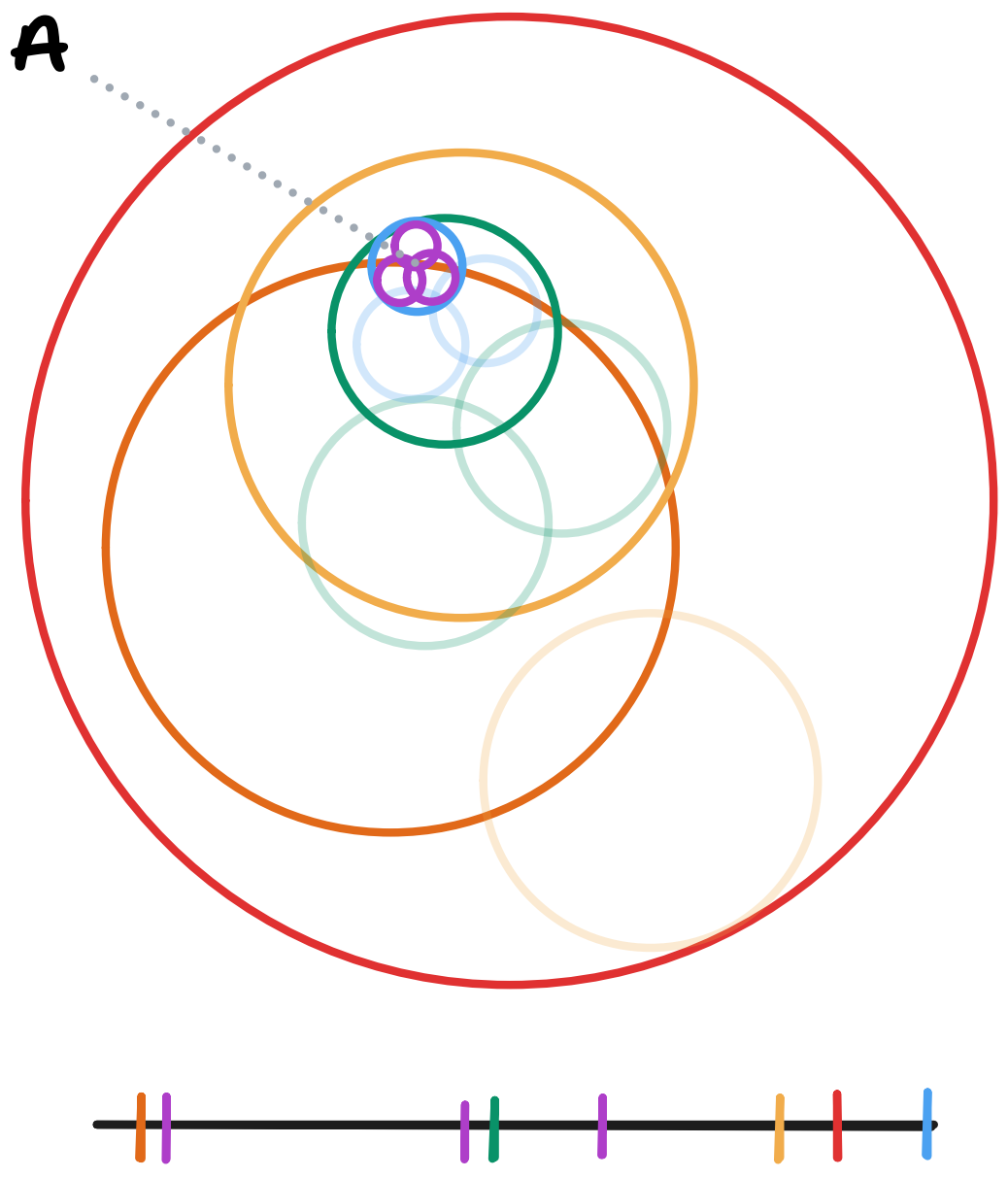

Figure 6: Two-dimensional position codes.

The nuances of constructing such a detector space are discussed in Chapter 6.

## 4.4. Cyclic coordinates and gradient codes

Even spaces with complex topology, such as cylindrical or toroidal, can be described in codes.

In such case, one or more spatial coordinates would be cyclic, and their codes should change smoothly. At the same time, the similarity property is still preserved in codes.

In engineering, a similar design is used in an absolute angle encoder. As the encoder shaft rotates, a disc, with a certain pattern on it, yields Gray codes [25, Section 7.2.1.1]. Notably, small changes in shaft position result in small changes in the output code. Specifically, adjacent positions always yield codes that are different by exactly one bit.

### 4.4.1. Sliding window method

The simplest way to obtain a topologically closed code space is to use a sliding window.

*Producing elements* that fall within the window are mapped to the output code. The window must be boundary-closed to produce a topologically closed code space (Figure 7).

A two-dimensional code with one topologically closed coordinate can be implemented similarly. For this purpose, the producing elements must be placed on a cylindrical surface.

On Figure 8, the $x$ coordinate specifies the angle and $y$ specifies the offset along the cylinder axis.

Moving the window along such a surface allows one to obtain a code representation for each of its points. The resulting code space will be topologically closed along the same coordinate as the generating cylinder.

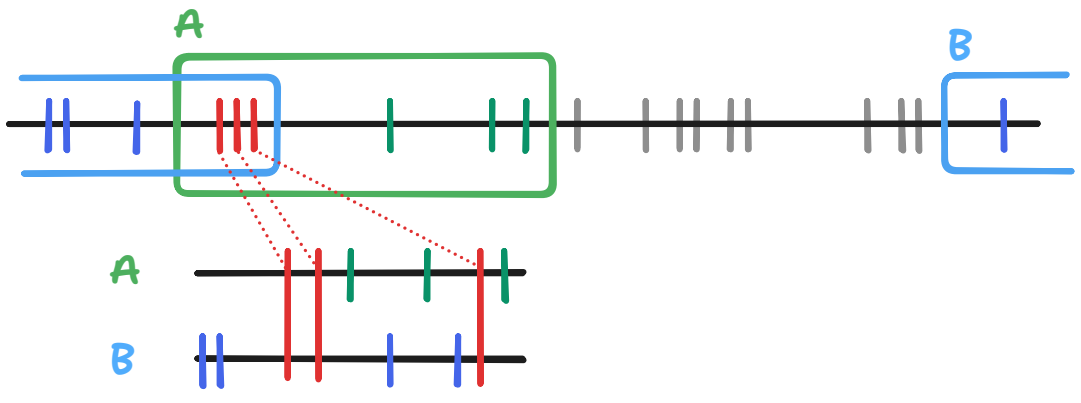

Figure 7: Sliding windows yield common bits.

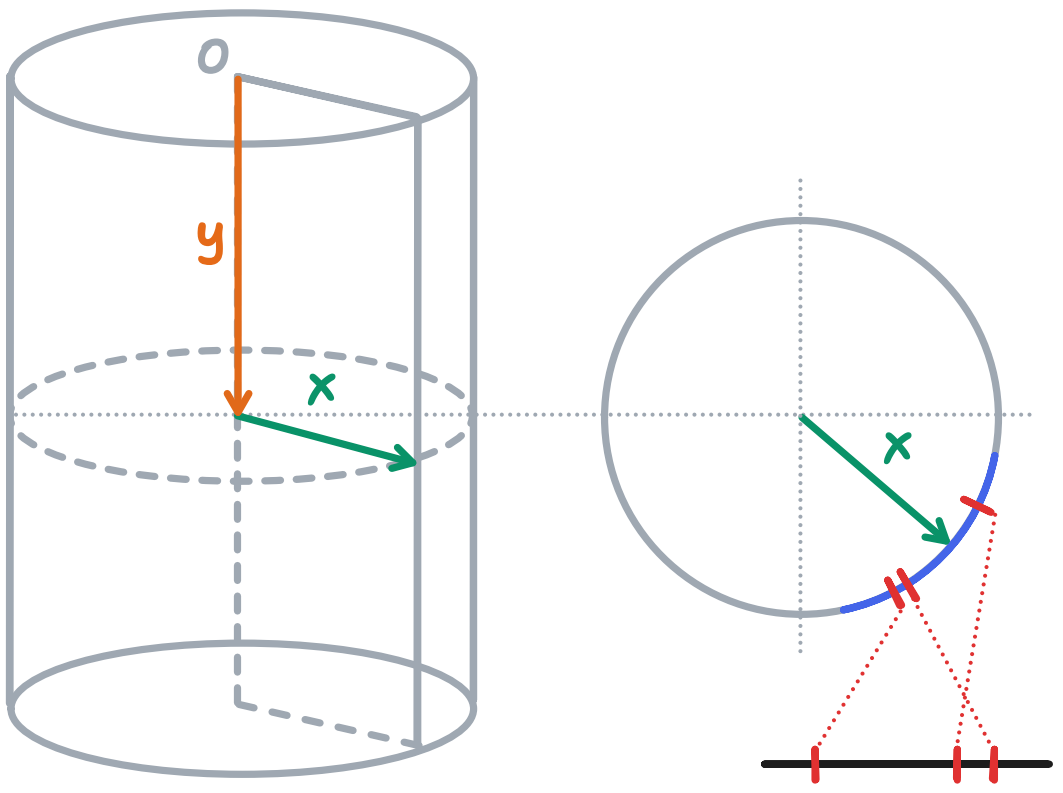

Figure 8: Generating cylinder.

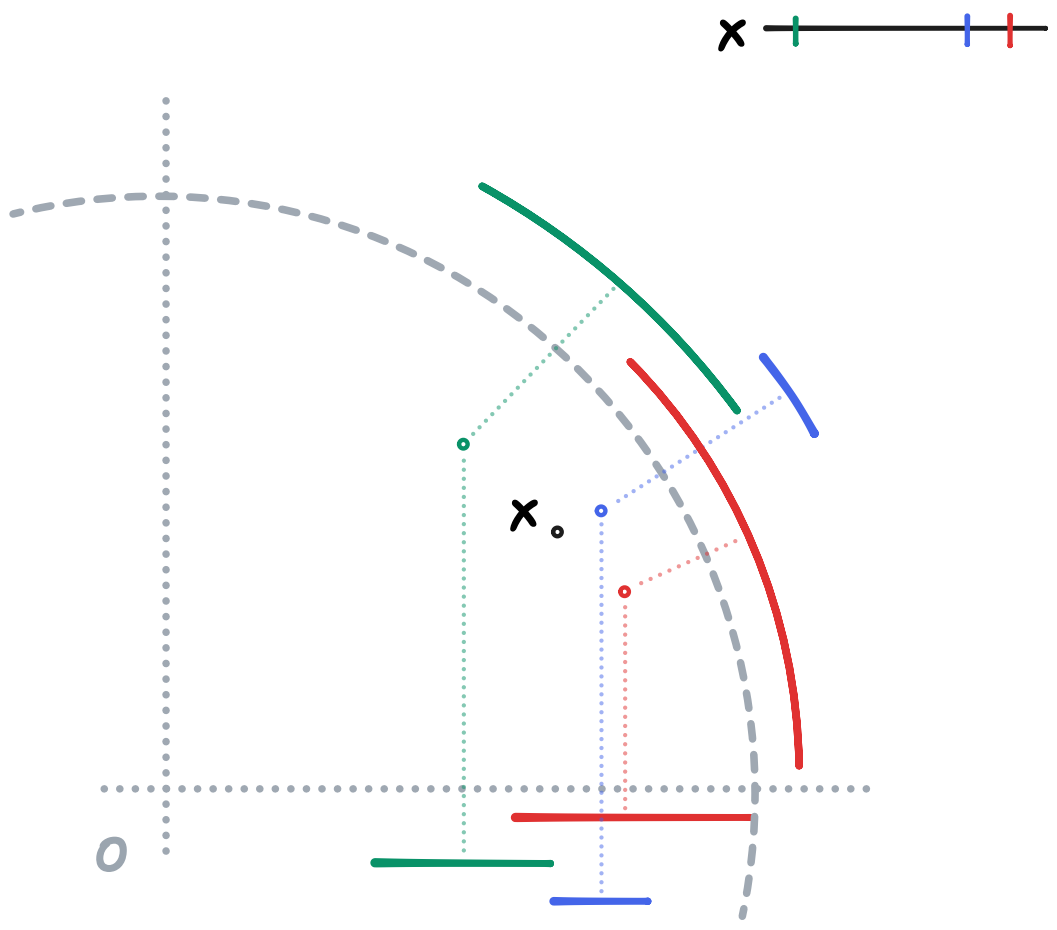

Figure 9: Encoding in polar coordinates.

If similarity must be ensured not only along $x$ but also along $y$, then instead of the secant plane, the region of the cylinder in some neighbourhood of $y \pm \delta$ should be chosen.

A torus or sphere can be used as a generating surface to obtain a topologically closed space in two coordinates.

In general, this method works, but the resulting codes have disadvantages:

- It isn't easy to control the density of the codes and the amount of overlap.
- The codes are not well compatible with colour-merging (Section 2.2.1).
- The code space comes out non-planar, so it is poorly suited for layout and detection. It is possible to use a conical producing surface to planarise the space, but this still does not solve all the problems.

#### 4.4.2. Geometric method

A more successful method of generating a topologically closed code space combines the aforementioned variants.

The plane where the detectors are located is used as the generating space. Each detector in polar coordinates is defined by its centre and the dimensions of the receptive field: $(a \pm \phi, r \pm \rho)$.

The larger receptive field corresponds to the red part of the "spectrum" and the smaller one to the violet part.

The Figure 9 show the encoding of point $x$ using three wide detectors. The boundaries of the detectors' receptive fields are shown as arcs of circles and projections on the axis.

As shown in ch 5, the code space obtained with such detectors is planar and can be decomposed into a pinwheel without folds.

#### 4.4.3. Component-wise assignment

The closure by angle can be obtained by specifying it component-wise. In this case, each of the coordinates is specified by a one-dimensional code, and the description code is obtained by combining the component codes:

$$a = f_1(\sin \phi) \mid f_2(\cos \phi) \mid f_3(\rho).$$

### 4.5. Short-range and long-range order

All the code systems described above have one crucial property — similarity in both near and far scales.

This means the code space formed by such detectors will have similarity between neighbouring and far-apart components.

To complete the picture, let's check examples of degenerate code spaces with some or other drawbacks.

- If, for example, we remove detector layers 1-6 from the linear space on Figure 3, leaving only 0 and 1, we get a space with only long-range order. In this case, the resolving power of the space will be very low. Thus, all points in the interval 100-1000 will have codes, at best, differing only by 1 bit.

- If we remove layers 0-4, leaving only 5 and 6, the codes will have only near-neighbour order. Neighbouring values will have similar codes, but distant values will be incomparable.

## 4.6. Multidimensional codes and continuous spaces

A space of n-dimension and any topology can be encoded using wide detectors.

However, in some cases, it makes sense to perform primary encoding in a continuous space and obtain codes after layout and detection (Chapter 6).

One of such examples is sound. Suppose we transform a sound signal from a temporal domain into a frequency domain, for example, using the discrete Fourier transform [42], [43]. Each temporal slice can be represented as a separate multidimensional vector, where each dimension has its own decomposition coefficient.

Interestingly, the values of Fourier decomposition coefficients can be considered activation levels of wide detectors in the presence of a given frequency within the signal. In this case, neighbouring frequencies will activate neighbouring detectors, even if frequencies match inaccurately. The same applies to the energy of mel filters.

This is discussed in detail in Chapter 7.

# 5. Code Space Layout

The manifold hypothesis states that a multidimensional dataset often corresponds to a nested subspace of lower dimension [44].

In other words, the eigen-dimensionality of the data is generally less than the dimensionality of the space in which it is defined.

This observation can be used not only for visualisation, but also for obtaining efficient domain-specific codes that optimally describe the space's topology while preserving the similarity property (Section 2.1.1).

## 5.1. Problem statement and requirements

The main idea of the layout is to obtain a mapping of a multidimensional code space onto a two-dimensional plane in a form convenient for subsequent detection (Chapter 6).

This way, dimensionality reduction and topological transformation of the stimulus code space into the detector code space are achieved.

We formulated the following requirements for the layout algorithm, which ultimately determined the features of the implementation.

### 5.1.1. Biological motivation

The algorithm mimics the operation of the neocortex. Within our model, each cortical minicolumn [45], [46] is associated with a single code.

Similar to afferent sorting [4], rearranging two points (codes) in place is an elementary step in organising our maps.

This largely determines the discrete nature of the algorithm.

### 5.1.2. Discrete nature

The algorithm should work on a discrete space (matrix), where each cell represents one cortical minicolumn. Each cell can contain only one code or be empty. The number of cells (cortex size) is limited and set externally.

One code represents a relatively short vector: in the case of binary vectors it is 128-256 bits, in the case of $\mathbb{R}^n$ feature vectors it is tens of elements.

Movement of codes is possible only by exchanging the contents of cells. The position of a point in the code space is fully determined by the coordinates of a cell holding its code. Thus, the layout algorithm resembles the classical cellular automaton [47], [48].

Nevertheless, we allow non-local exchange within the model when far-away points are rearranged.

### 5.1.3. Compactness and hierarchy

The layout algorithm should generate compact, densely packed clusters. Clusters should organise in a hierarchical structure.

This is important for building detector space. Dense packing saves cortical area and simplifies activation. The hierarchy of clusters allows the construction of a corresponding hierarchy of detectors.

In addition, the algorithm must handle the highly inhomogeneous structure of the code space, which can contain both very small and very large clusters.

#### 5.1.4. Locality

Although we allow the exchange of far points, in the limit where the long-range order is already consolidated, it is desirable that the algorithm can efficiently handle a local geometric neighbourhood without having to compute $O(n^2)$ interactions and traverse the entire code space.

The algorithm should work off the current state of the space without significant additional memory and with minimal preprocessing.

#### 5.1.5. Simplicity is more important than efficiency

We did not aim to build the most efficient[14] or universal dimensionality reduction algorithm capable of replacing classical methods such as PCA [49], t-SNE [9] or UMAP [2].

However, we wanted to design a conceptually simple algorithm that could be executed on a cellular automaton, would not require additional memory, and would be reasonably efficient on short vectors.

### 5.2. DAMP layout algorithm

The core of our algorithm[15] resembles the optimisation phase of the UMAP algorithm and combines features of two-dimensional sorting with simulated annealing [50], [51].

By exchanging a randomly selected pair of points, it is possible to estimate the energy impact of such an exchange and keep the beneficial variant.

The Figure 10 schematically depicts a partially organised code space consisting of two clouds (clusters) and a pair of unsettled points $A$ and $B$, currently far out of their place.

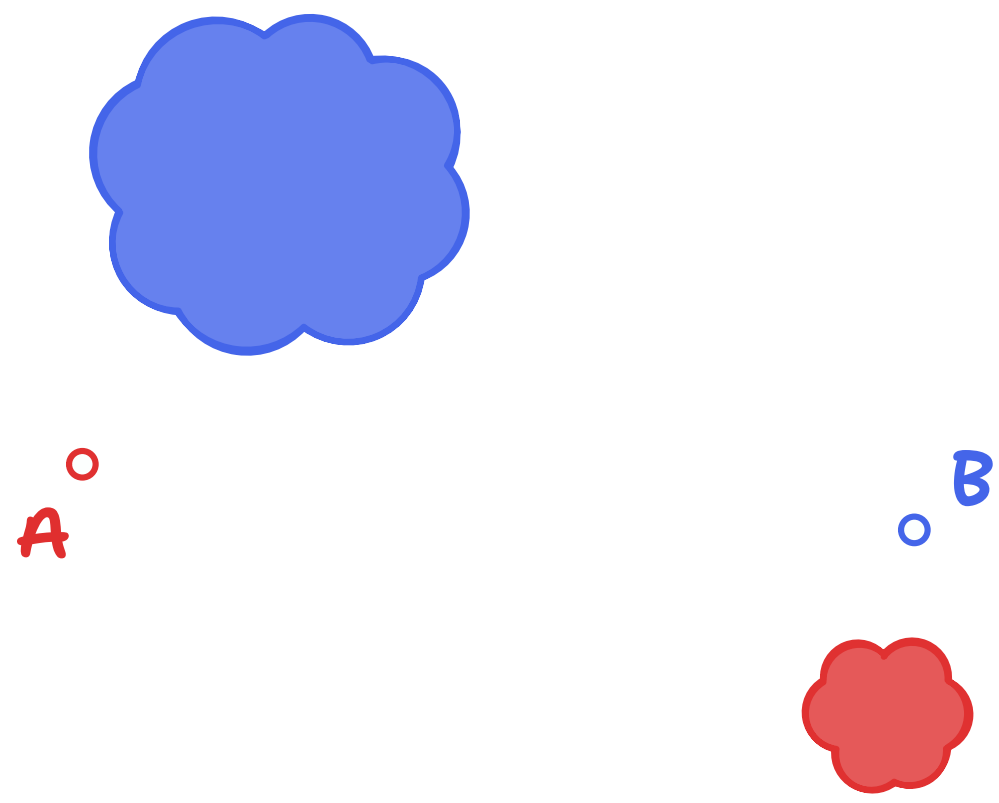

Figure 10: Point clusters.

If points $A$ and $B$ are swapped, the system will be in an energetically favorable position because the points will be closer to the clouds of their colour.

### 5.3. Test pair selection

In general, pairs are chosen randomly. A set of pairs is selected from all points of the code space, which become swaps hypotheses and are tested for energies.

In practice, a much more efficient method is to choose the first point of a pair randomly and the second point within a certain radius of the first.

As the space gets consolidated, successful candidate points happen to appear nearby, so this selection method considerably improves overall performance.

### 5.4. Formal definition

Given a matrix $\mathbf{V}$, of dimension $m \times n$:

$$\mathbf{V} = \begin{pmatrix} v_{11} & v_{12} & \cdots & v_{1n} \\ v_{21} & v_{22} & \cdots & v_{2n} \\ \vdots & \vdots & \ddots & \vdots \\ v_{m1} & v_{m2} & \cdots & v_{mn} \end{pmatrix},$$

where each element $v_{ji}$ is either a bit vector of length $k$ or a *vector of features* $(f_1, f_2, \cdots, f_k) \in \mathbb{R}^k$, and for each pair of elements $a, b \in \mathbf{V}$ a similarity measure with threshold is given

[14] Subsequently, an optimisation (Section 5.9.4) was found bringing the algorithm closer in efficiency to UMAP.

[15] Working name, an abbreviation of the phrase Discrete Approximation of Manifold Projections. Also, it echoes the nature of the neocortex we were inspired by.

$$\text{sim}_\lambda(a, b) = \tau(\text{sim}(a, b)) \in \mathbb{R},$$

where sim is the similarity measure in the code space, $\tau$ is the threshold function, $\lambda$ is the threshold value, and $\eta$ is the slope coefficient of the sigmoid curve in the neighbourhood of the threshold:

$$\tau(x) = x \cdot \sigma(\eta \cdot (x - \lambda)), \quad \sigma(z) = \frac{1}{1 + e^{-z}}.$$

Depending on the nature of a code space, the function sim can be performed using either an analytical measure of the similarity of the original elements (if possible) or a measure based on the similarity of their codes.

The $\lambda$ threshold filters out the metric's noise and gradually shifts the focus towards strongly connected points as the layout progresses.

Given a matrix of pairs of dimension $p \times 4$:

$$\mathbf{P} = \begin{pmatrix} x_{11} & y_{11} & x_{12} & y_{12} \\ x_{21} & y_{21} & x_{22} & y_{22} \\ \vdots & \vdots & \vdots & \vdots \\ x_{p1} & y_{p1} & x_{p2} & y_{p2} \end{pmatrix},$$

where each line is two pairs of coordinates of points to be tested for exchange.

We calculate the energy matrix for cases when the points remain in their places and when they are exchanged.

## 5.5. Pair energy calculation

In general, the energy of a system is comprised of $n^2$ interaction energies of each point with all other points.

However, when we consider the *relative change* of the system energy when a single pair of points get swapped, we see that individual pair's contribution is negligibly small. Also, swapping of one pair of points does not affect the interaction energies of the remaining points with each other.

The energy of the test pair itself does not affect the result either, because neither the similarity nor the distance between the points changes, so the energy remains the same.

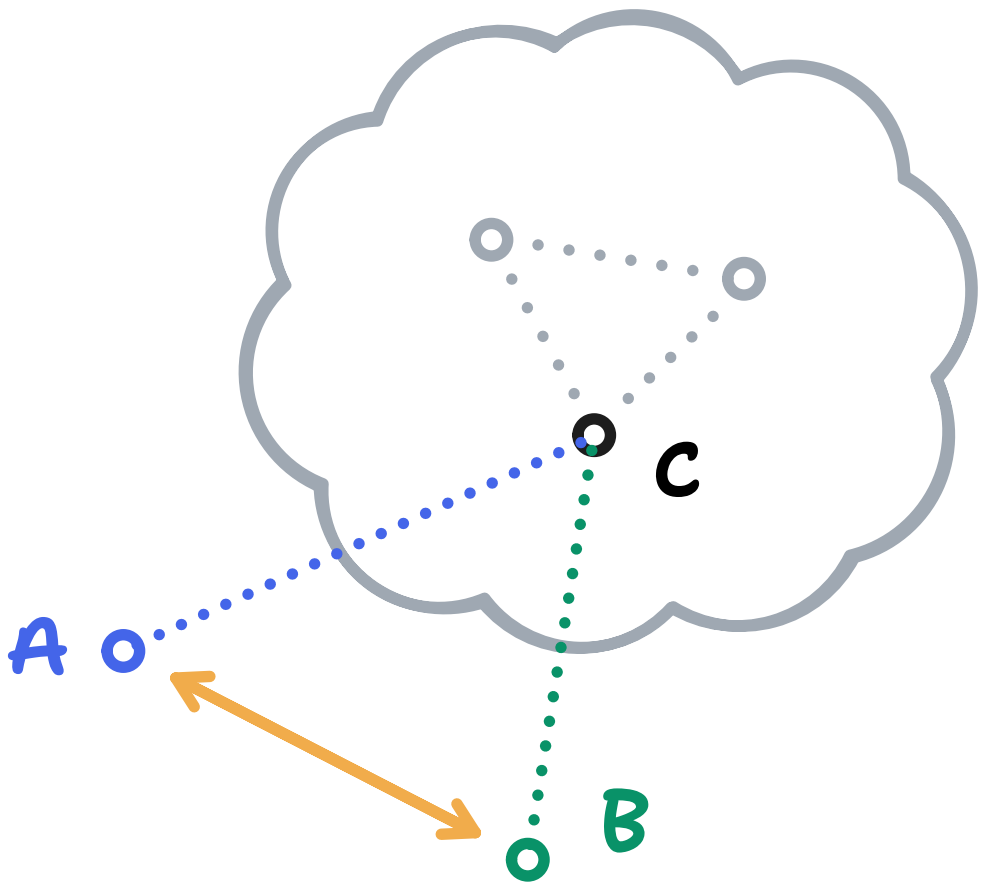

Figure 11: A test pair $A$, $B$ and a cloud of other points of the code space. The point $C$ interacts with $A$, $B$, and the different cloud points (not considered in the pair energy).

Assuming that the effect of an individual pair of points on the system's total energy is small, we can speculatively calculate the energies of many pairs, neglecting their interaction with each other.

We are interested in the relation energy between selected points and all other points in space (see Figure 11).

For each pair of points with coordinates[16] $(y_1, x_1, y_2, x_2)$, we determine the Euclidean distances to the selected point with coordinates $(y, x)$:

$$d_1 = \sqrt{(y_1 - y)^2 + (x_1 - x)^2},$$

$$d_2 = \sqrt{(y_2 - y)^2 + (x_2 - x)^2},$$

and then calculate the system's energy: $\varphi_c$ и $\varphi_s$.

### 5.5.1. Long-range layout

This algorithm considers the relations between points on long range and is therefore computed over the entire $\mathbf{V}$ space.

Similarity of points:

[16] We use the notation adopted in linear algebra libraries, in which the most frequently changing index is specified the last. For example, $a_{ji}$ and $\mathbf{A}_{ji}$ select $i$-th element in $j$-th row.

$$s_1 = \operatorname{sim}_\lambda\left(\mathbf{V}_{yx}, \mathbf{V}_{y_1 x_1}\right),$$
$$s_2 = \operatorname{sim}_\lambda\left(\mathbf{V}_{yx}, \mathbf{V}_{y_2 x_2}\right).$$

The pair energy when the points remain in their places:

$$\varphi_c = \sum_y \sum_x (s_1 \cdot d_1 + s_2 \cdot d_2).$$

The pair energy when the points are swapped:

$$\varphi_s = \sum_y \sum_x (s_2 \cdot d_1 + s_1 \cdot d_2).$$

The long-range layout goal is to **minimise** global energy of the system.

The product of similarity and distance has the effect that the system “penalises” strongly correlated points that are *distant* from each other, and forces them to move towards each other.

#### 5.5.2. Short-range layout

This is a fast variant of the algorithm that runs in the local neighbourhood $\mathbf{R} \subseteq \mathbf{V}$ of a test pair:

$$c_y = \frac{y_1 + y_2}{2},\ c_x = \frac{x_1 + x_2}{2},$$

$$d_c = \sqrt{\left(y - c_y\right)^2 + \left(x - c_x\right)^2},$$

$$r \geq \sqrt{\left(y_1 - y_2\right)^2 + \left(x_1 - x_2\right)^2},$$

$$\mathbf{R} = \left\{v_{yx} \in \mathbf{V} : d_c \leq r\right\}.$$

Here $r$ is the circle’s radius with the centre at the middle of the segment connecting the points of the test pair. The larger $r$ is, the more accurate the layout will be.

Since the algorithm does not consider the relations between points over long distances, it makes sense to apply it only for “polishing” the short-range order when the long-range order has already been established and the distance between swapping points is small.

Similarity of points, now by $\mathbf{R}$:

$$s_1 = \operatorname{sim}_\lambda\left(\mathbf{R}_{yx}, \mathbf{R}_{y_1 x_1}\right),$$
$$s_2 = \operatorname{sim}_\lambda\left(\mathbf{R}_{yx}, \mathbf{R}_{y_2 x_2}\right).$$

The pair energy when the points remain in their places:

$$\varphi_c = \sum_y \sum_x \left(\frac{s_1}{d_1} + \frac{s_2}{d_2}\right).$$

The pair energy when the points are swapped:

$$\varphi_s = \sum_y \sum_x \left(\frac{s_2}{d_1} + \frac{s_1}{d_2}\right).$$

The short-range layout goal is to **maximise** local energy of the system.

Unlike the long-range algorithm, here we do not penalise distant correlated points, but instead encourage *nearby* ones.

#### 5.5.3. Energies

The result is a $p \times 2$ matrix of energies:

$$\mathbf{E} = \begin{pmatrix} \varphi_{1c} & \varphi_{1s} \\ \varphi_{2c} & \varphi_{2s} \\ \vdots & \vdots \\ \varphi_{pc} & \varphi_{ps} \end{pmatrix}.$$

The two columns correspond to the energies before and after the swap, respectively.

### 5.6. Point exchange and layout

A pair of points is swapped when the energy $\varphi_s$ turns out to be favourable. Otherwise, points remain in their places.

Each step of the layout algorithm results in the exchange of a certain subset of pairs. The steps are repeated until the number of exchanges per step falls below a threshold or until the layout quality function reaches a certain value.

### 5.7. Point energy

For each point $c \in \mathbf{V}$ with coordinates $\left(c_y, c_x\right)$ we define a neighbourhood $\mathbf{R} \subseteq \mathbf{V}$ of radius $r$:

$$d_c = \sqrt{\left(y - c_y\right)^2 + \left(x - c_x\right)^2},$$
$$\mathbf{R}(c, r) = \left\{v_{yx} \in \mathbf{V} : d_c \leq r\right\},$$

the point energy with a threshold $\lambda$:

$$E(c, r) = \sum_y \sum_x \frac{\operatorname{sim}_\lambda\left(c, \mathbf{R}(c, r)_{yx}\right)}{d_c},$$

the normalised point energy:

$$\hat{E}(c,r) = \frac{E(c,r)}{E_{max}}, \quad E_{max} = \max_{v \in \mathbf{V}} E(v).$$

In some cases, it makes sense to compute $E_{\max}$ over a neighbourhood of $\mathbf{R}$ rather than over the whole space $\mathbf{V}$.

The energy of a point should be taken as a *measure of relevance* or *fitness* of a point to its environment. The more compact and homogeneous the environment is, the higher the overall energy of its points.

To visualise the layout process, an energy matrix is used, obtained by calculating the normalised energy for each point in the code space within a radius $r_e$:

$$\hat{\mathbf{E}}_{ji} = \hat{E}(c_{ji}, r_e), \quad \forall c_{ji} \in \mathbf{V}.$$

In most cases, $r_e \leq 5$ is sufficient.

## 5.8. Layout quality assessment

The average energy of the space can be used to estimate the global quality of the layout:

$$\bar{E} = \frac{1}{|\mathbf{V}^+|} \sum_y \sum_x \hat{E}(\mathbf{V}_{yx}, r_e).$$

Here $|\mathbf{V}^+|$ denotes the number of *nonzero* elements in the matrix:

$$\mathbf{V}^+ = \{v \in \mathbf{V} : v \neq 0\}.$$

## 5.9. Algorithm optimisations

In the most general case, only one pair of points ($p = 1$) is evaluated per algorithm step.

However, as shown before, if we neglect the interaction of individual pairs with each other, we can significantly improve performance by calculating energy and making substitutions speculatively and in parallel, and still not lose in accuracy.

### 5.9.1. Pair selection and early cut-off

It was noted above that the second point in a pair should be chosen within some radius of the first, because as the space is laid out, the successful candidate points tend to appear nearby. This narrows the random search area and reduces the number of unsuccessful pairs.

In some tasks, the early cutoff for pair selection has proved useful. Namely, when selecting a pair of points $a, b \in \mathbf{V}$, the condition $\text{sim}(a,b) \geq t$ can be used to filter weakly correlated points.

Such an optimisation can be helpful at later stages of the layout, when similar points must be moved. In this way, it is possible to avoid calculating the energy of obviously unsuccessful pairs.

If the code space contains zero points, only one of the points should be zero when selecting pairs.

### 5.9.2. Energy calculation

When calculating distances between the points, we can keep the values as sums of squares and not extract the roots, since we care about the differences of energies, not their absolute values.

### 5.9.3. Probabilistic first point selection

As the code space is laid out, we can periodically calculate the energies of all points and use this information to select points with probabilities proportional to their energies during the layout.

The probability of choosing a point $p$ is determined by its normalised energy:

$$P(p) \propto -\ln \hat{E}(p, r).$$

As long as the code space is weakly organised, the energies of points are close to zero. This corresponds to the state where the probability of choosing any particular point is relatively equal.

As the space gets laid out, more clusters are gradually formed, so the point energies begin to increase as well.

Thus, the selection probability will be shifted towards weakly organised parts of the code space.

### 5.9.4. Calculation on a subset

UMAP is fast because it considers only $k$ nearest neighbours [2, p. 3.1] when computing the gradient at a point.

We can imagine a variant of the long-range algorithm that would make a pass over subset of $\mathbf{V}$.

In particular, using the space activation map as a point selection mask is promising.

#### 5.9.5. Similarity matrix

For long feature vectors, it makes sense to precompute a similarity matrix, the values of which are used in the layout. This would be benificial if the memory overhead is less than the overhead of recomputing the similarity in place.

## 5.10. Parallel and distributed processing

The layout algorithm is based on the principle of speculative calculation of point pair energies. Since the calculation of pair energies, both in the short-range and long-range variants, depends only on $\mathbf{V}$, each pair can be calculated independently and in parallel. The same applies to the distributed computation over many nodes.

In addition, in the distributed variant, it is possible to speculatively continue the calculation of new pairs even if the exchange lists from other nodes have not yet been applied. This can avoid idle time and even out the energy consumption spikes.

All this is possible because the probability of simultaneously selecting the same point on multiple nodes is relatively small, and the effect of an individual substitution on the total system energy is negligible.

Therefore, in the case where the number of swaps is much smaller than the number of points in the space, speculative processing cannot significantly affect the result.

Even in the worst case, when the same point appears several times in the swap list, it is possible either to apply only one of the swaps, or to apply all swaps and save resources for uniqueness checking, at the cost of a small number of erroneous swaps, which will be corrected later.

If we synchronise pair generation so that within the entire cluster, each point only enters one pair per period, there will be no issues with swap conflicts at all.

## 5.11. Asymptotic complexity

To calculate the energies of a single pair, we must *linearly* traverse $m \times n$ nonzero elements of the matrix $\mathbf{V}$ once.

Since the energies of individual pairs are independent of each other, all of them can be computed simultaneously in $O(m \times n \times p)$ operations. There are $p$ pairs in total, so the exchange of points is done in $O(p)$ operations.

Thus, the asymptotic complexity *grows linearly* and depends on the size of the code space and the number of pairs.

Given that typically $m \times n \gg p$, the asymptotic complexity of a single layout step can be estimated as

$$O(m \times n \times p) + O(p) \approx O(m \times n).$$

## 5.12. Accelerated GPU implementation

Due to its nature and practical lack of data dependencies, the layout algorithm is well-suited for GPU computation.

Since $\mathbf{V}$ can be very large (millions of elements) and the number of pairs in $\mathbf{P}$ is in the thousands, an efficient implementation should use $\mathbf{V}$ only once[17].

To efficiently utilise GPU caches, it is important to group the computation so that data would be processed locally whenever possible.

Each point from $\mathbf{V}$ contributes to energy of each pair from $\mathbf{P}$. Therefore, we have to either accumulate the results in parallel in the output tensor (that would require memory synchronisation), or to use additional memory per batch, which would later be reduced to get the final result.

The best performing implementation is one in which points are placed in shared workgroup memory.

Details and implementation aspects:

- Each thread computes **only one pair** $p \in \mathbf{P}$.
- Each thread operates on a subset of $\mathbf{B} \subseteq \mathbf{V}$.

[17] This section describes an accelerated implementation without optimisations mentioned in Section 5.9.4.

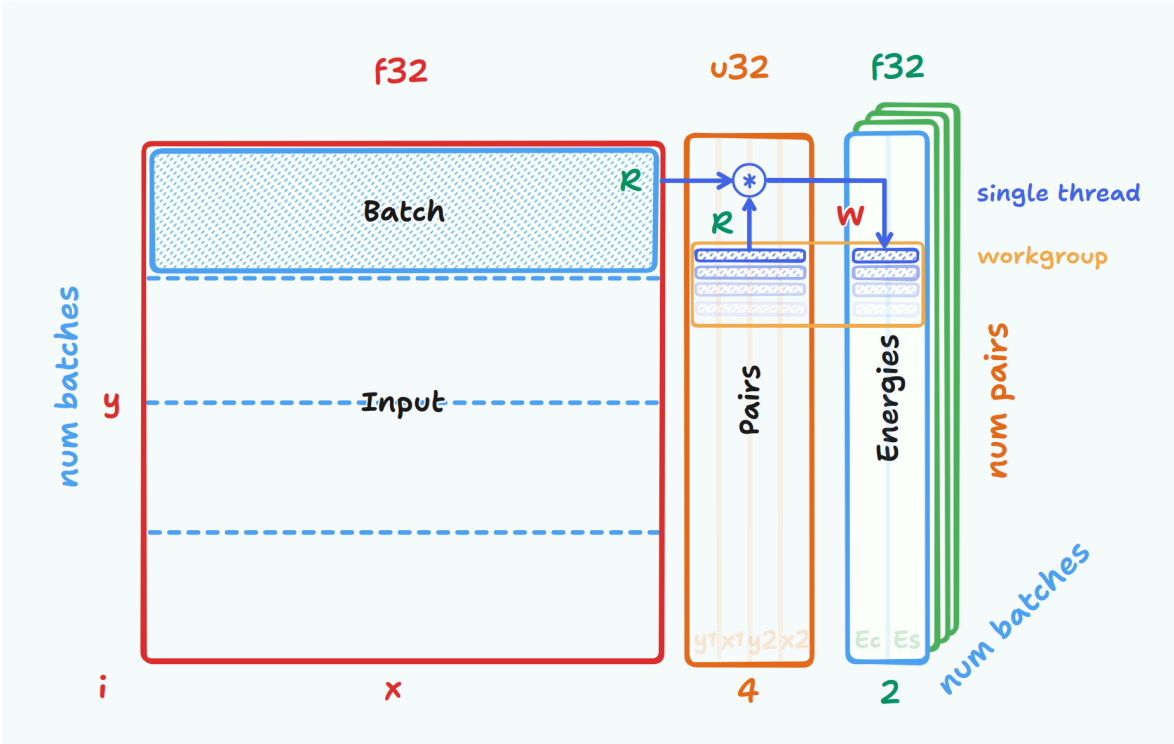

Figure 12: A scheme for accelerated energies calculation.

- At startup, a thread reads the codes or vectors corresponding to the points of its pair from $\mathbf{V}$ and stores them in the workgroup's shared memory as a cache.
- Each thread reads only constants from $\mathbf{B}$ and $\mathbf{P}$, accumulates the sums of the energies in local memory, and writes a single pair of values to $\mathbf{E}_k$.
- The results of the batches are summed: $\sum \mathbf{E}_k$.
- Many threads simultaneously read a limited chunk of $\mathbf{V}$, which should efficiently utilise the cache and memory coalescence.
- By varying the size of the batches, from a single row to the size of a matrix, an optimal variant for the GPU architecture can be found.

## 5.13. Code requirements

Several conditions must be fulfilled for a correct code space layout. If conditions are not met, the layout and, in turn, detector codes may produce inadequate results.

Below are examples of successful and unsuccessful code space layouts on a synthetic problem of two-dimensional gradient codes layout.

### 5.13.1. Importance of long-range order

For an appropriate layout of the code space, codes must have similarity over a wide range, especially if the space is topologically closed in one or more dimensions.

The Figure 13.a shows an example of an autistic layout of gradient codes. The colours in the image encode the value of component $x$ of a two-dimensional linear gradient, where each point $(x, y)$ corresponds to a unique combination of components $x, y \in \{0, 1, \ldots, 99\}$. Red corresponds to points where $x = 0$, and purple corresponds to points where $x = 99$.

Ideally the layout should result in a smooth transition from red to purple, sequentially through all the colours of the spectrum.

We can see that the local structure was somewhat formed, but it is discontinious, and the long-range order is significantly distorted. The layout appears to have several "crystallisation centres", which "compete" for the attention of other points.

An attempt to build a detector space (Chapter 6) over such a code space would result in a situation where unrelated concepts could end up in the same detector and therefore, would get common bits. Vice versa, some conceptually close points would get distant codes, because they were not laid out properly.

The Figure 13.b shows a pathological case of a two-dimensional gradient code layout with closed topology in which the long-range order is completely broken. Laying out and semantically interpreting such a space is practically impossible.

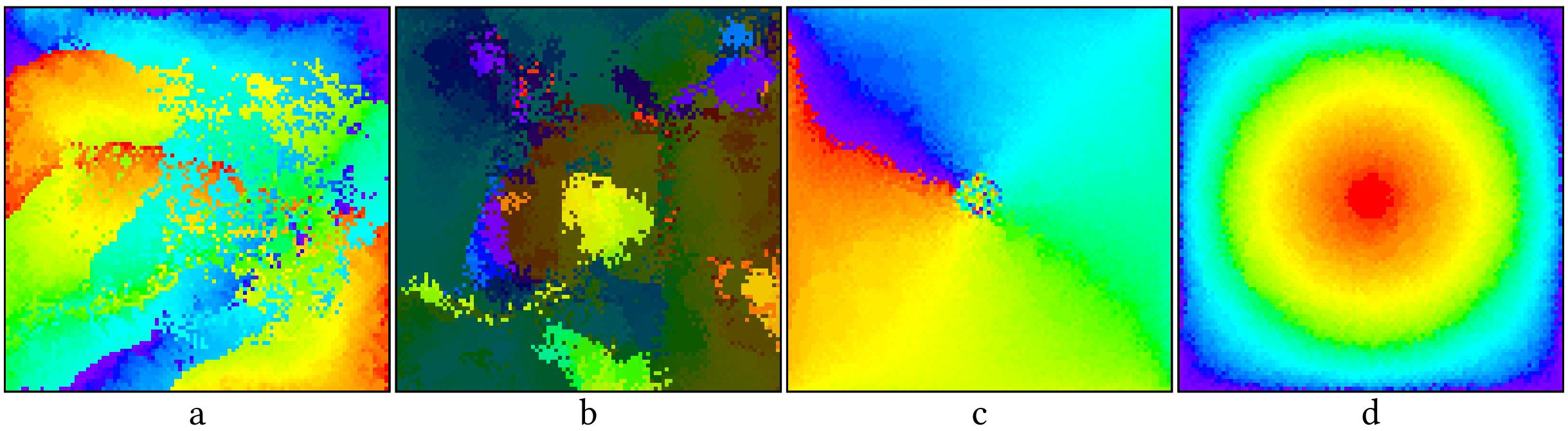

Figure 13: Examples of layout distortion. **a, b**: the long-range order is broken; correlated sections are highlighted in **b**; **c, d**: unsuccessful layout of gradient angle and modulus components, respectively.

Several regions are highlighted according to the similarity of a point to the centre of corresponding area. The value of point similarity is mapped to the brightness component of the HSV colour model [52].

In both cases, the problem was not in the layout mechanism, but in the code space itself. Insufficient overlap of codes of separate points led to a shift of pair energies towards shot-range order.

Thus, long-range order (Section 4.5) is essential when performing long-range layout to organise the code space according to domain topology.

### 5.13.2. Mapping continuity

In addition to similarity, the smoothness of the code space organisation is also essential.

The chapter on wide detectors gave an example of code space generation by cylindrical surface (Section 4.4.1). Such codes will be topologically closed and have good modulus similarity. However, this is not enough to realise a smooth layout, since a uniform mesh applied to a cylindrical surface cannot be projected onto a plane without rips and wrinkles.

#### 5.13.2.1. Analytical metric

The analytical approach gives good results, but choosing the right similarity metric and topology is essential.

Figures 13 **c** and **d** show layouts of the components of a two-dimensional (conical) gradient. The first image shows the gradient angle, and the second shows its modulus.

The Cartesian distance between the points representing the gradient components in the polar coordinate system was chosen as the metric. In other words, similarity is expressed by the difference of gradient vectors:

$$\mathrm{sim}\left(\vec{a},\vec{b}\right) \propto |\vec{a}-\vec{b}|.$$

It can be seen that the angle component is laid out incorrectly in the vicinity of zero. This is because the points have parasitic similarity near the origin. Despite its symmetry, the modulus component is distorted in the vicinity of maximum values (purple), which are clumped together at the corners of the map. However, they should have formed an outer concentric ring instead.

Figures 14 **a** and **b** show a better layout using similar analytic metric. However, the gradient with zero modulus corresponds to a vector with some minimum length, sufficient to make the gradients considered different in angle, even if their moduli are close to zero.

The layout is quite smooth, except for folds caused by the need to fit the space into a square. The codes would gather a smooth and rounded pinwheel if a larger matrix is chosen as the code space.

Therefore, for a smooth layout, the cyclic similarity of codes and the distribution of points are essential.

To implement an topology without using an analytic metric, the code space must be organised so that its topology is as similar as possible to equivalent analytic version.

#### 5.13.2.2. Layout by code

Figures 14 **c** and **d** show an example of an unsuccessful attempt to lay out a code space that repeats the analytic variant described earlier.

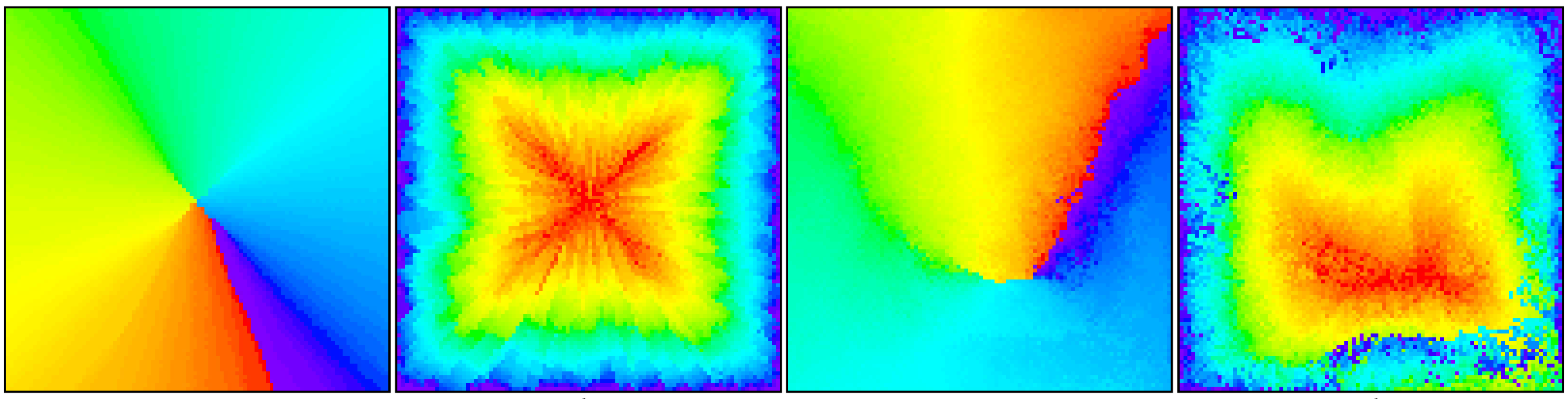

a b c d

Figure 14: Examples of conic gradient layout. **a, b**: successful layout by analytic metric; **c, d**: unsuccessful layout by codes with topology similar to the analytic one.

The topology seems right but still has significant long-range order distortions. This is due to the insufficient overlapping of the codes of neighbouring points, which allowed arranging only the shot-range order.

#### 5.13.2.3. Adjusting encoding parameters

The Figure 15 shows a successful variant of code space layout, giving long-range order and moderate code accuracy. The space was properly laid out at the corners into a more or less symmetric pinwheel.

The space has good overlap and low density (33 bits out of 128), but with a relatively low resolution of 2.54. Out of 10 000 of angle and gradient combinations, the space could only encode 3 926. In the worst case, 43 points are mapped to the same bit vector.

On the left is the laid out code space, on the right is the generation space of the primary detectors, and the detectors are active for a given combination of angle and modulus (yellow dots).

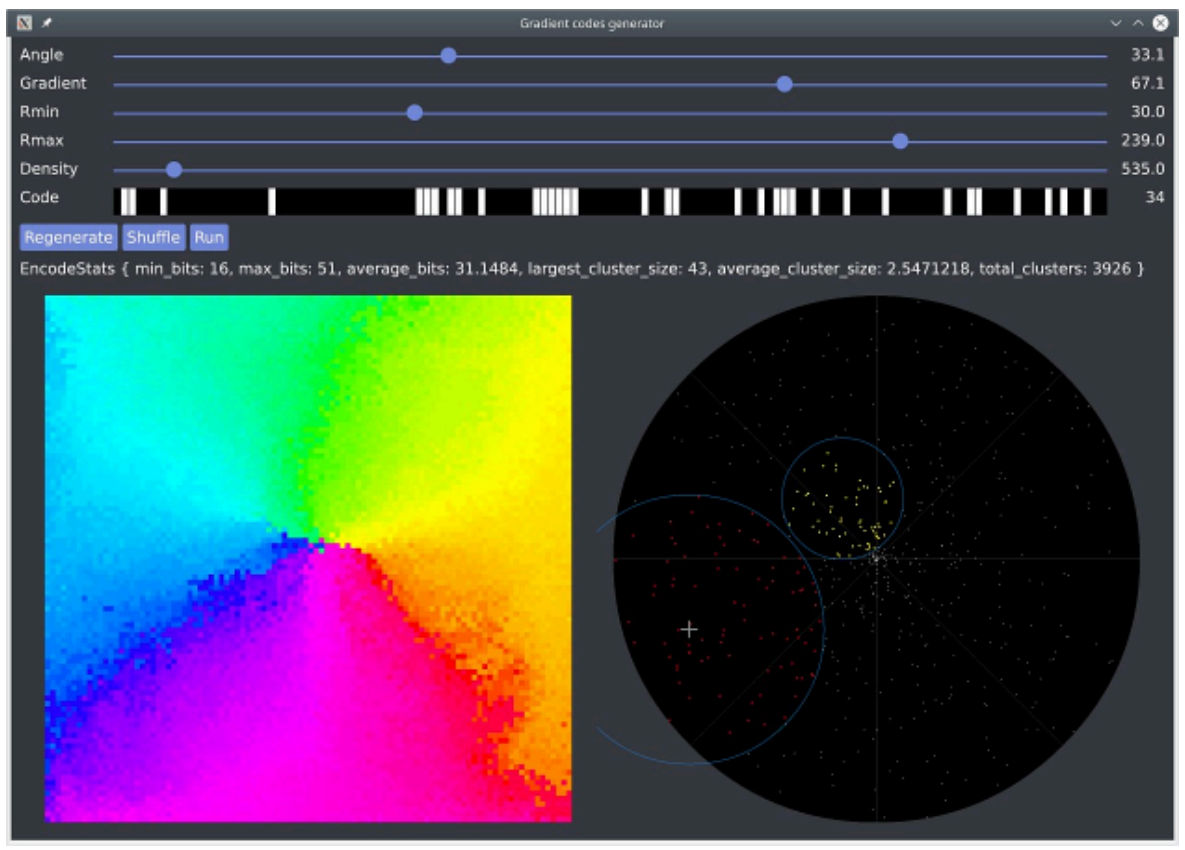

Figure 15: Parameter selection interface with an example of a somewhat successful layout.

#### 5.13.2.4. Dynamic code modification

Fully redesigned encoding variant, which directly sets the overlap in angle and modulus, in the current settings gives the highest resolution at the cost of low overlap.

In the experiment on Figure 16.b, encoding parameters were changed as the layout progressed. Initially, codes with maximum overlap and low resolution were used; as the space was laid out, the overlap decreased, shifting the emphasis to short-range order.

#### 5.13.2.5. Colour coding

Colour coding (Section 3.2) allows the space to be laid out in single pass, without changing the codes as the layout progresses.

The Figure 16.c shows a space with well-tuned colour codes. Angle overlap 170°, modulus overlap 30/100, total resolution per code $10\,000/5\,622 \approx 1.78$, largest cluster size 11.

A fragment of the same code space is highlighted on Figure 16.d. The brightness component of the points is proportional to the similarity of point codes to the selected one. It can be seen that conceptually similar points have their codes consolidated in a dense cluster.

### 5.13.3. On the use of neural network embeddings

Neural network embeddings usually have many dimensions (hundreds, thousands), much larger than typical length of our vectors (tens of ele-

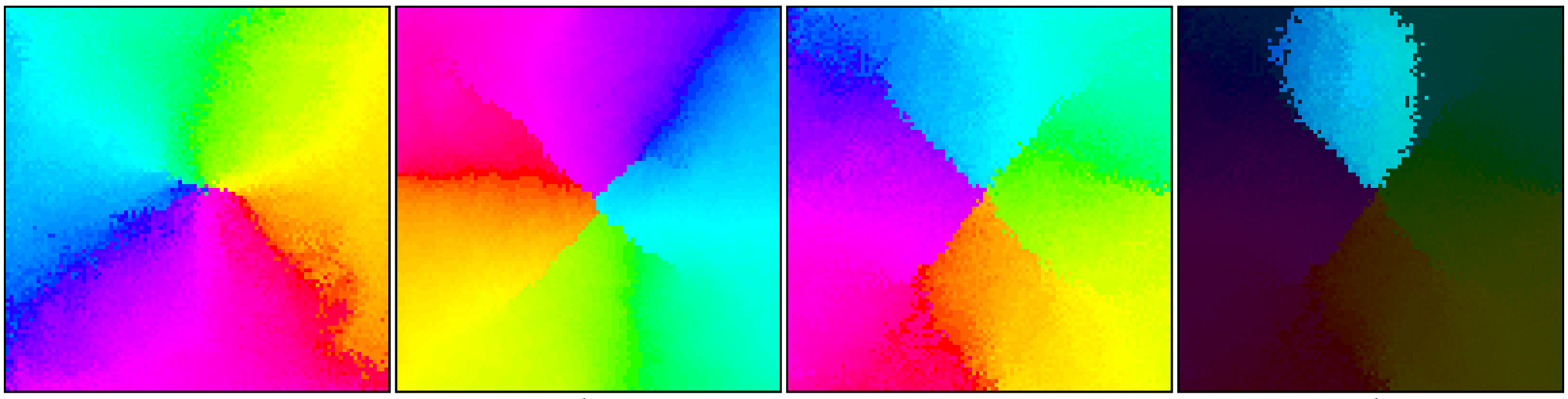

a b c d

Figure 16: Examples of successful gradient layouts (corner component shown).
**a**: a copy of the low-resolution space shown earlier on Figure 15.
**b**: layout by changing codes. **c**: colour code layout in one pass.
**d**: visualisation of the space activation from variant **c**, by cosine metric with $\lambda = 0.6$.

ments). In addition, their topology can be very complex and non-planarizable.

For this reason, it seems questionable to directly use neural network embeddings as input codes for the layout.

Nevertheless, since existing nonlinear dimensionality reduction algorithms, such as t-SNE and UMAP, can deal with them and construct a relatively smooth map, our algorithm theoretically could manage as well.

#### 5.13.4. Conclusions

In order to get a proper layout, the code space's topology must match the original domain's topology as closely as possible while still being suitable for a smooth mapping to the plane. Therefore, it is important not only to have codes with similarity, but also have a decent amount of overlap in them.

A similar picture can be observed in neurophysiology. In particular, the map of orientation sensitivity of mini-columns of the visual cortex [4, fig. 7] closely resembles our maps. This gives grounds to speak about the potential similarity of the informational nature of the ongoing processes.

### 5.14. Laid out space structure

Let's take another look at the code space layout process (Figure 17).

First, an unorganised set of points was randomly scattered over the coding space (column **a**).

The colour of a point in space is calculated as the colour sum of the unit bits of the vector representation, where the least significant bit corresponds to red and the most significant bit to violet.

The square in the centre is an artefact that does not affect the subsequent layout. The dots were added gradually; as the space was filled, it was enlarged, giving this effect.

During the process of far layout (column **b**), the points were reorganised into groups. Point energy maxima correspond to clusters of points of close colours.

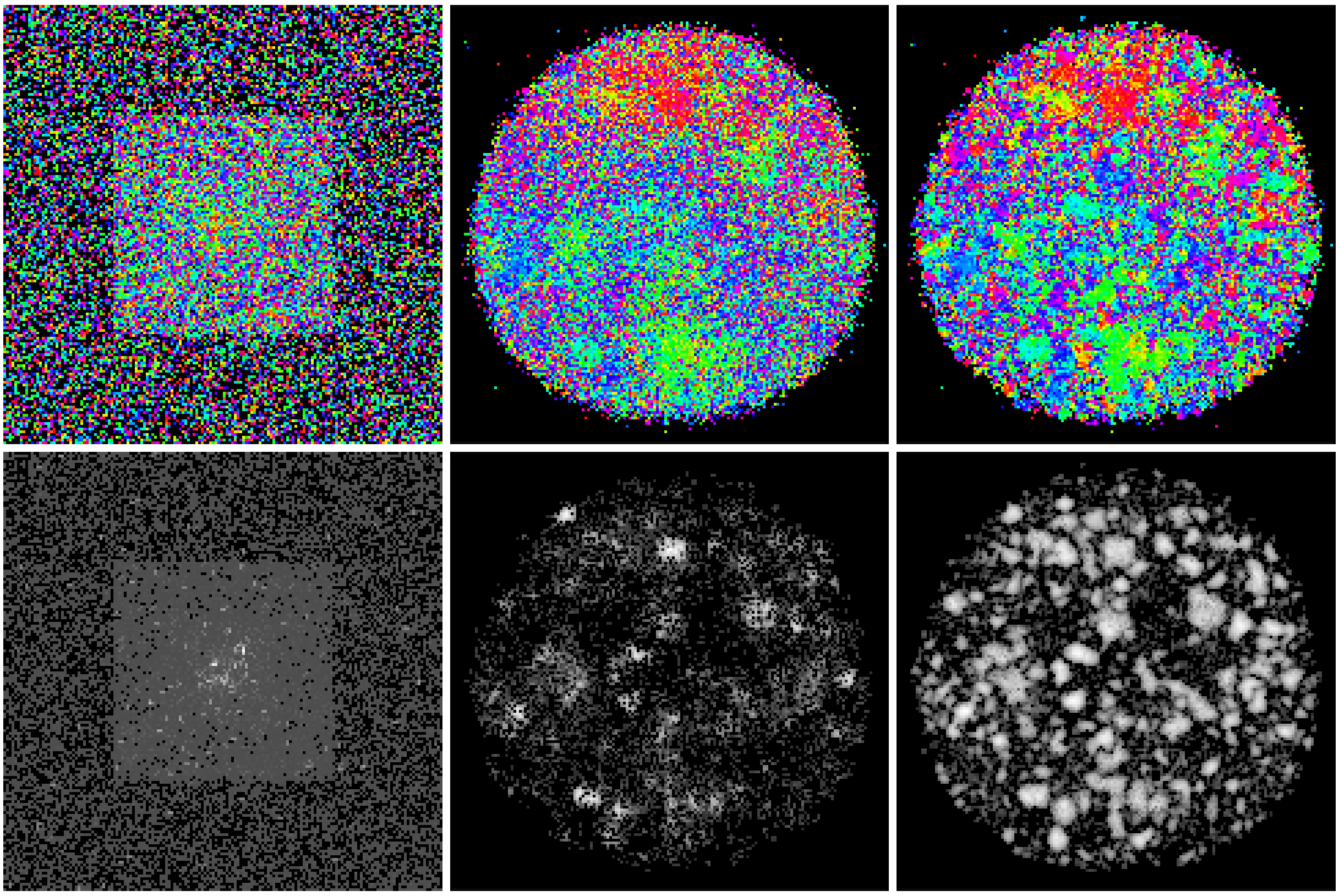

a b c

Figure 17: Code space layout process. The code space $\mathbf{V}$ and its corresponding energy matrix $\hat{\mathbf{E}}$ are shown. **a**: the beginning of the layout, **b**: the middle of the process, **c**: the final state.

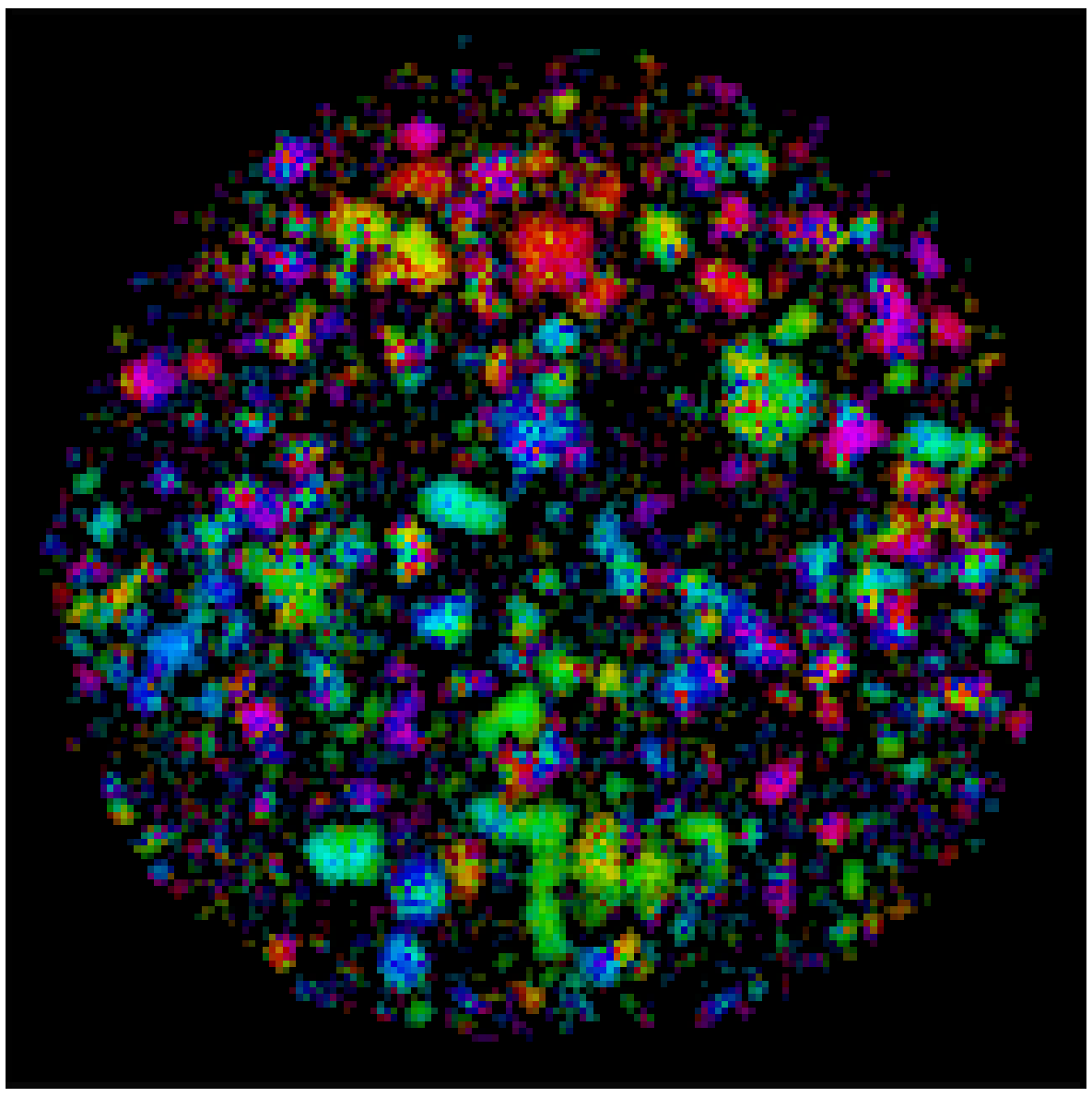

Figure 18: Composite visualisation of the laid out space. The colour of a point is defined by the code $\mathbf{V}$, its brightness by the energy $\hat{\mathbf{E}}$.

After switching to the near layout algorithm (column **c**), the space quickly evolved, the point energyies increased rapidly, and many compact regions with distinct boundaries appeared.

A good way to visualise the code space is to project the matrices $\mathbf{V}$ and $\hat{\mathbf{E}}$ onto the same map, so that a bit code gives the pseudo-colour of a point and its brightness determines its energy (Figure 18). Such an approach helps to highlight well-organised groups and hides unorganised "rubbish".

The following regular elements of the structure can be identified:

- *Clusters and pinwheels* are dense regions where the similarity between elements is greater than with elements of the environment. Unlike clusters, pinwheels have a radial, often cyclically closed substructure reflecting the local topology of a space.
- A *hyper cluster* is a cluster of individual clusters or pinwheels. The elements of such a cluster have similarity, but it was not enough to merge the whole set of points into a single dense cluster.
- *Bridges* or *threads*, usually look like thin lines spanning from one part of a space to another. They can be a sign of not fully laid out space, problems in codes, violation of long-range order or presence of strong connections of elements that is hard to express in 2D. In a well laid out space, the number of bridges is minimal.
- Unorganised areas with low energy, resembling "colourful static". Typically, these are "rubbish" codes that have not found their place and have no pronounced similarity to other points, except for random bit collisions.

The structure of the laid out code space is essential because it allows us to describe the subject domain in domain-specific codes.

## 6. Detection

In Chapter 5, we mentioned that the idea and method of layout are based on the manifold hypothesis, stating that a multidimensional dataset can often contain a nested subspace of lower dimensionality.

We can construct a mapping of the original space into the nested space and describe it using a compact system of codes. This turns out to be more efficient than the description in terms of the original space of higher dimensions.

A detector space constructed over a code space represents such a mapping.

The detectors described in this chapter are not fundamentally different from those from Chapter 4. Here we describe detectors that are formed based on the characteristics of underlying code space (Section 5.14) and, therefore, can be distributed non-uniformly.

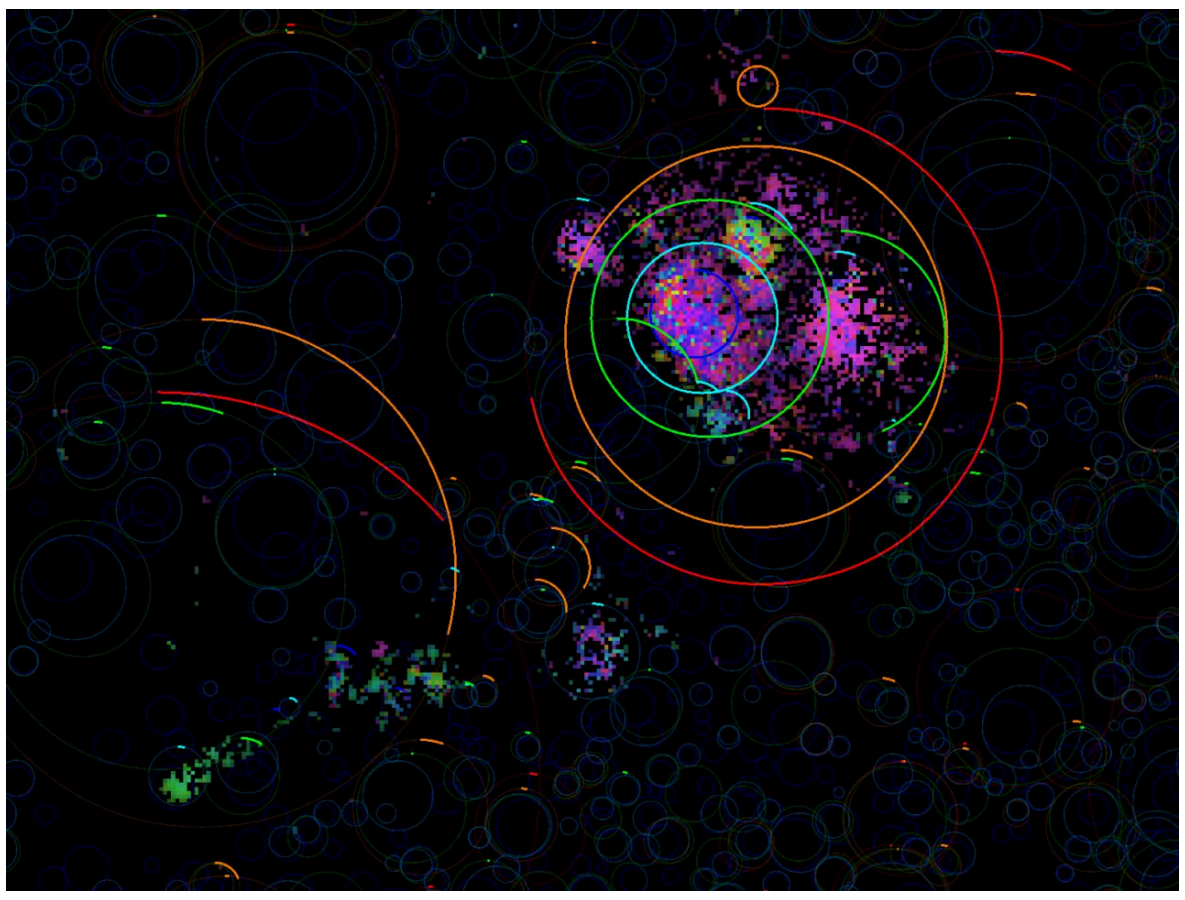

Figure 19: Activated detectors.

The Figure 19 shows an activated fragment of a morphology space (Section 7.1) with $\lambda_a = 0.6$, a detector space (background circles) a the set of activated detectors (bright arcs and circles). An arc length is proportional to a detector activation level, its colour corresponds to a detector threshold: red $\lambda_d = 0.5$, blue $\lambda_d = 0.75$.

## 6.1. Code space activation

For each point of a code space $\mathbf{V}$ we apply a similarity function with a threshold $\lambda$:

$$a_{ji} = \mathrm{sim}_\lambda(c, v_{ji}),$$
$$\forall j, i : a_{ji} \in \mathbf{A}_\lambda, v_{ji} \in \mathbf{V}.$$

The resulting matrix $\mathbf{A}_\lambda(c)$ represents the activation of the space $\mathbf{V}$ by the code $c \in \mathcal{V}$. Generally speaking, the code $c$ can be anything and does not need to be taken from the code space itself.

When constructing the detector space, instead of calculating the whole matrix, a local fragment is used, analogous to the point-energy calculation (Section 5.7):

$$a_{ji} = \mathrm{sim}_\lambda(c, v_{ji}),$$
$$\forall j, i : a_{ji} \in \mathbf{A}_\lambda, v_{ji} \in \mathbf{R}(c, r).$$

In other words, for a chosen center point $c \in \mathbf{V}$, we apply the similarity function to every point $v$ in the code space that is within radius $r$ of $c$, and we put the results into the activation matrix $\mathbf{A}_\lambda(c, r)$ at the corresponding coordinates.

## 6.2. Detector hierarchy and embeddings

When a code space is activated by an input, only part of the space has an activation energy above the $\lambda$ threshold. Also, because of the layout, the activation preserves local structure of the code space.

The Figure 20 shows a fragment of a morphological code space, its corresponding detector space, and their activation at two points. Note, that the hierarchy of active detectors quite accurately describes the location of activated regions of the code space.

Based on the assumption that the activation pattern is unique enough and behave like a "fingerprint" of the stimulus, it can be used as an embedding prototype. Our experiments (Chapter 7) make us believe the hypothesis is valid, but we do not provide formal proof yet.

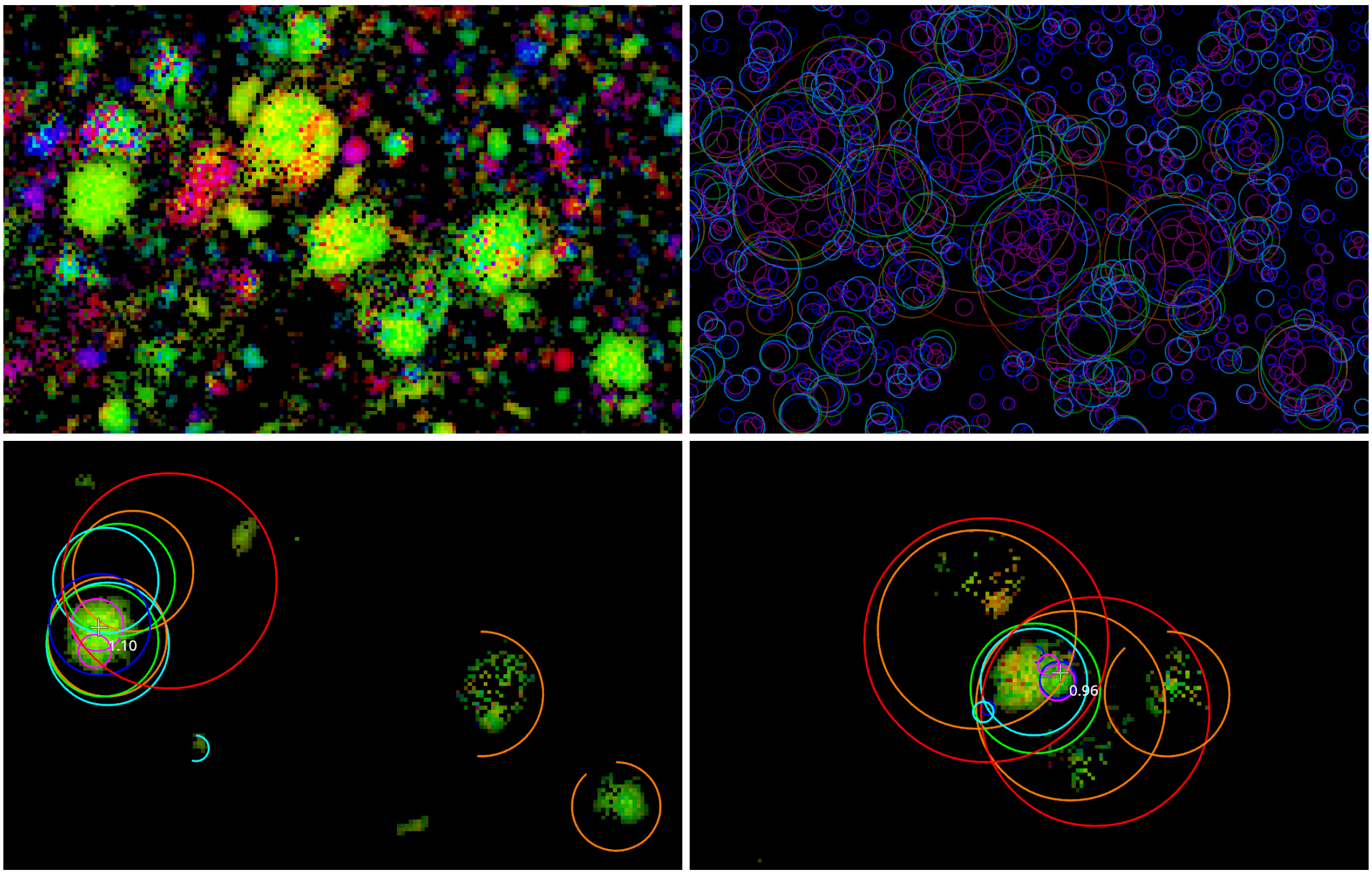

Figure 20: Code space, detector space, and their activation.

If we describe the activation pattern in terms of active detectors, the code they form will also carry the pattern's "imprint" and have similar properties.

In particular, similar activation patterns will correspond to identical sets of active detectors, and therefore, their codes will also be similar. All points within the violet detector (the smallest circles) in the example above will give nearly identical codes.

By doing this, we also preserve the original topology in the derived detector code space. Codes that inherit the stimuli topology can be combined, transformed, and laid out at the next level of the hierarchy.

## 6.3. Detector parameters

Each detector has several parameters that determine how it was created, when it should activate, and how it affects the resulting embedding code.

These parameters include:

- Detector center $c_d \in \mathbf{V}$ with coordinates $(c_y, c_x) \in \mathbb{R}^2$ or $\mathbb{Z}^2$.
- The radius of the receptive field $r_d \in \mathbb{R}$.
- The value of the activation threshold $\lambda_d \in \mathbb{R}$ used during the detector creation.
- The number of points $n_d$ with energy higher than the minimum $\mu$ that were in the *receptive field* of the detector at the moment of its creation:

$$n_d = |A_\mu|,$$
$$A_\mu = \left\{a_{ji} \in \mathbf{A}_{\lambda_d}(c_d, r_d) : e_{ji} \geq \mu\right\}.$$

  Here $a_{ji}$ is the level of threshold activation of points in code space by the code from the center of the detector, and $e_{ji} \in \hat{\mathbf{E}}$ is normalised energy of the corresponding point in space taken at the same coordinates.
- Detector total energy:

$$e_d = \sum a_{ji} \cdot e_{ji}, \quad \forall j, i : a_{ji} \in A_\mu,\ e_{ji} \in \hat{\mathbf{E}}.$$

- Detector activity output code $b_d \in \mathcal{C}$, usually a vector with random one of its bits set.

These parameters are set once when creating the detector and usually are not changed afterwards.

## 6.4. Construction of detector hierarchy

The construction is performed stochastically. All detectors are organised into several *layers*. The number of layers defines the depth of the *detector hierarchy*, directly affecting the number of set bits in the output code during its activation.

First, for each *layer*, a certain activation threshold $\lambda$ is fixed, which defines $\lambda_d = \lambda$ for all detectors created on this layer.

For each new *detector*, a random point $\tilde{c} \in \mathbf{V}$ is selected and interpreted as an *approximate* centre of said detector. Once the centre is selected, point clustering and optimal detector radius calculation are performed. Then the pending detector is tested against all existing detectors in its region and inserted to the hierarchy if proven to be suitable. This is repeated, until the whole layer is filled with detectors and no new detectors can replace existing ones.

### 6.4.1. Point clustering

A code space is activated $\mathbf{A}_{\lambda_d}(\tilde{c}, r_a)$ by a code from $\tilde{c}$ within the activation radius $r_a \geq r_d$ and a threshold $\lambda_d$. Activated points are then grouped into separate clusters (or pinwheels) $P \subseteq \mathcal{P}$ by one of the clustering algorithms.

DBSCAN [53], [54] is well-suited as a clustering algorithm.

The $k$-means method [55], [56] is undesirable because the number of classes is unknown in advance and can vary greatly depending on the code space's topology and the chosen point.

At the same time, clusters (pinwheels) of the code space are ideal for density-based clustering.

### 6.4.2. Cluster centre calculation

For each cluster (pinwheel) $P$, its centroid is calculated as a weighted average of coordinates of cluster's points. Here, $p_i \in P$ are points, and $w_i \in W$ are their *weights*, defined as a product of the point's activation and its energy:

$$\vec{c}_d = \frac{\sum \vec{p}_i \cdot w_i}{\sum w_i}, \quad W = \mathbf{A}_{\lambda_d}(\tilde{c}, r_a) \cdot \hat{\mathbf{E}}.$$

### 6.4.3. Optimal detector radius

The goal is to obtain a detector that surrounds the cluster (pinwheel) as tightly as possible. In reality, a cluster boundary is reasonably close to a circle, but may also include some points scattered at a distance around dense "core". Therefore, we need to filter out the outliers without negatively affecting the detector accuracy.

The distance from a point $p$ to the centre $c_d$ is

$$r(p) = \|\vec{p} - \vec{c}_d\|.$$

The ratio of the number of points within the radius (as an approximation of the area they occupy) to the ideal occupancy for the given radius (the circle area) is

$$f(p) = \frac{|\{q \in P : r(q) \leq r(p)\}|}{\pi \cdot r(p)^2}.$$

The search for the optimal radius of a detector comes down to finding a point $p \in P$ for which $f(p)$ is maximal:

$$r_d = r\left(\operatorname*{argmax}_{p \in P} f(p)\right).$$

### 6.4.4. Statistical methods

The circle method described above gives acceptable results with minimum cost for simple cases and clusters with about zero eccentricity.

In complex cases, the principal component analysis [49] and the Mahalanobis statistical distance [57] can be used to calculate the parameters of a circumscribed ellipse that will define the boundary of the detector's receptive field.

### 6.4.5. Detector insertion

Based on $\lambda_d$, $c_d$ and $r_d$, the remaining detector parameters such as $n_d$ and $e_d$ are calculated, and a random output code $b_d$ is generated.

When inserting a new detector $d \in \mathcal{D}$ into the detector hierarchy $D \subset \mathcal{D}$, it is crucial to ensure that it does not overlap the centres of existing detectors on the same layer of the hierarchy:

$$D_\lambda = \{e \in D : \lambda_e = \lambda\}$$

and that existing detectors from $D_\lambda$ do not overlap its centre $c_d$.

That is, the distance between $c_d$ and a centre of any existing detector $c_e$ must be greater than the radii of both detectors:

$$\|\vec{c}_d - \vec{c}_e\| > r_e, \quad \|\vec{c}_d - \vec{c}_e\| > r_d, \quad \forall e \in D_{\lambda_d}.$$

If overlap does occur, the fill factor of the new detector must be higher than that of any existing detector overlapping with it:

$$\frac{n_d}{r_d} > \frac{n_e}{r_e}, \quad \forall e \in D_{\lambda_d} : \|\vec{c}_d - \vec{c}_e\| \leq r_d.$$

Then, depending on the fill factor, the new detector is either discarded or *replaces* all overlapping detectors from $D_{\lambda_d}$:

$$D' = \{d\} \cup D \setminus \left\{e \in D_{\lambda_d} : \|\vec{c}_d - \vec{c}_e\| \leq r_d\right\}.$$

This is important for the stability and convergence of the algorithm.

## 6.5. Stimulus detection

The detection allows us to describe the reaction of a code space to the presentation of one or more stimuli, in terms of detector activity.

An activation $\mathbf{A} \equiv \mathbf{A}_{\lambda_a}(S)$, where $S \subset \mathcal{V}$ is a set of presented stimuli, is performed over the entire code space, so that reactions of the space to individual stimuli are combined:

$$a_{ji} = \max_{s \in S}\left(\text{sim}_{\lambda_a}(s, v_{ji})\right),$$
$$\forall j, i : a_{ji} \in \mathbf{A}, \; v_{ji} \in \mathbf{V}.$$

The subset of active points $a_{ji}$ having energy $e_{ji}$ above threshold $\mu_e$ and falling within the receptive field of detector $d$:

$$A_{\mu_e}(d, \mathbf{A}) = \left\{a_{ji} : e_{ji} \geq \mu_e, \left\|\vec{a}_{ji} - \vec{c}_d\right\| \leq r_d\right\},$$
$$\forall j, i : a_{ji} \in \mathbf{A}, e_{ji} \in \hat{\mathbf{E}}.$$

A detector activation level is defined as a ratio of the activation energy of the stimulus to the detector energy at a time of its creation:

$$E(d, \mathbf{A}) = \frac{1}{e_d} \sum a_{ji} \cdot e_{ji},$$
$$\forall j, i : a_{ji} \in A_{\mu_e}(d, \mathbf{A}), \; e_{ji} \in \hat{\mathbf{E}}.$$

The subset of active detectors $D_{\mu_d}(\mathbf{A}) \subseteq D$, with activation level above the threshold $\mu_d$:

$$D_{\mu_d}(\mathbf{A}) = \{d \in D : E(d, \mathbf{A}) \geq \mu_d\},$$

The output code is calculated by colour merging (Section 3.3) codes of all active detectors:

$$C(\mathbf{A}) = \bigcup_{d \in D_{\mu_d}(\mathbf{A})}^{\sigma} (b_d, \lambda_d).$$

Here, the operator $\cup$ denotes colour merge operation, where $\lambda_d$ is interpreted as the "colour" of the detector code $b_d$, and the resulting number of bits (saturation) should not exceed $\sigma$.

The resulting code $C(\mathbf{A})$ is a *structural embedding* describing the response of a meaningful subset of a code space to presented stimuli and mapping it to a bit vector of a given saturation.

## 6.6. Analogy with neural networks

One can notice the similarity between the described model and a neural network in which the code space $\mathbf{V}$ is the input layer and the detector space $D$ corresponds to a hidden layer of substantially smaller size connected to *some subset* of input layer neurons and to one or more neurons of the output layer .

What is important here is that the receptive subset of the detector (Figure 21) is *local*, unlike a fully connected neural network (FFN) where all neurons of layer $D$ are initially connected to all neurons of layer $V$.

The link weights will change as the network is trained, but the connectivity will remain non-local (Figure 22).

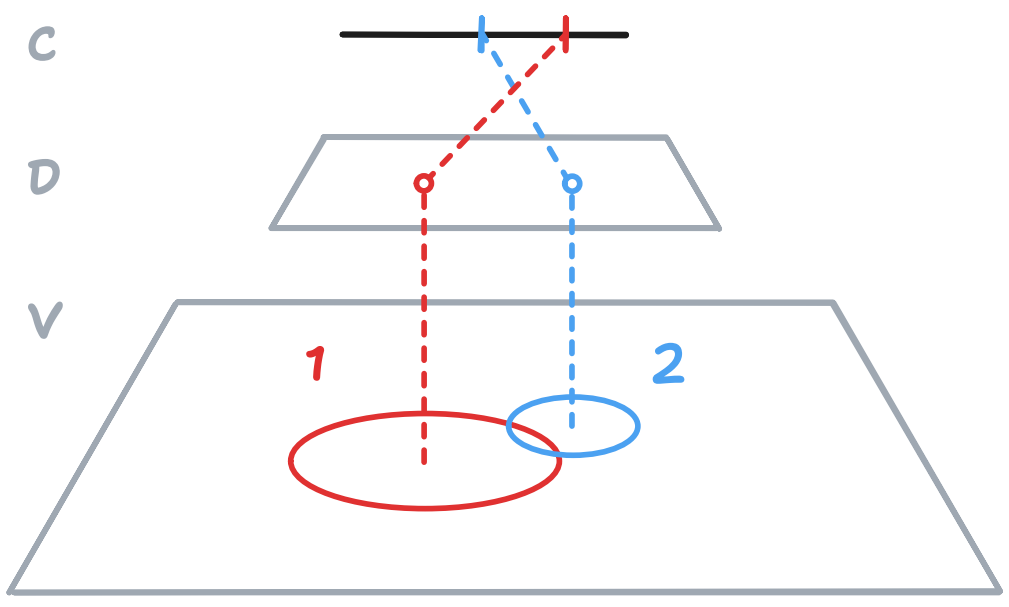

Figure 21: Local mapping of detector receptive fields to code space.

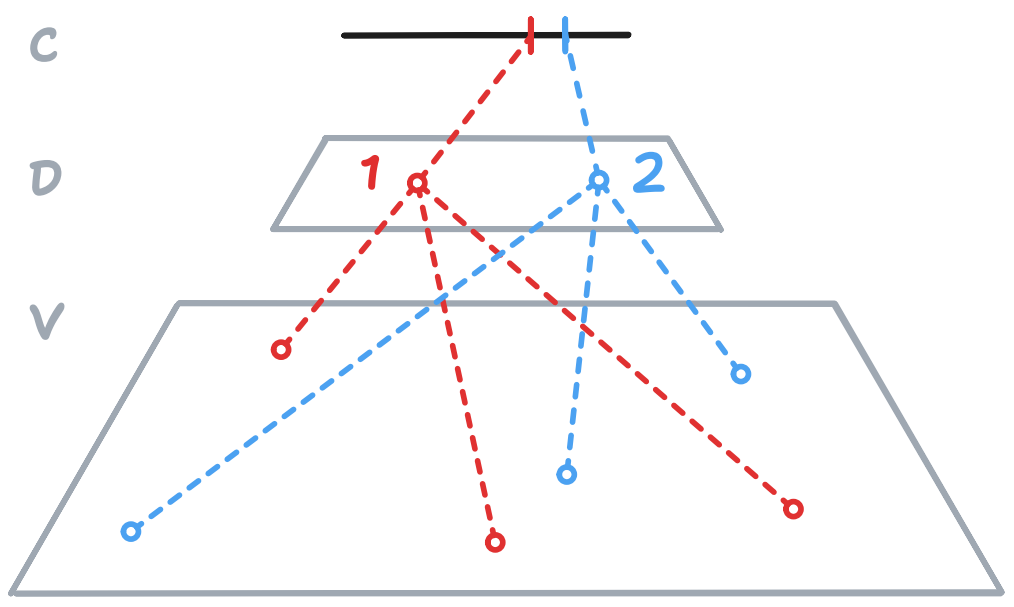

Figure 22: Non-local connections in artificial neural networks.

The fundamental difference between the two approaches is that we solve the problem in several steps:

1. First, we arrange the neurons of the input layer so that neurons encoding similar concepts are located next to each other.
2. Based on the topological features of the input layer, we *determine the number* and construct hidden layer neurons by mapping them onto a *local subset* of the input layer.
3. We calculate the preferred size and saturation of the output layer based on the average number of activated detectors when stimuli are presented.

Thus, the problem itself "tells" us what the model parameters should be for optimal description of the subject.

In a sense, we solve the problem by following the principle of "looking where the light is" or "catching a lion in a desert" by topologically turning the "cage" inside out.

By combining the resulting codes and repeating the process *layer by layer*, we can obtain complex descriptions reflecting the structure and properties of the stimulus code space.

In this sense, our model resembles the autoencoder stack [11], deep Boltzmann machines [14], Deep Belief Network [13], Layered SOM [10], and other architectures where model training occurs layer by layer.

The goals and methods of training also differ. In our case, the goal is to form a discrete code space

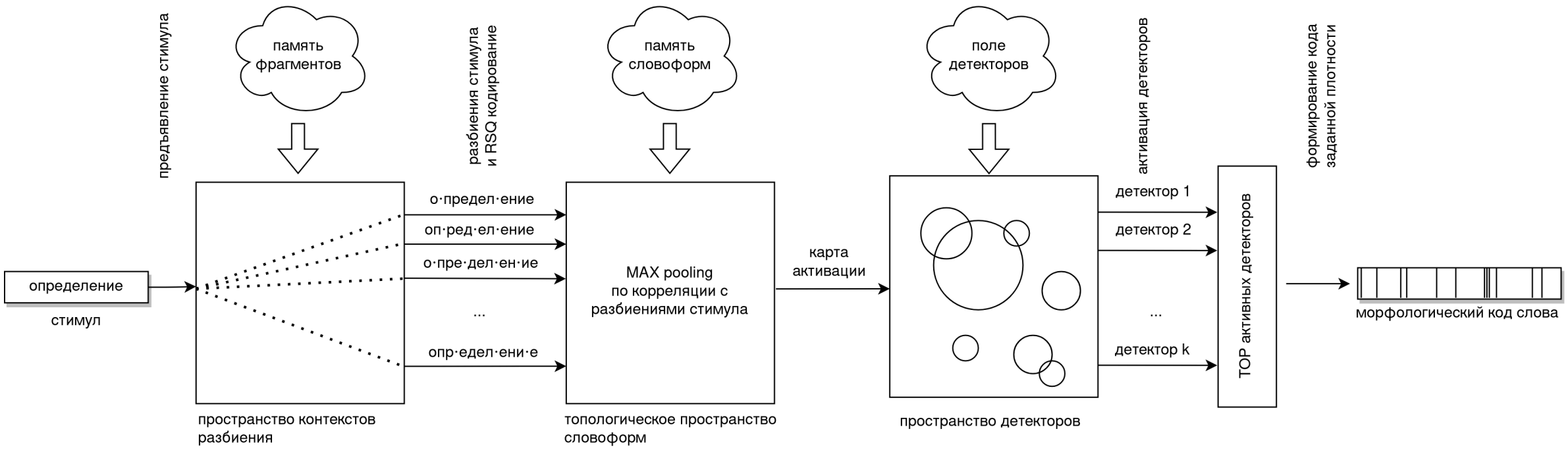

Figure 23: The model structural diagram. The processing path from *stimulus* to morphological *embedding* is shown.

by memorising the primary stimuli and solving the NP-complete problem of their layout.

# 7. Practice

This chapter discusses several practical examples to illustrate how the discrete approach can be applied to data of different modalities.

We would like to emphasise again that the resulting embeddings are *structural* and not *semantic*.

They represent the structure of concepts in a form convenient for further processing, but they do not reflect the semantics. Semantic embeddings will be considered in subsequent articles.

In all the cases described, we wanted to test the theory in practice rather than obtain a product-quality solution. Nevertheless, with proper effort, this is possible.

## 7.1. Morphology encoding

We aim to implement structural morphological embeddings so that morphologically similar words would get similar codes (Figure 23).

This will make it possible to perform operations and find relationships between individual words and whole groups of words. The particular words will be able to reinforce each other during the learning process, thus increasing learning efficiency.

Considering the "bitter lesson" [58], [59], we don't want to impose our idea of exactly how codes should be obtained on the model, but we are also careful not to waste information unnecessarily.

### 7.1.1. Motivation and specifics of the approach

Various tokenisation methods, such as BPE [60], WordPiece [61], SentencePiece [62] and others, are used for primary text encoding in neural network language models.

They form a token dictionary and then partition the input text into tokens, usually in a single way. Usually, a partitioning is chosen that optimises one of the parameters, e.g., the number of tokens or the total weight of tokenisation, calculated as the sum of probabilities (weights) of the input tokens. The approach works, of course, but it has its drawbacks.

The attention mechanism of transformers in general has complexity $O(n^2)$ depending on the number of tokens, so tokenisers are primarily tuned to minimise the total number of tokens.

This leads to morphologically close words often represented by entirely different sets of tokens. Even the same word at the beginning and middle of a sentence can be encoded by different tokens.

Such tokenisation loses the similarity relation between morphologically close words and forces the model to recover this information during training.

This requires time and resources, and, most importantly, leads to the "Swiss cheese" problem, when a seemingly adequate model can suddenly fail an obvious task simply because such relations were poorly represented in the training dataset. Therefore, models have to be trained on vast amounts of data, but even this does not guarantee the result, as the infamous example of counting

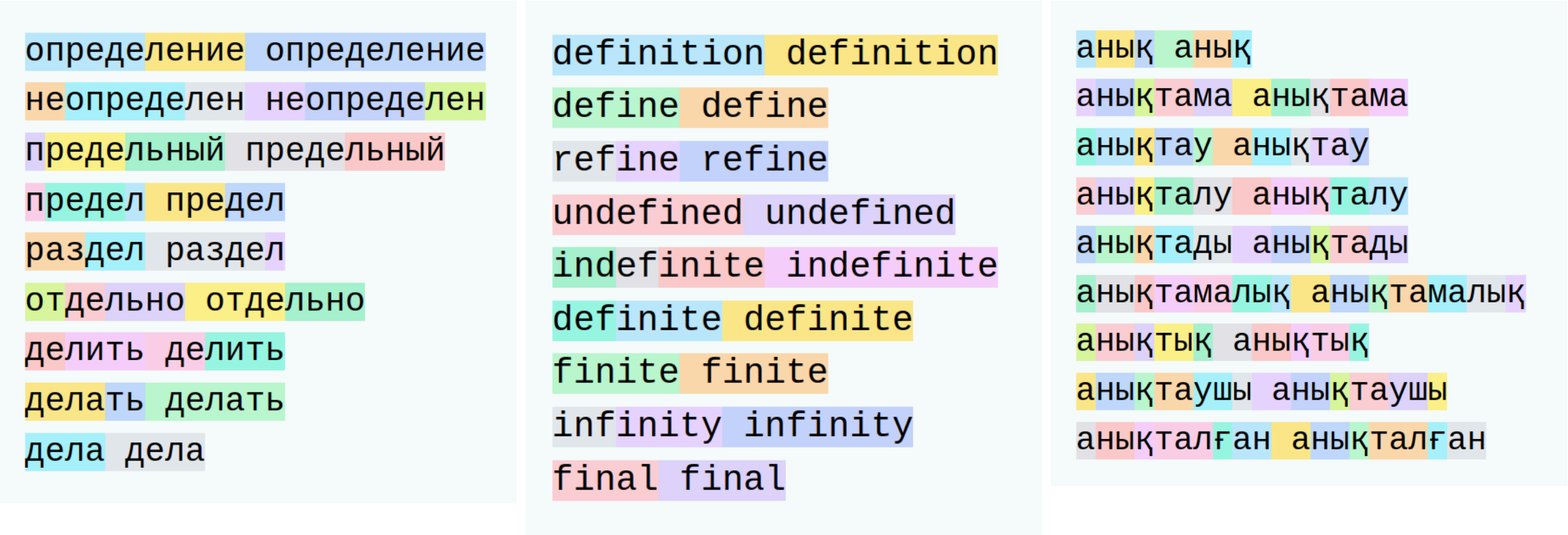

Figure 24: Example of tokenisation of DeepSeek-R1 model [63] for Russian (left), English (middle) and Kazakh (right) languages. The images were generated using Tiktokenizer [64]. It can be seen that in the general case, tokenisation of words has nothing to do with morphology. The same word can be encoded with different tokens depending on the word's position in the text.

the number of letters 'r' in the word "strawberry" [65] showed.

We build models that preserve such information in the coding system and reflect the domain topology. If any information or relationship between entities is present in the source domain, ideally, it should be preserved and reflected in the code.

In addition, we implement the principle of "multiple perspectives", which allows us to describe a certain subject without being limited by any single representation.

### 7.1.2. Existing dictionaries

There may be a natural desire to take a morphological dictionary, write out all morphemes, assign a unique code to each morpheme, and encode all words as unions of such codes.

However, the "bitter lesson" also applies here: human dictionaries are "contaminated" by human interpretations. For example, Russian words "занять", "нанять", "принять" and even "изъять" are described as consisting of a single stem, whereas native speakers see those words as constructed from several productive morphemes.[18]

In many words, fusion, assimilation, and reduction of morphemes make naive parsing challenging.

By blindly using such definitions we risk losing affinity to other structurally similar words.

### 7.1.3. Preprocessing and token dictionary

For the experiment, we implemented a simple algorithm for building a token dictionary, recursively dividing words into fragments.

1. Preprocessing
   - The text corpus is divided into words by non-letter characters (punctuation, spaces, etc.), which are removed from the stream.
   - Words are converted to lower case, augmented with start and end markers; words shorter than 2 and longer than 20 characters are removed.
2. Dictionary formation
   - Each word is split into a list of unique prefixes longer than 2 letters. The lists are merged while collecting frequency statistics.
   - Each word is split into parts: a prefix and a suffix, which are added to the dictionary as tokens. Then, for each part, the operation repeats recursively. Afterwards, the next possible split of a given word is processed.
3. Reinforcement
   - All tokens that can be used to make a whole word without gaps or overlaps are reinforced proportionally to their contribution: the ratio of the token's length to the whole word's length.

[18] This is somewhat similar to English words like "today", "understand", "overcome", "hereinafter", and so on.

This way, we get a dictionary of tokens ranked not by their frequencies, but by their ability to construct whole words.

In principle, nothing stops the use of tokenisation dictionaries from existing tokenisers (and even from a morphological dictionary), provided that they contain tokens corresponding to morphemes and allow you to get multiple options for a single word.

The subword segmentation method presented in [66], used for regularisation during the ULM model training, looks like a good candidate. In our case, we can directly use partitioning variants to populate the code space.

### 7.1.4. Token encoding

Based on a dictionary of fragments (tokens), several split variants of various appropriateness can be obtained for a given word.

Some variants conform to the division into morphemes accepted by linguists, others fuse several morphemes, remaining are incidental (Table 1).

| definition | определение | анықтама |
|---|---|---|
| definit·ion | определен·ие | анық·тама |
| defin·ition | о·предел·ени·е | ан·ық·там·а |
| de·finit·ion | о·пре·дел·ение | анық·та·ма |
| de·fin·it·ion | о·пре·деление | а·нықтама |
| d·efi·ni·ti·on | опр·ед·ел·е·ние | а·ны·қт·ама |

Table 1: Variants of word fragmentation.

Neural network tokenisers usually choose one of the options. In our models for a single word, we take *several* successful fragmentations and *use them all* to build a morphological code space.

Each word fragmentation provides similarity with the same fragments, in other words. Together, these allow for obtaining similarity without regard to tokenisation.

In Section 2.3.5, a simple variant of word encoding was considered, where each character in each position corresponds to its unique code. It can be adapted to encode tokens.

#### 7.1.4.1. Formal definition

First we define a function $f$ that maps all combinations of characters, or more specifically, Unicode code points [67] at each position to a bit vector:

$$f : \mathbb{N} \times \mathbb{U} \to \mathcal{C}, \quad \mathbb{N} = \{0, 1, ...\},$$
$$\mathbb{U} = \{\text{all Unicode code points}\}.$$

An alphabet $\mathcal{A}$ is a set of codes of all such character-positions:

$$\mathcal{A} = \{f(a) : a \in \mathbb{N} \times \mathbb{U}\}.$$

Each word fragment is mapped to a set of codes $C$ of its characters and a token weight $w$ obtained during dictionary construction. The positions of characters in *each* of the fragments are counted starting from zero.

$$\mathcal{F} = \{(C, w) : C \subset \mathcal{A}, w \in \mathbb{R}^{+}\}.$$

The fragmentation code of a word can be obtained by colour merging codes of all its fragments $F \subset \mathcal{F}$ (Formula 1):

For example, the fragmentation variant `о·предел·ени·е` will be encoded as:

$$c_w \equiv \{\text{о}_0,\ \text{п}_0|\text{р}_1|\text{е}_2|\text{д}_3|\text{е}_4|\text{л}_5,\ \text{е}_0|\text{н}_1|\text{и}_2,\ \text{е}_0\}.$$

Let $\mathrm{s}$ be a vector of characters representing one word, where each character $s_i$ is one Unicode code point:

$$\mathrm{s} = (s_1, s_2, ..., s_n), \quad \text{where } s_i \in \mathbb{U}.$$

We define a partitioning function $p$ that maps a word $\mathrm{s} \in \mathcal{S}$ to a set of its fragmentations $W \subset \mathcal{W}$:

$$p : \mathcal{S} \to W.$$

For each word $\mathrm{s}$, we can obtain a subset of its good fragmentations by selecting them with weights above a certain threshold $\lambda_p$:

$$\mathcal{W} = \left\{ (c_w, w_w) \mid \exists F \subset \mathcal{F} : c_w = \bigcup_{(\mathrm{C},\cdot) \in F} \bigcup_{c \in C} c, \quad w_w = \sum_{(\cdot, w) \in F} w \right\}.$$

Formula 1: The set $\mathcal{W}$ of *all* partitions of *all* words comprises the fragmentation codes of all words $c_w$ and is augmented by their combined weights $w_w$.

$$W_{\lambda_p}(\mathbf{s}) = \{(c, w) \in p(\mathbf{s}) : w > \lambda_p\}.$$

Later, $W_{\lambda_p}$ will become points of a code space that will be used to encode the morphology.

As a threshold for a given word $\mathbf{s}$, we used the median weight among all its fragmentations:

$$\lambda_m(\mathbf{s}) = \lambda \in \mathbb{R}^+ : \frac{|W_\lambda(\mathbf{s})|}{|p(\mathbf{s})|} \sim \frac{1}{2}.$$

We applied a different saturation limit $\sigma_i$ for each character, depending on the index of its string representation and the total word length.

This variable saturation is essential for *accenting* words (Section 7.1.4.5).

#### 7.1.4.2. Analytic codes

In the simplest case, the density for the $i$-th character *in a word* (not in a fragment) is given analytically using one of the cumulative distribution functions:

$$\sigma_i \propto \mathrm{CDF}(i) \cdot \sigma_{\max}.$$

We used the normal distribution. Practice has shown that this method works well for long words, but is inapplicable for short words.

#### 7.1.4.3. Per-character encoding

For short words of length $|\mathbf{s}| < 11$, we used table assignment (Table 2).

| $\lvert\mathbf{s}\rvert$ | $\sigma$ |
|---|---|
| 1 | 7 |
| 2 | 7, 5 |
| 3 | 7, 5, 3 |
| 4 | 8, 7, 3, 2 |
| 5 | 6, 5, 4, 3, 2 |
| 6 | 6, 5, 4, 3, 2, 1 |
| 7 | 6, 5, 4, 3, 2, 1, 1 |
| 8 | 6, 5, 4, 3, 2, 1, 1, 1 |
| 9 | 5, 5, 3, 3, 3, 1, 1, 1, 1 |
| 10+ | 5, 5, 3, 3, 3, 2, 1, 1, 1, (1,) … |

Table 2: Setting saturation according to the length of the word and the index of the *character* in the word.

| $\lvert\mathbf{s}\rvert$ | $\sigma$ |
|---|---|
| 1 | 15 |
| 2 | 15, 5 |
| 3 | 15, 7, 3 |
| 4 | 15, 7, 3, 2 |
| 5 | 15, 6, 5, 3, 1 |

Table 3: Setting saturation as a function of word length and *fragment* index.

For example, each fragmentation variant of a word of 4 characters were encoded with 20 bits, 8 bits for the first character, 7 for the second, and so on.

Longer words were encoded analytically using the cumulative normal distribution formula, but in most cases, 1 bit per character is sufficient to encode the "tail".

#### 7.1.4.4. Per-fragment encoding

The idea of this encoding variant is that instead of smoothly increasing the bit density from one end of the word to the other, we redistribute the density *per fragment*.

The code of the whole fragmentation is set as a union of fragment codes. Fragment codes are obtained by uniformly mixing character codes, considering the available bit budget for a given fragment (Table 3).

Thus, the codes of short fragments look like a union of the character codes they contain; the code of long fragments, on average, gets one bit from each letter and mostly works as a hash.

This approach better reflects the partitioning semantics and allows for *accenting* even one-character fragments.

#### 7.1.4.5. Accents

In reality, for each word fragmentation, we generate two codes: one for the word's "head" and one for its "tail", since it is important to emphasise both, the head of the word (where prefixes and roots are) and the tail (where suffixes and endings are).

Here is an example of a fragmentation for a Russian equivalent for "definition":

- **о·предел**·ени·е (accent to head)
- о·предел·**ени·е** (accent to tail)

Here, *the accent* — the part that receives more bits than the rest of the word — is in bold. In tail emphasis, the reverse index $j = |\mathbf{s}| - i - 1$ is used, i.e., $\sigma_j$ is indexed from the end.

This is necessary for the code space points to form clusters with the similarity profile we need. In turn, the codes of the detector hierarchy will express common morphological properties of words whose fragments happen to be included in same clusters (Table 4).

| Accent | |
|---|---|
| **on prefix** | **on suffix** |
| **о·предел**·ени·е | о·предел·**ени·е** |
| **предел** | дел·**ени·е** |
| **предель**·ный | на·стро·**ени·ем** |
| **преду·пре**·ди·ть | нап·ряж·**ени·я** |
| **поддел**·ка | отнош·**ению** |
| **про**·ток·ол | с·на·ряж·**енн·ый** |

Table 4: Different word fragmentations give similarity to other words with similar fragments.

Each fragmentation gives a different perspective of morphology and similarity. By combining all fragmentations, we see the picture from all sides. This helps obtaining the morphological similarity of words without regard to their tokenisation.

### 7.1.5. Code space layout

We used the Tatoeba [68] corpus as the training dataset because it is compact, contains sentences in a colloquial language, and is well-suited for our tasks.

We wanted to know if it is possible to getting reasonably high-quality embeddings on an unorganised corpus.

From the Tatoeba corpus of the Russian language, of about a million sentences, we selected those with 5 words or less, totalling 565 668 samples or about 24 MB of text. After two passes, this yielded 505 193 unique tokens (fragments).

### 7.1.6. Space filling

For all words $\mathbf{s} \in \mathcal{S}$ of the corpus, we compute possible fragmentations, both with head and tail emphases, which we injectively map onto the matrix $\mathbf{C} \in \mathcal{C}^{d \times d}$, at random coordinates so that the new points were placed at zero elements of the matrix without overwriting already added points:

$$\mathbf{C} \overset{0}{\hookleftarrow} \bigcup_{\mathbf{s} \in \mathcal{S}} W_{\lambda_p}(\mathbf{s}) \cup W^{-1}_{\lambda_p}(\mathbf{s}), \quad \lambda_p = \lambda_m(\mathbf{s}).$$

Here $W^{-1}_{\lambda_p}(\mathbf{s})$ is a set of fragmentations encoded using the function $p^{-1}(\mathbf{s})$, which emphasises the word tail by indexing characters in the reverse order.

We end up with a partially filled matrix $\mathbf{C}$ containing all codes and some remaining zeros.

To minimise the empty space, it makes sense to set the dimension of the matrix $\mathbf{C}$ so that $d = \sqrt{n + \varepsilon}$, where $n$ is the number of points in the code space, and $\varepsilon$ is the margin of empty space, about 15% of $n$. This reserve is necessary for the space to unfold smoothly, without rips and wrinkles.

### 7.1.7. Layout parameters

We used the discrete cosine measure (Section 2.2.4.1) as the similarity function for the layout.

Initially we set the similarity threshold value to $\lambda = 0.65$ and then increased it to $\lambda = 0.8$ in 0.5 increments as the layout progressed. Instead of sigmoid in this test, we used a hard similarity cutoff for all $x < \lambda$, equivalent to $\eta = \infty$.

At the start of the layout, the point spread radius was set to $r = d/2$ and then gradually decreased to 1.

As long as the code space is loosely organised, the large radius allows to select points all over the space, effectively making wide steps and distant swaps. As the space gets laid out, candidate points tend to get closer. Therefore, by reducing the radius of point selection, we increase the probability of selecting successful pairs, increasing the efficiency of the process.

In this test, parameter values were set manually, taking into account the number of swaps made per unit time. As this value fell to 1% of the initial value, the parameters $\lambda$ and $r$ were changed.

The layout parameters can be changed automatically, similar to the classical annealing method. To control the rate of change, it makes sense to take into account the specific number exchanges and the evaluation of the layout quality (Section 5.8).

### 7.1.8. Russian morphology

The layout of the Russian language morphology was done in several stages; the parameters and fragmentation methods were adjusted on the fly, and the activation profile was evaluated by test words.

Although the model was built around the Russian morphology, the same methods can be used for other languages, especially for inflective and agglutinative languages.

In the process, mistakes and poor decisions were made that impacted the topology and the quality of detection.

Nevertheless, we decided to document everything "as is", as it proved to be a good illustration of the interpretability of the model and the ability to make incremental changes without losing the progress.

The errors were corrected by hot-patching the codes as the layout progressed.

#### 7.1.8.1. Preliminary layout

Initially, we used 128-bit codes, where each character position was encoded by 40 coloured bits. The code of the whole word was obtained by colour-merging fragment codes, taking into account the absolute position of characters within the word.

At this stage, we did not use the tabular assignment of saturation of individual characters (Section 7.1.4.3). Instead, we used the cumulative distribution function (CDF) for all words, regardless of their length.

The maximum saturation of a word embedding code was set to $\sigma = 40$.

After clusters were consolidated (Figure 25), a detector hierarchy (Section 7.1.10) was constructed, and cluster activation levels were evaluated. The detection quality was estimated by analysing saturation of produced morphological embeddings and their similarity.

#### 7.1.8.2. Roots insertion

We found a number of word embeddings containing enough bits from suffixal clusters, but where roots were poorly represented. If ignored this would lead to a situation where morphological embeddings of such words would be considered close to other words with similar affixes, but words with similar roots would not be considered close.

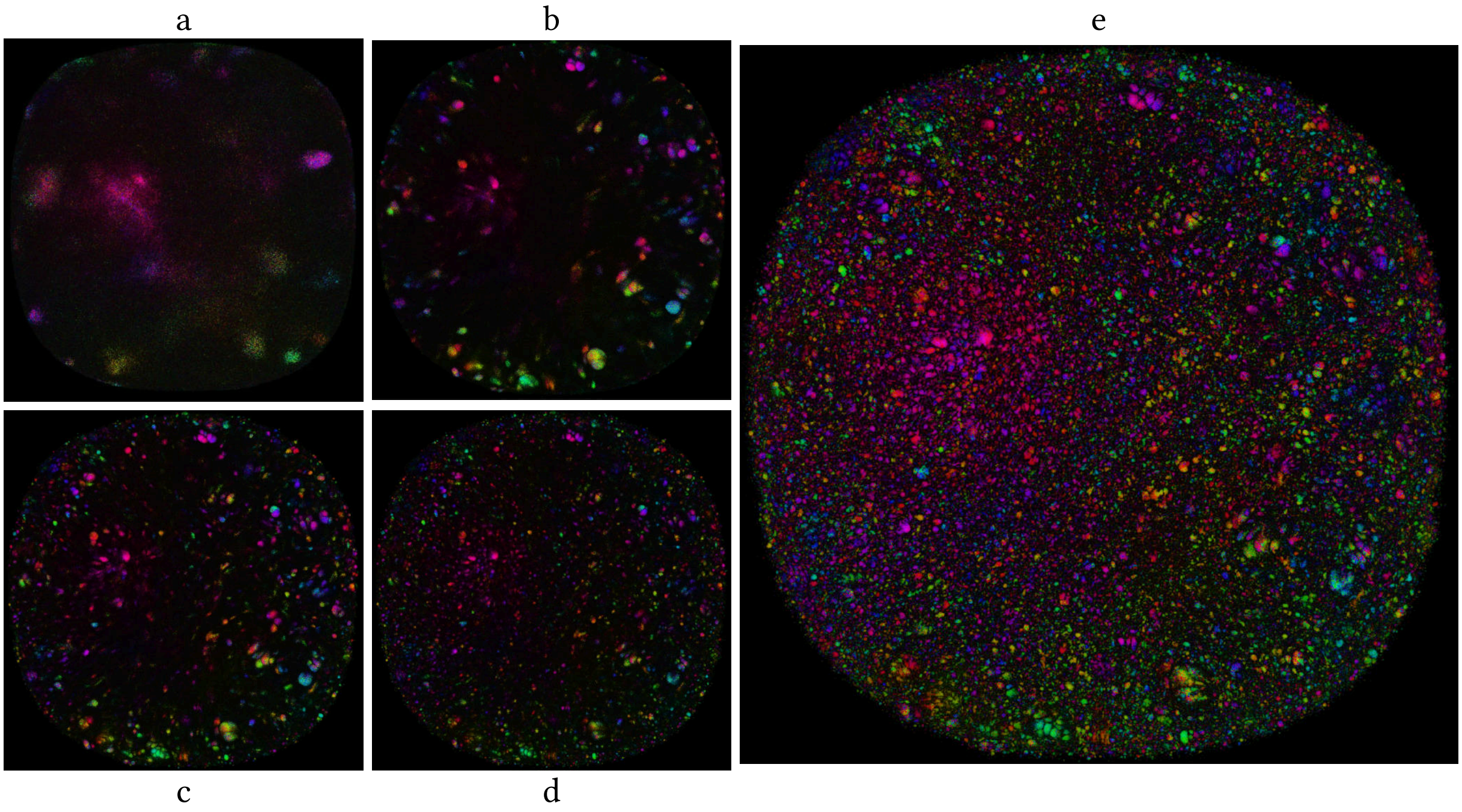

Figure 25: Visualization $\mathbf{C} \cdot \hat{\mathbf{E}}$ of the pre-layout, $n = 419\,566$, $d = 1316$.
Values of $\lambda$ in alphabetical order of images: 0.65, 0.71, 0.77, 0.81, 0.85.

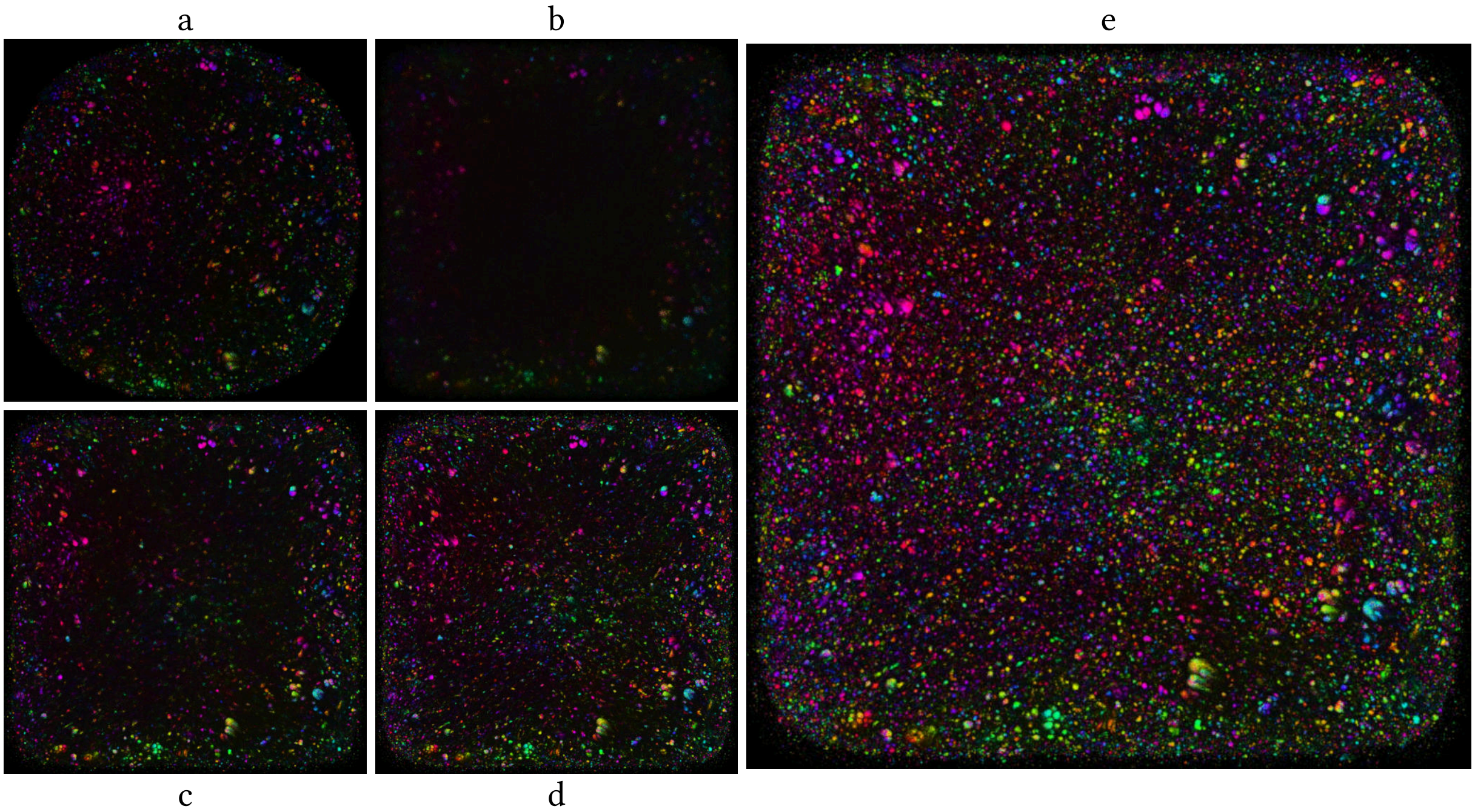

Figure 26: Layout after the root addition. $n = 2\,414\,645$, $d = 1587$.
**a**: layout start; **b**: space reconfiguration; **c, d**: processing; **e**: result.

It was decided to process weakly represented words from Tatoeba again and add root-accented fragmentations $W_{\lambda_p}(\mathbf{s})$, but without corresponding suffix-accented $W_{\lambda_p}^{-1}(\mathbf{s})$.

After the addition of root fragmentations, the morphological code space grew to $2\,414\,645$ points at $d = 1587$.

The layout (Figure 26) was then done again for added points to find their places in existing clusters.

Until the new points have gained sufficient energy, they were not visible on the visualisation (**a**). Initially, they were located in the corners of the matrix, and then they gathered in the centre, pushing the old points to the periphery (**b**).

Notably, the existing code space has preserved its topology even after adding a large number of new points (Figure 26.e).

This allows us to detect morphology even after a significant reorganisation of the code space so that old and new embeddings will be similar to a certain degree.

This can significantly reduce model training costs in real-world applications as incremental updates become possible.

#### 7.1.8.3. Defect analysis and re-layout

Subsequently, it turned out that the recently added root-codes $W_{\lambda_p}(\mathbf{s})$ erroneously include word start and end markers.

Therefore, *all* these new codes had parasitic similarity. These new codes gathered in the centre, forming a hyper cluster one-third the size of the whole space (Figure 26.d).

After fixing the bug and patching the space codes, it had to be laid out again (Figure 27).

To do this, $\lambda$ was again reduced to 0.65 to dissociate the space, and then gradually brought up to 0.9 as the layout progressed.

The space returned to its original rounded shape, and the new root codes mostly integrated into pre-existing clusters.

#### 7.1.8.4. Suffixes insertion

After recalculating the detectors and analysing the detection quality, we found out that previously added roots do indeed work. However, the space was still weakly activated in some cases,

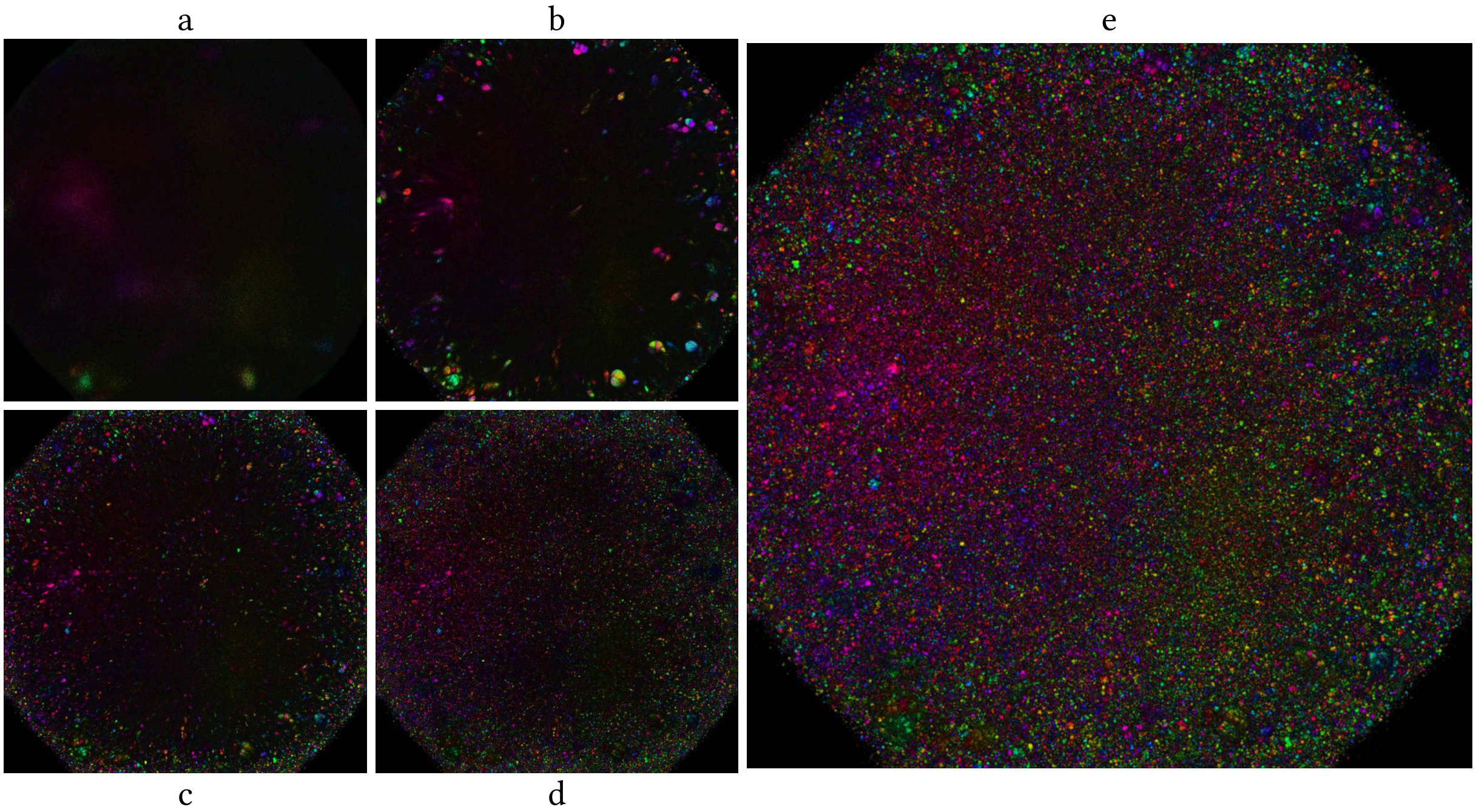

Figure 27: Layout after code correction. **a**: dissociation of previous layout; **b, c, d**: space reconfiguration; **e**: refinement by the near algorithm. Values of $\lambda$ in alphabetical order of images: $0.65, 0.75, 0.8, 0.85, 0.9$.

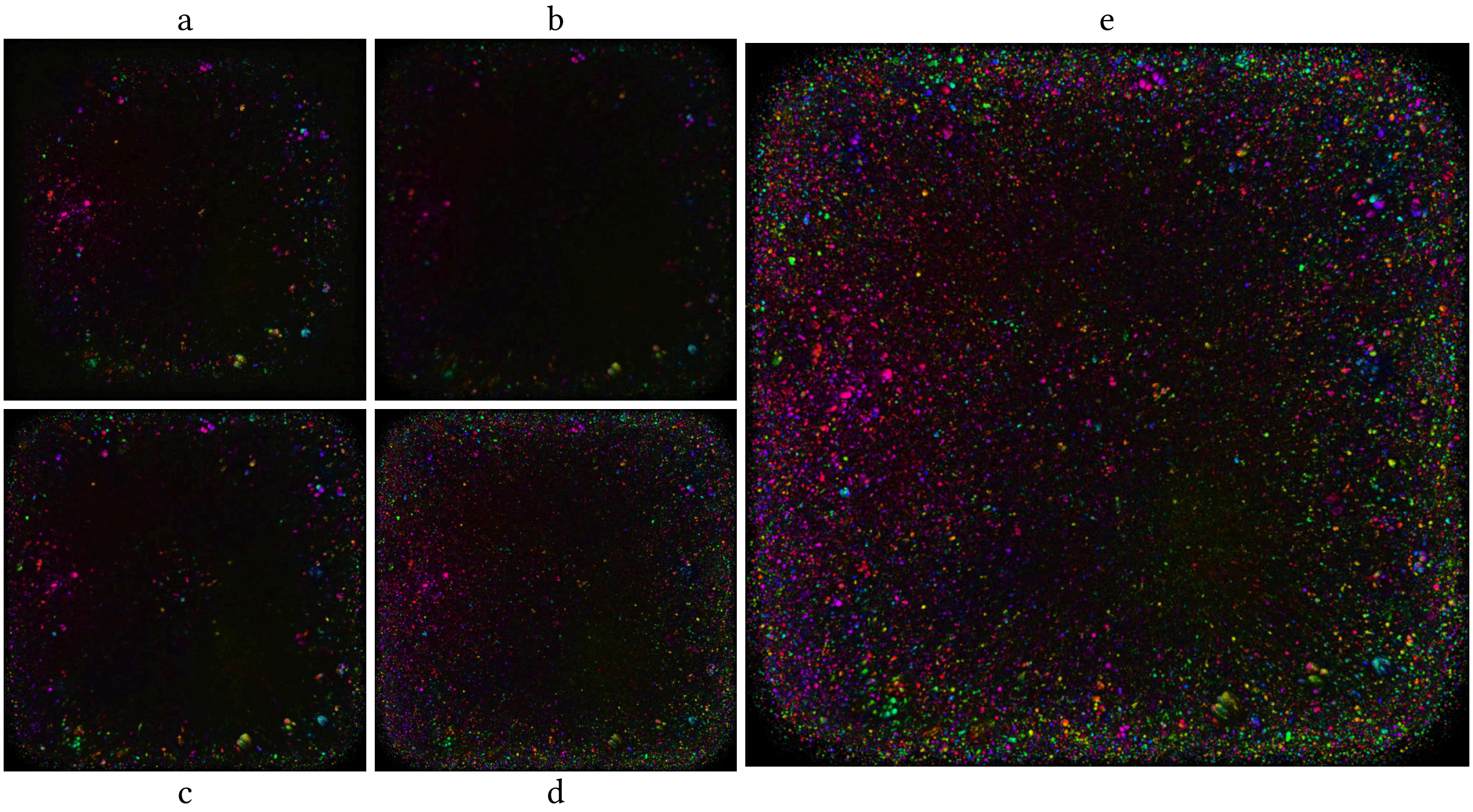

Figure 28: Layout after adding suffixes. **a**: matrix augmentation, code addition and dissociation; **b**: migration of old codes to the periphery; **c**: layout of the central region, migration of clusters; **d**: continued layout; **e**: result.

especially on short suffixes and endings $\leq 2$ characters.

Similar to the root problem, it was decided to add poorly represented suffixes from Tatoeba in hope that it would help to consolidate clusters and get better detector codes.

Another layout (Figure 28) resulted in new suffixes, but not all of them were reliably detected, even with active points being present.

For example, the short ending `·ая` did not get its cluster even though there were 8772 points encoding suffix-accented fragmentations with `·ая`

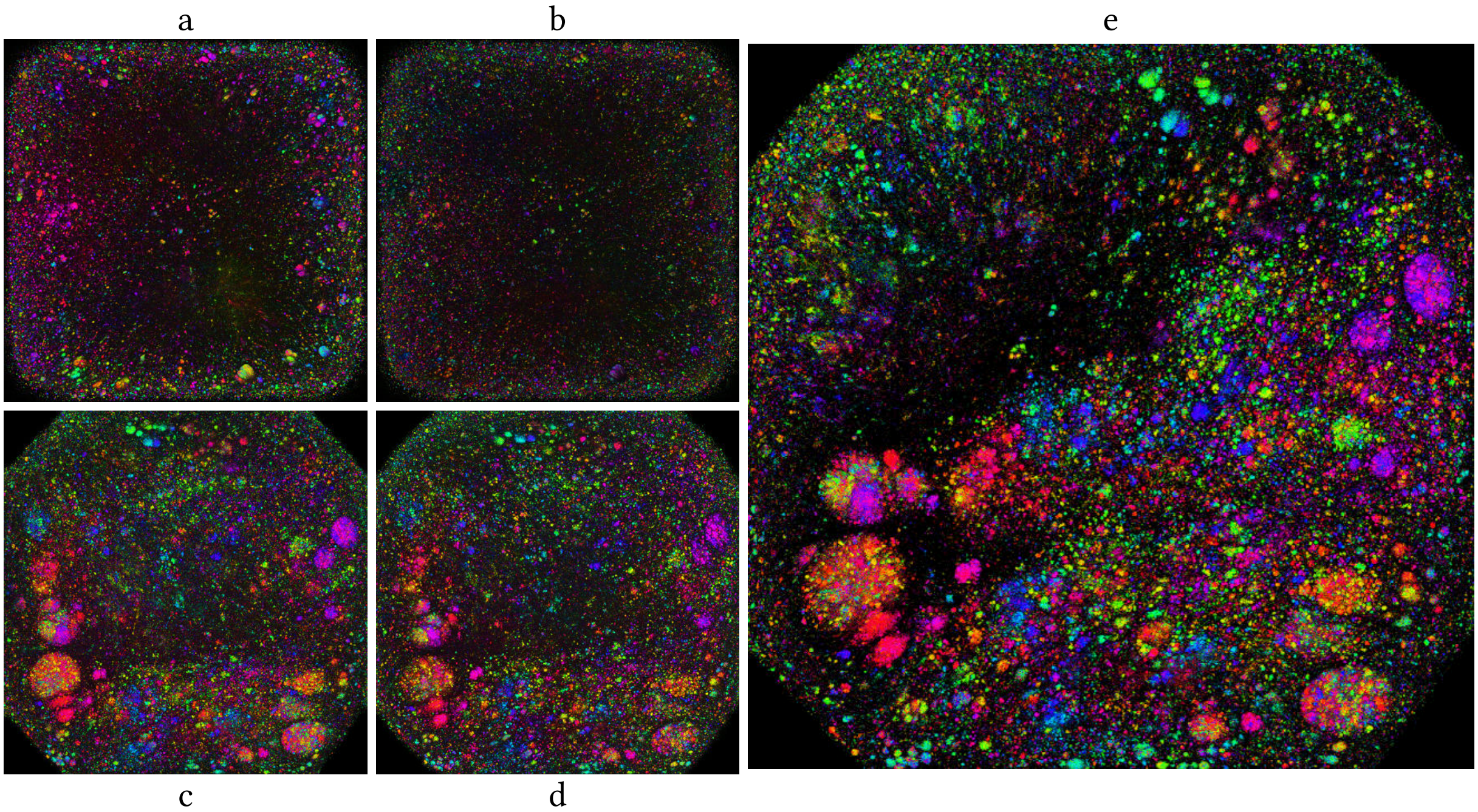

Figure 29: Code replacement and re-layout. **a, b**: space *before* and *after* in-place code replacement (without layout), $\lambda = 0.65$; **c**: new codes space after layout in the range of $\lambda = 0...0.85$, visualisation done at $\lambda = 0.65$; **d**: layout after switching to tabular codes, $\lambda = 0.65$ was set for visualisation, while the space was laid out with $\lambda = 0.85$; **e**: actual layout result, complex hierarchical cluster structure and low energy dark regions are visible, $n = 3\,037\,878$ points, $d = 1764$, visualised at $\lambda = 0.55$.

at the end. There were enough points for such a cluster, but still it did not form.

After analysing the situation, we concluded that the problem lies in the codes. Analytical assignment of saturation of individual characters led to the fact that the distribution of bit density between fragments was insufficient for cluster consolidation. The similarity of many pairs appeared to be outside $\lambda$ and did not affect the energy of the system. Therefore, there was no evolutionary pressure on these points.

In addition, significant code saturation resulted in bit collisions and parasitic similarity of the codes, degrading the signal-to-noise ratio.

#### 7.1.8.5. Per-character table codes

To solve both problems, a tabular option for specifying saturation of individual character positions (Section 7.1.4.3) was implemented.

Since we knew word fragmentations for all generated points in the code space, we recomputed all fragmentation codes using the new alphabet and patched the space in place with the new codes.

Afterwards, the space was laid out again.

After the layout, the space noticeably changed, a clearly defined hierarchy of clusters appeared, and the code "dynamic range" expanded. Before the update, $\lambda < 0.6$ caused the space to "glow", as all points were treated as similar; after the layout, even at $\lambda = 0.45$, a clear structure is visible: clusters are highlighted, unorganised "rubbish" codes are dark.

After rebuilding the detector hierarchy, embedding quality was re-evaluated. We found that the new space became good at detecting adjectives and other word forms with endings longer than one character.

However, the space still failed to consolidate clusters with single-character fragments in words like вод·а, ед·а, мам·а, etc.

Apparently, the bit density of a single-character fragment was still insufficient to affect the layout and consolidate the points.

#### 7.1.8.6. Per-fragment table codes

Once again, the primary coding system was redesigned to address the problem.

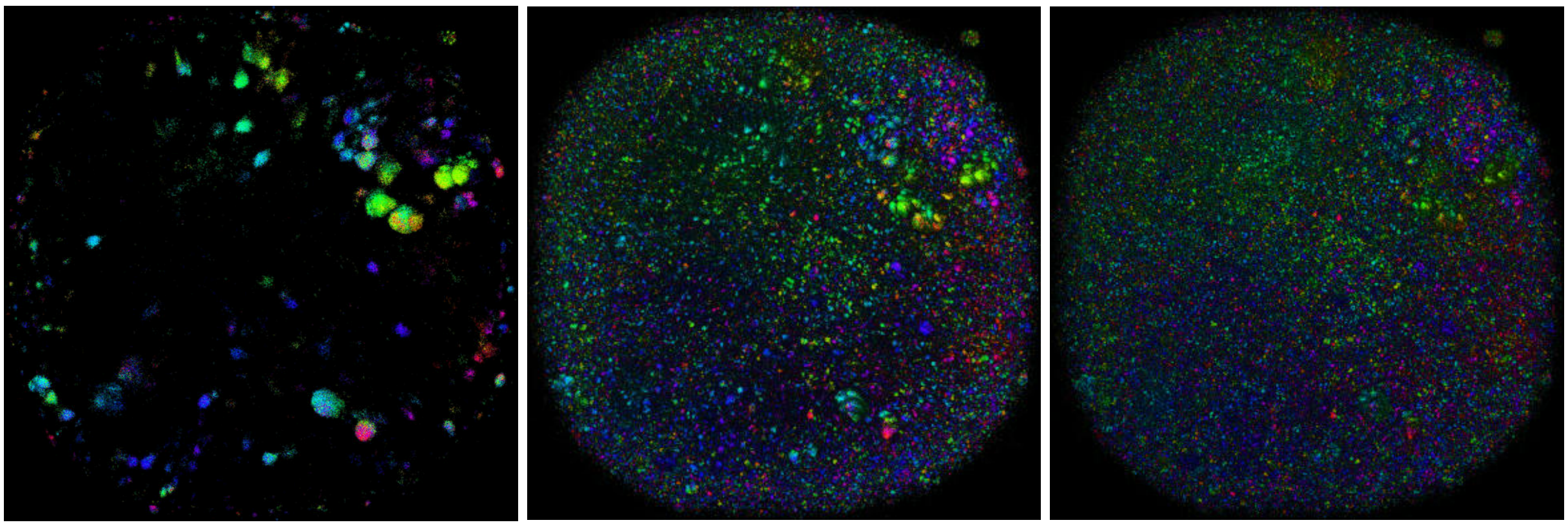

a b c

Figure 30: Visualisation $\mathbf{C} \cdot \hat{\mathbf{E}}$ of the layout of English morphology. It can be seen that the space was successfully clustered, despite the different morphological nature of the language.
**a:** beginning of layout, $\lambda = 0.8, r = 21$; **b:** process completion, $\lambda = 0.85, r = 4$; **c:** result.
The enclave in the upper-right corner is a numeric literals that happened to be in the dataset.

Instead of character-based encoding, we applied a variant in which each fragment is given a specific "budget" of bits distributed among the characters of a fragment (Section 7.1.4.4).

#### 7.1.8.7. Results

After tweaking the hyperparameters and performing another layout cycle, we obtained the actual variant of the morphological code space shown in Figure 29.e:

It is worth noting that the space is still not ideal: it contains problematic and poorly organised elements. After all the changes to the coding system, quite a lot of "rubbish" remained that was not consolidated into clusters but takes up space. Many clusters have not been consolidated completely.

Since we were not aiming for a product-quality solution, we settled on this option as it was sufficient for the proof of concept.

### 7.1.9. English morphology

English, as a language, is quite modest inflectionally, but even so, we still can identify a fair amount of productive prefixes and suffixes that prove useful in morphological analysis.

The article's length does not allow us to describe all the details of the process, so we will limit ourselves to presenting only the main steps of the morphological space layout (Figure 30).

### 7.1.10. Detector hierarchy construction

The construction of a detector hierarchy is done by selecting several activation thresholds $\lambda_a$, at which characteristic elements are extracted from the code space.

Each $\lambda_a$ value defines its layer of the detector hierarchy. The more such levels there are, the higher the saturation of the final code would be.

The minimum value of $\lambda_a$ should be noticeably above the noise level. The maximum is determined experimentally, but it usually lies between 0.8 and 0.85 and roughly corresponds to the level of $\lambda$ when the space layout was finalised.

Each detector layer should focus on characteristic structural features of the underlying code space:

- A region of a code space;
- A group of clusters (hypercluster);
- A single cluster;
- A cluster region or a pinwheel sector;
- A compact neighbourhood of a cluster, or multiple points within a pinwheel sector.

This allows us to describe the activation of a code space in terms of its structural features to obtain a structural embedding code.

Table 5 shows threshold values that were chosen for the previously laid out morphological space of the Russian language.

| $\lambda_d$ | detector colour |
|---|---|
| 0.5 | red |
| 0.6 | orange |
| 0.7 | green |
| 0.75 | teal |
| 0.8 | blue |
| 0.85 | purple |

Table 5: Detector threshold levels.

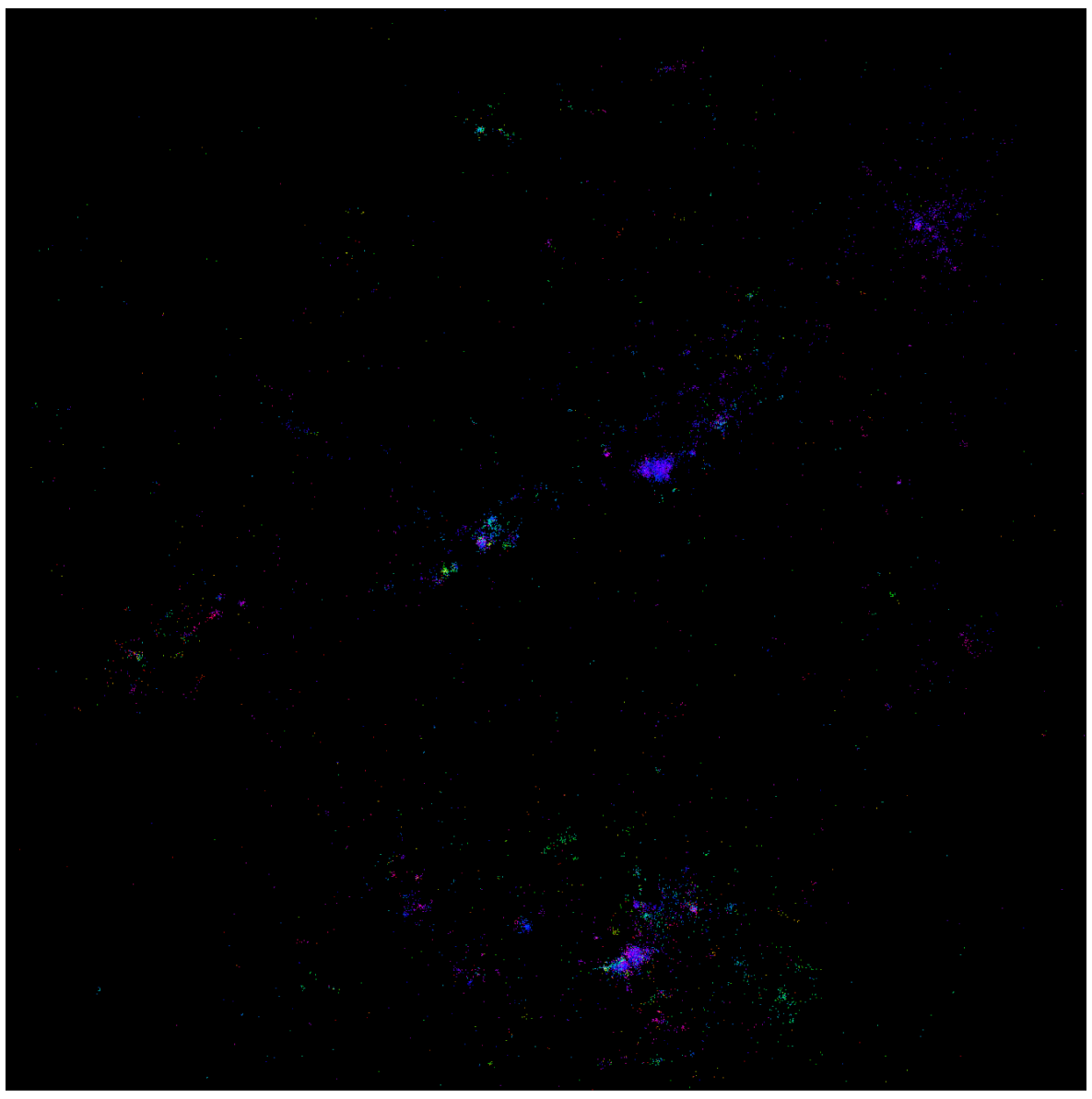

Figure 31: Visualisation $\mathbf{C} \cdot \mathbf{A}^{\mathrm{s}}$, $\mathrm{s} =$ «красивая», $\lambda_a = 0.55$.

### 7.1.11. Activation and detection

To activate a code space by a stimulus (a word) $\mathrm{s} \in \mathcal{S}$, we first obtain all successful head- and tail-fragmentations, and then compute an activation matrix (Section 6.5) and a set of active detectors:

$$\mathbf{A}^{\mathrm{s}} \equiv \mathbf{A}_{\lambda_a}\left(W_{\lambda_p}^{\pm 1}(\mathrm{s})\right),$$
$$D_\mu^{\mathrm{s}} \equiv D_\mu(\mathbf{A}^{\mathrm{s}}), \quad \lambda_p = \lambda_m(\mathrm{s}).$$

The activation matrix $\mathbf{A}^{\mathrm{s}}$ expresses the response of a code space to the stimulus $\mathrm{s}$. For example, Figure 31 visualises the activation pattern of the space when presented with the stimulus "красивая".

It can be seen that the space responded to the stimulus unevenly:

- Most of the points are silent, and activation in these points is essentially zero.
- Weak activation is observed in some regions of space, with activated points scattered randomly.
- Dense activation is observed in some areas, but the activation level is still relatively low.
- Finally, *strong activation* is observed in a few places, especially where the points are *densely concentrated.*

We are interested in the latter group where a lot of points in one or more clusters *coherently* respond to a stimulus.

This causes detectors built on such a space to activate exactly where there is a dense activation of points in the space, comparable to the activation level at the time of detector creation (Figure 32).

If we look closely at the strongly activated clusters, we would find that all of them were formed by points containing fragments also found in the stimulus.

Some of the clusters correspond to roots, some to suffixes. Together, they give an idea of a word's morphological profile, which is then encoded in the embedding.

Two words with similar roots or suffixes will activate the code space in roughly the same areas, so there is definitely be a common subset of detectors triggered by both words.

By encoding such a detector activity as codes, we will obtain discrete morphological embeddings that will also happen to be similar.

### 7.1.12. Morphological embeddings

Given a set of detectors $D^{\mathrm{s}}$, it is possible to obtain an activation code $C^{\mathrm{s}}$ by selecting detectors $d$ with activation levels higher than $\mu_c$ and combining their output bits $b_d$ with a threshold $\lambda_d$.

$$C^{\mathrm{s}} = \left\{(b_d, \lambda_d) : d \in D^{\mathrm{s}}_{\max(\mu_d, \mu_c)}\right\},$$
$$\mu_c = \mu \in \mathbb{R}^+ : |D_\mu^{\mathrm{s}}| \sim 50.$$

In the example in Figure 33, we used bit 256-bit output codes and limited the density so that, during a strong activation, at most 50 bits would be included in the code.

In addition bit vectors, detector activity levels can also be used as an embedding. Such an embedding would also have similarity properties and technically can be passed to neural networks, since its components are changing smoothly in an interval $[0, 1]$.

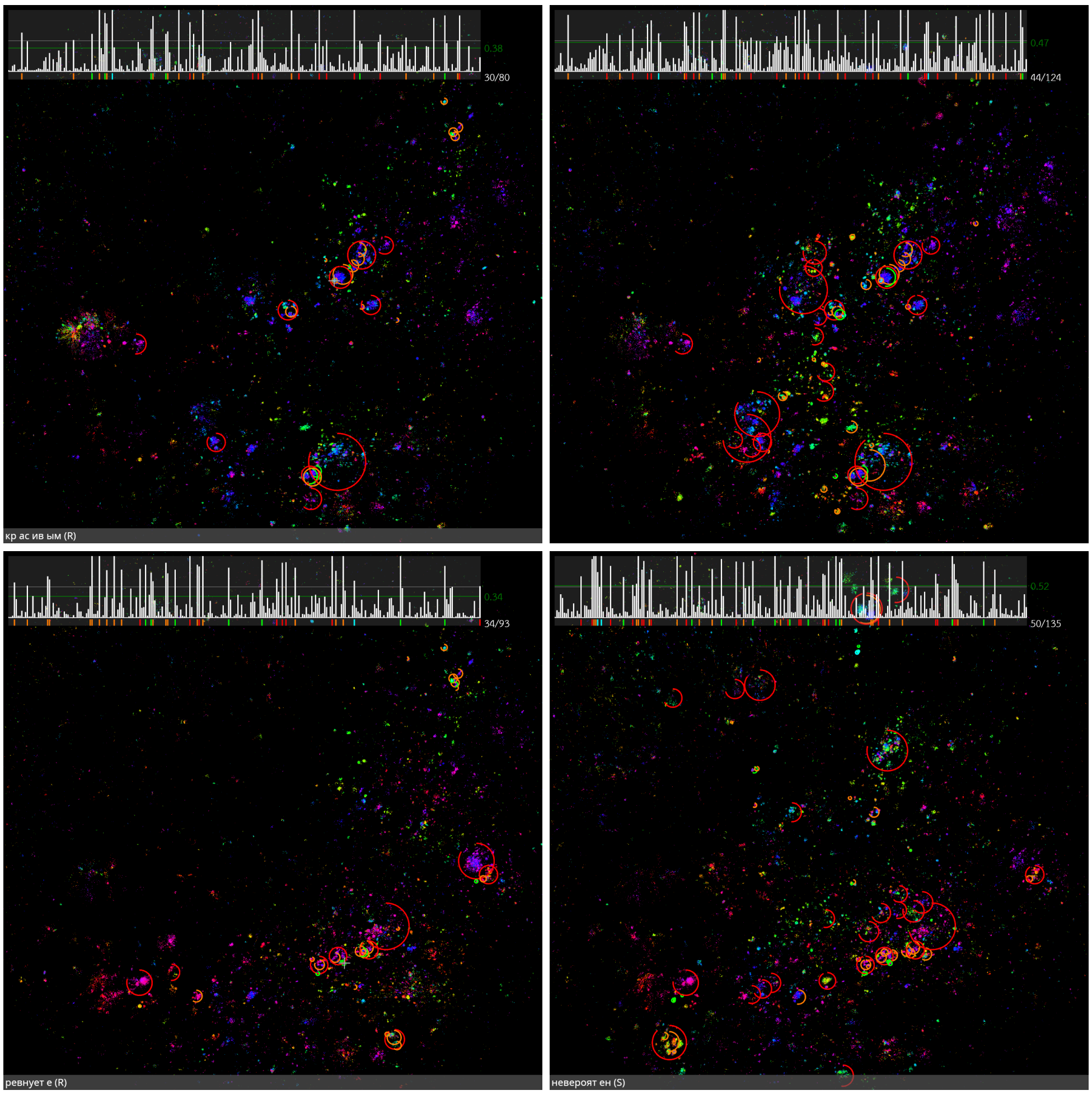

Figure 32: Visualisation of $\mathbf{C} \cdot \mathbf{A}^{\mathbf{s}}$ space activation and cluster detection $D^{\mathbf{s}}_{\mu_d}$.
Right to left and top to bottom: «красивая», «красота», «ревнивая», «ревность».

### 7.1.13. Embedding properties and analysis

As mentioned many times above, morphological embeddings allow us to obtain a code space in which different words having similar morphemes will have similar structural embeddings (Figure 34).

For example, words with similar root morphemes would have common bits in their embeddings. In theory, this property should be practically independent of word lengths or absolute morpheme positions. In a well-organised space, the words «предопре**дел**енность» and «**дел**о» would get

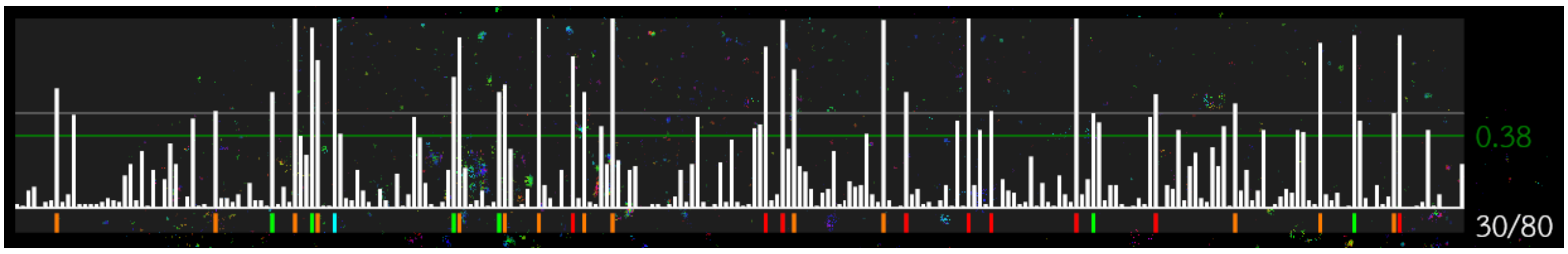

Figure 33: Diagram of detector activation and output code for the stimulus $\mathbf{s}$ = «красивая». Detectors are mapped surjectively, according to the indices of their output bits. Upper part of the image - activation levels of detectors $D^{\mathbf{s}}$, bottom - output code $C^{\mathbf{s}}$, the colours of the bars match $\lambda_d$. Total 80 detectors activated, of which the code included 30 with an activation level above $\mu_d = 0.5$ (gray line) and above $\mu_c = 0.38$ (green line).

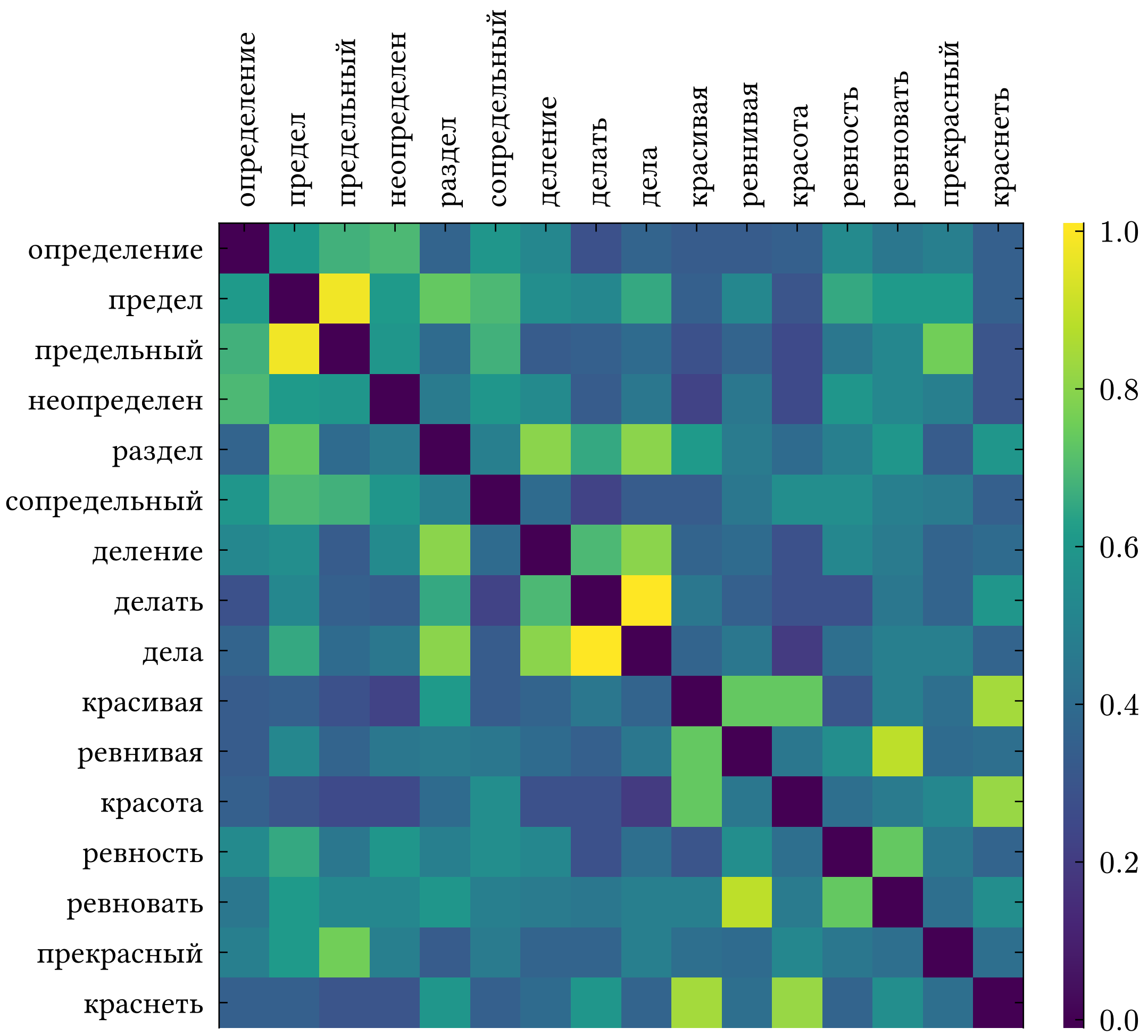

Figure 34: Heat map of morphological similarity of some Russian words. The map values are calculated by the formula $(\mathrm{sim}_{0.5}(a, b))^{1.5}$, and the discrete cosine measure (Section 2.2.4.1) was used. Elements on the diagonal are zeroed for better readability of the diagram.

common bits. The same applies to individual suffixes and endings.

In addition to the similarity property, morphological embeddings have other interesting properties.

#### 7.1.13.1. Positions of characters in a word

An individual word *fragmentation* does not contain information about positions of its fragments. Nevertheless, it turns out that embedding, as a superposition code of all word fragmentations, do contain indirect information about the absolute order of word's characters.

For example, fragmentations `о·предел·ени·е` and `е·ени·предел·о` are encoded in exactly the same way. However, the whole point is that we are not restricted to only one variant of word fragmentation. Therefore, there has to be another variant, such as `о·пре·деление`, whose fragments give a different view of the character sequence. Combining all views makes it possible to reconstruct the original word exactly.

It is reminiscent of the genome assembly task, where many individual reads provide insight into the whole genome and allow us to reconstruct the entire nucleotide sequence [69].

#### 7.1.13.2. "Cross-pollination" of word forms

Existing neural network architectures can, of course, solve the task of morphological word interpretation to a certain degree. However, they are forced to deal with inherently poorly organised data.

As it was shown in Section 7.1.1, two morphologically close words can be tokenised in very different and unpredictable ways. Modern language models have no choice but to learn word similar-ity relations individually for each for word form and each representation.

This is possible, but it requires trillions of training sample tokens, and even then, success is not guaranteed. The problems of “Swiss cheese” and “strawberry” persist even in the largest language models. “Garbage in — garbage out.”

Our morphological space benefits from the fact that it encodes information uniformly, not depending on word length and its position in the text.

Moreover, different word forms reinforce each other: some provide information about roots, other about suffixes, and together, they cover the whole variety of morphemes of a language. That is, the space derives general morphological patterns rather than representations for each particular word form.

The relationships between words learned using such embeddings would also be universal. For example, the pair “краси**вая** дев**ушка**” (an adjective in ·вая and a concordant feminine noun in ·ушка) will reinforce all such pairs, e.g.[19], “плюше**вая** игр**ушка**”. This allows model knowledge to be multiplied without presenting all possible combinations.

We are confident that a semantic model built on top of such a space will have a “linguistic sense” and will be able to “feel” the language and human speakers, without the need to flood the model with data.

#### 7.1.13.3. Encoding unknown words

A native speaker can tell quite a lot about a word’s role in a sentence solely by its form, even if the word itself is unknown to the speaker. A neural network can do this only for those words, whose tokenisation sufficiently matches other well-known words. If a word is encoded with unknown or poorly represented tokens, the neural network would not be able to infer much about the word’s meaning and its role in a sentence[20] until it has seen sufficient examples of its use.

On the contrary, a properly trained morphological space describes all morphemes occurring in a given language. Therefore, the space can interpret a word form even if it has never occurred in the training sample.

At the same time, for known words, it is possible to estimate the quality of interpretation by evaluating the activation levels of clusters and the contents of active points of the code space. It is possible to clearly distinguish a situation when a space knows a certain word form *exactly*, from a situation where the interpretation of a word was obtained in parts, but the word itself is unknown.

## 7.2. Layout of histochemical markers

This section analyses the immunohistochemical markers based on public data from histologic sections [70], [71].

**Note:** we do not claim scientific significance or reliability of the obtained results. As for the other examples, our primary goal was to test the layout and detection algorithms in practice and verify that they can work on data of different modalities.

Nevertheless, the results look interesting and potentially useful for subject matter experts.

### 7.2.1. Subject and problem statement

The tertiary lymphoid structure (TLS) is an additional lymph node, created on the periphery of inflammation focus, to better control tumours and infections.

Multispectral images of histological sections were obtained using immunohistochemistry techniques. Each colour channel corresponding to one histochemical marker highlights a particular protein.

The combinations of active proteins allow us to judge the types of cells in the slices, their states and operation modes at the time of biomaterial preparation. For example, the CD45RO protein is specific to activated T-lymphocytes, whereas Ki67 is specific to cells in the active phase of the cell cycle. Less specific proteins exist, such as CD45, which are expressed on the membranes of almost all leukocytes.

[19] English analogue: “thought**ful** deci**sion**”, “help**ful** sugges**tion**”, “beauti**ful** abstrac**tion**”, etc.

[20] Indirect information can be provided by the attention mechanism [27, chapter 3.2].

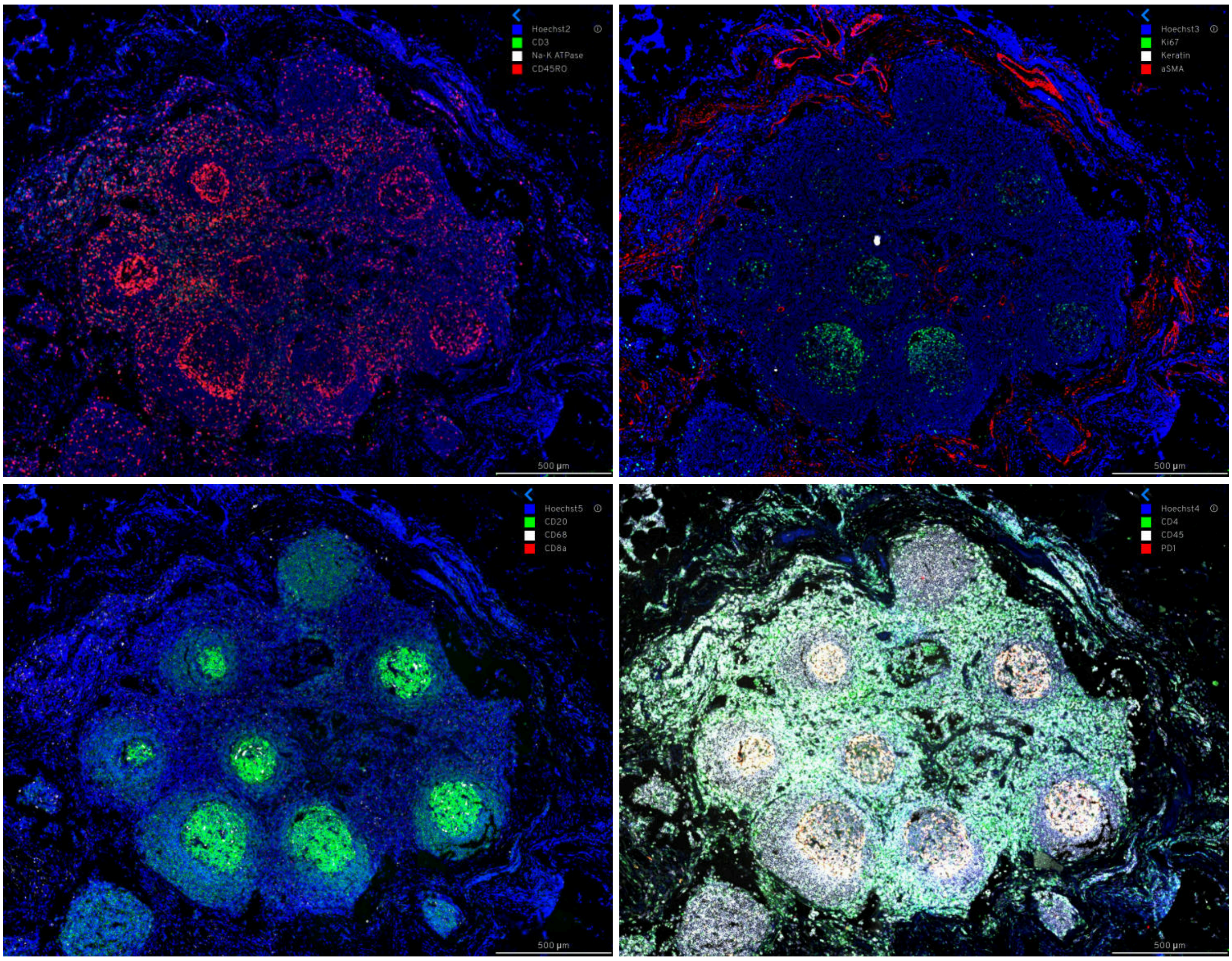

Figure 35: Multispectral image of tertiary lymphoid structure (TLS) from CRC08. From left to right, top to bottom: localisation of activated T-lymphocytes (CD45RO); cell proliferation (Ki67); activated B-lymphocytes (CD20); leukocytes (CD45).

In the TLS region various processes occur, such as T- and B-lymphocyte interaction, antigen presentation by dendritic cells, B-lymphocyte proliferation, antibody production and others.

TLS was chosen as our object of study because we had an idea of what kind of result we would get in an ideal case.

The goal is to construct a primary coding system and obtain a structural description expressing the similarity of code points in the domain.

### 7.2.2. Primary encoding

Histochemical markers well represented in the TLS domain (Table 6) were used in the analysis.

Note that the CD45 marker (not to be confused with CD45RO) was not used in the layout. Due to its low specificity, it provides almost no new information in the TLS area (all cells of interest express CD45 at approximately the same level).

In the original TIFF file, each marker is encoded with a separate 16-bit channel. Values were normalised to reflect the actual range and histogram of values. Each channel was assigned a pseudocolour for visualisation purposes.

In contrast to the complex system of primary morphology coding, everything was implemented significantly simpler here.

Each point in the code space, “a cylinder”, is a normalised $\mathbb{R}^{13}$ vector describing a single cylindrical “slice” through all “layers” of channels.

The cylinder size was determined experimentally; in this case, $r_c = 4$ was chosen. At each layer, the values of all points falling within the radius of the cylinder were averaged; therefore, each component of the 13-vector expresses the average level of marker presence in its radius.

| | | Range | | Pseudo-color | | | |
|---|---|---|---|---|---|---|---|
| **K.**[21] | **Marker** | **min** | **max** | **R** | **G** | **B** | **Immunologic role** |
| 5 | CD3 | 2160 | 10652 | 216 | 0 | 224 | a major component of the T-cell receptor |
| 7 | CD45RO | 3270 | 25688 | 0 | 255 | 238 | activated T-lymphocytes |
| 9 | Ki67 | 5164 | 24010 | 204 | 255 | 0 | marker of cell proliferation and mitosis |
| 11 | aSMA | 3667 | 42724 | 242 | 12 | 135 | smooth muscle actin; a marker of fibrotic changes and tissue remodeling |
| 13 | CD4 | 4927 | 15475 | 12 | 73 | 242 | MHC-II activation cofactor, a specific marker of T-helper cells |
| 15 | PD-1 | 2251 | 9981 | 12 | 242 | 12 | non-specific marker of T- and B-lymphocytes, immunosuppression |
| 17 | CD20 | 2292 | 33606 | 242 | 73 | 12 | A marker of activated B-lymphocytes, absent in plasma cells |
| 18 | CD68 | 710 | 12227 | 12 | 242 | 104 | Non-specific marker of macrophages, monocytes, dendrocytes |
| 19 | CD8a | 1708 | 7130 | 135 | 12 | 242 | MHC-I activation cofactor, a specific marker of T-killers |
| 22 | FOXP3 | 767 | 4333 | 216 | 23 | 230 | A specific marker of CD4+ CD45+ regulatory T lymphocytes |
| 25 | E-cad | 2064 | 9933 | 230 | 23 | 161 | cell adhesion, epithelial architecture |
| 31 | CD31 | 1592 | 11052 | 217 | 57 | 33 | Marker of vascular endothelium and angiogenesis; non-specific marker of monocytes, neutrophils, platelets and T-lymphocytes; lymphocyte integration and adhesion, signaling |
| 33 | PCNA | 8456 | 24331 | 106 | 33 | 217 | DNA polymerase cofactor; replication, DNA repair, cell cycle, proliferation |

Table 6: Histochemical markers and proteins from the TLS region.

Most of the markers described above are *non-specific*, i.e., they appear in many functionally distinct cells. Therefore, their layout leads to a complex topology.

Further variants are possible:

- If a point weakly expresses specific markers, it will adhere to one of the major nonspecific components.
- If a point is highly specific and is similar to its group, it will form an isolated cluster.
- "Rubbish" points will either end up on the periphery of clusters to which they have some affinity (due to noise or accidental match on some channel), or be pushed to the edges of the matrix.

### 7.2.3. Layout results

The layout used a simplified cosine similarity metric in the interval $\lambda$ from 0 to 0.6.

The Figure 36 shows a general view of the laid out code space.

A complex structure is immediately apparent and very different from the rounded clusters of morphology space (Section 7.1.8.7).

The current understanding is that this is due to the complex topology. Many cells express the same membrane proteins but belong to different classes.

[21]The channel number in the original TIFF file.

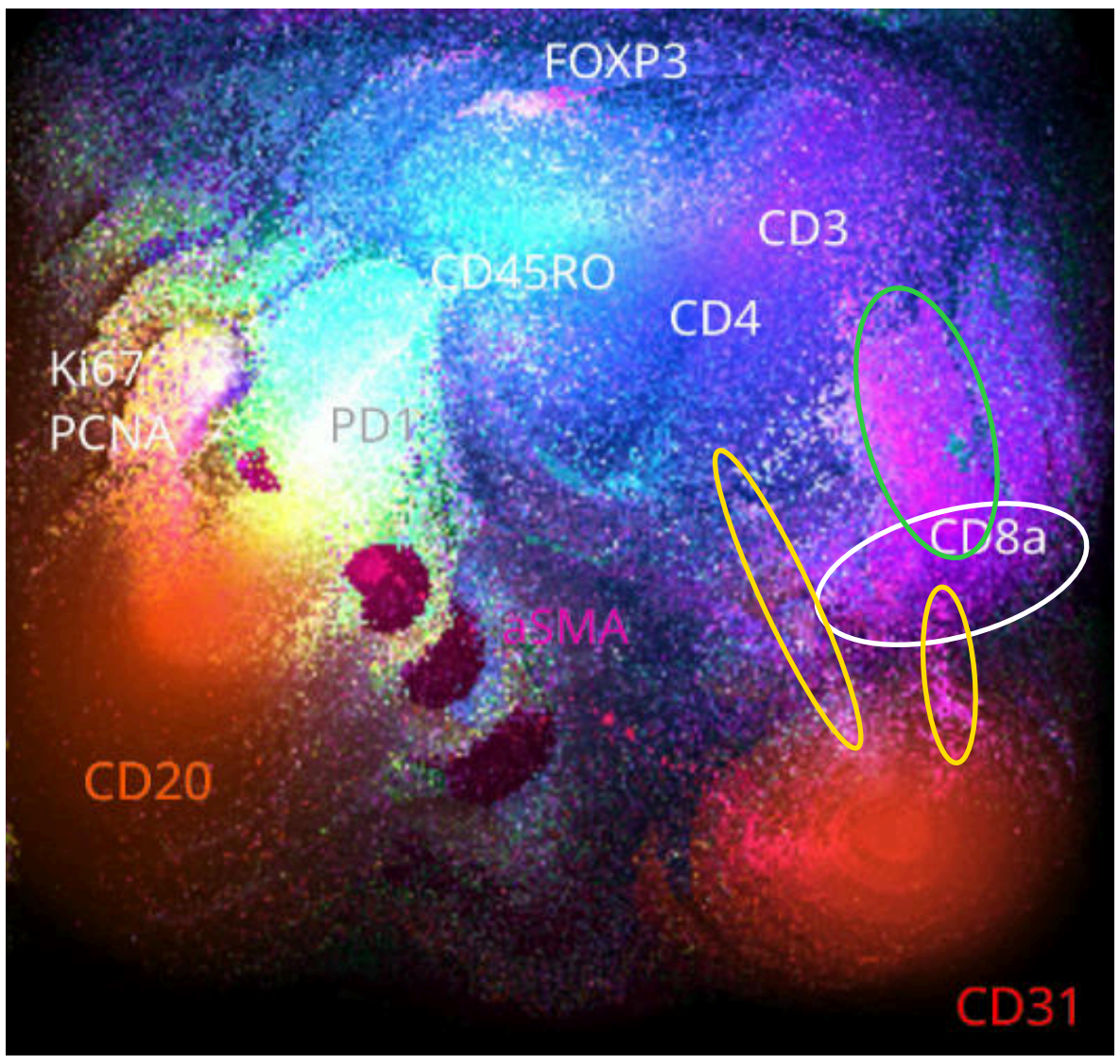

Figure 36: The laid out space. Visualisation **V**, weighted sum of channel pseudocolours, $\lambda = 0.6$.

Previous example of morphology coding had two grouping centres (by roots and by suffixes), whereas the histochemistry layout makes it appear that each nonspecific marker "pulls" space points onto itself.

#### 7.2.3.1. Highlights

- The layout algorithm was able to handle 13 dimensions.
- The rubbish points have dispersed to the corners.
- CD31 and aSMA do not interfere with the others, because they hardly interact with them (only through the edges of the cylinders).
- CD4 is accurately located between CD45RO+ and CD8+.
- CD4/CD45RO+ are accurately located next to CD20.
- There is a separate pinwheel for FOXP3.
- Ki67 and PCNA share the same cluster, which is also accurate.
- Overall, the topology of the obtained space reflects the known patterns of cell interaction.

#### 7.2.3.2. Interesting patterns

On Figure 36, colored ellipses mark somewhat interesting regions.

- White region: T-killers (CD8a) that do not express CD4.
- Green region: possibly naive CD8a.
- Yellow regions: possibly a reference to CD31 in its role in mediating cell adhesion and signalling, rather than just a marker of vascular endothelium.

Interestingly, the CD4 protein is specific to T-helper cells (MHC-II). However, the layout map shows that CD4 is also present in T-killers, but not in all of them.

It turned out that this is not a defect of the algorithm, but a pattern objectively present in the original data. CD4 and CD8a share a common place inside TLS (Figure 37.a), so they inevitably fall into the same cylinders, which, when laid out, give this effect.

The same applies to CD4 and PD-1 (Figure 37.b).

### 7.2.4. Conclusions

- Overall, the layout works, even in 13-dimensional space.
- The results are encouraging. The codes show the patterns that are represented in the data.
- Nevertheless, there are some false-positive results.

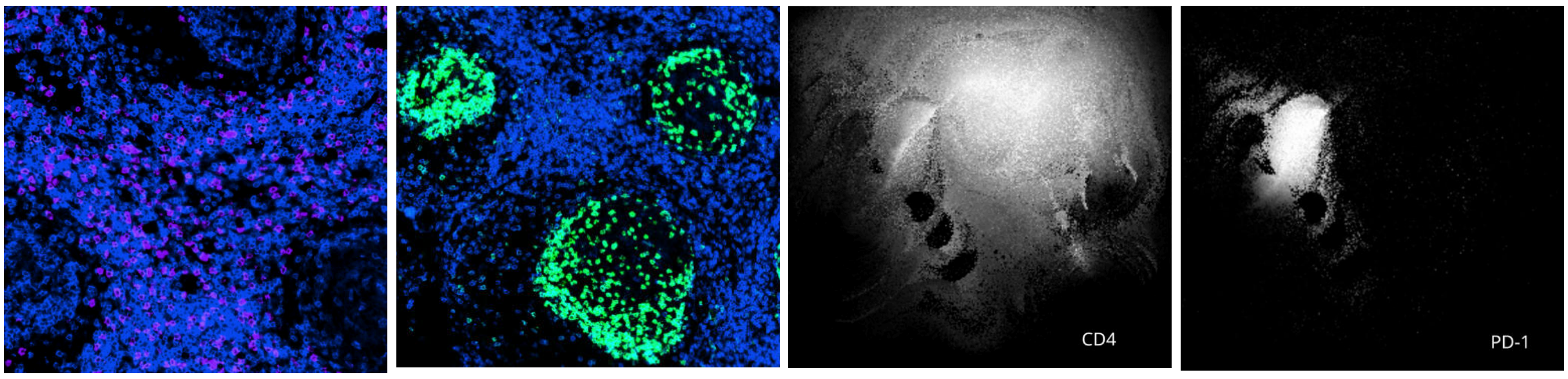

a b c d

Figure 37: **a:** Localization of CD4 (blue) and CD8a (purple) in the TLS region; **b:** CD4 (blue) and PD-1 (cyan) localization; **c, d:** Separate CD4 and PD-1 channels, respectively.

- Probably the 8×8 cylinders are too large and capture a lot of extra points.
- In the future, it makes sense to try to capture the protein composition of cells more accurately by targeting cell centroids (Hoechst).
- It is possible that parasitic connections of complicated topology produce false positive results. Without such noise, the topology of the space would be simpler and cleaner.
- A better similarity metric should give cleaner results. We computed the map on Figure 37 using a simplified cosine metric, which provides distortions. A strict cosine and Jaccard metric would probably work better.

# 8. Advantages and Specifics

In this chapter, we focus on highlighting the advantages of the discrete approach to machine learning. Some of the advantages are well-known from other methods, some are unique.

## 8.1. Taming combinatorics

Model training has always been associated with the "curse of dimensionality" and the "last percent problem".

A large number of features in a model leads to a "combinatorial explosion"; insufficient training examples lead to poor performance and the "Swiss cheese" problem.

We approach the combinatorics problem from different angles:

- Concepts are encoded as binary vectors so that close concepts correspond to close vectors at all hierarchy levels.
- The organisation of the code space allows for the description of concepts never before encountered in the training dataset.
- Facts and experience in a model are stored as independent, discrete memories with similarity properties.
- Semantic information processing uses contextual and aspectual transformations (to be described in future papers).

All this allows us to implement "cross-reinforcement" in learning, when one elements help to reinforce the others, including those encountered in third contexts.

At the same time, our models are robust to the problem of catastrophic forgetting because essentially memories do not conflict or accidentally overwritten.

We believe that our approach will significantly reduce model training costs, and improve overall model quality and performance.

## 8.2. Interpretability

When training hidden layers of neural networks, it is rarely beneficial to group neurons geometrically according to their meaning. The weights are assigned randomly, so neurons are generally connected chaotically; by looking at a single neuron, it is difficult to say which group it belongs to and which features it highlights.

In our case, clusters of points in a code space have affinity to some feature. Observing the contents of a cluster and its activation pattern makes it relatively easy to understand its essence.

The stimulus that generated each point of the code space at each level of the hierarchy is known and can be preserved. This information can be used afterwards to determine cluster dominance, assess the layout and detector hierarchy performance, and debug space activation and detection.

## 8.3. Editability

A significant disadvantage of modern neural network architectures is the limited ability to edit and retrain an already trained model.

The discrete approach solves many of the associated problems.

### 8.3.1. Separation of structure and semantics

In our models, structure is separated from semantics and data representation is separated from model memory. This allows us to replace one without affecting the other, and to adapt the trained model to changing conditions and data dynamically.

### 8.3.2. Merging spaces

The result of one level of our models is a sparse binary code, usually of small saturation and redundant length.

An interesting consequence is that different code spaces can be painlessly merged together without the need to retrain the model or change its topology.

For example, it is possible to combine Russian and English morphological *embedding spaces* together, as if they were originally been generated by a single detector hierarchy. From the model's point of view, only the number of codes changes, the size of the vectors remain the same.

It is also possible to concatenate the matrices of separate *code spaces*, preserving the detector hierarchies and the codes they generate.

If, for example, it is necessary to obtain a common code space for several languages using Latin alphabet, it can be done by combining the code spaces and re-consolidating them. However, in this case, the detector codes will change.

#### 8.3.3. Lossless training

In case of a neural network, every training step can alter all its weights with the exception of explicitly fixed layers. Thus, each new memory can potentially affect all existing memories.

This is related to the problem of catastrophic forgetting and sudden loss of model knowledge. Without the original dataset used to train the base model, it can be challenging to assess and monitor the degradation of model's skills and knowledge.

Our models are built using discrete memories that do not change and can only be deleted explicitly, if necessary. New data may change existing *data representation* (due to migration of clusters and changes in detector codes), but on its own this cannot destroy the actual code points.

Due to pinwheel migration during training, the activation codes of a code space may change. However, since activation codes also have a similarity property, so it is possible, to a certain extent, to use both, the old and the new codes without significant quality degradation and without rebuilding previous codes that were cached or written to a vector database.

#### 8.3.4. Online training

Our models can be trained continuously. In such a case, vector databases and caches can be updated at each access, replacing old codes with their refined versions (memory reconsolidation).

#### 8.3.5. Memories adjustment and alignment

Our models store memories as discrete, interpretable elements. Therefore, it is possible to make spot changes, block or delete elements without disturbing the rest of the memory.

This can correct errors and tweak potentially harmful memories or copyrighted material without retraining or rolling back the model.

Afterwards, it is always possible to check and ensure that the model is indeed free of unwanted memories.

#### 8.3.6. Topology change without retraining

Modern neural network architectures are very limited in changing the topology of an already trained model. Typically, the model must be trained again when the architecture or topology changes. In our case, this is not the case.

The model can grow incrementally as new data is added. The organisation of the code system makes this process seamless.

When a code space was already laid out, at each level, the detector hierarchy can be rearranged to balance the density and sensitivity of the output code. It is possible to select the optimal output code size and density without retraining the model.

The same is true for changes after the model was fine-tuned. After adding new points to a code space, laying it out and updating the detector hierarchy, it can happen, that the existing output code length may be insufficient to describe all active detectors. In this case, it is possible to increase the vector length so that the saturation would again be within the normal bounds.

In some instances (as Section 7.1.8 has shown for morphology), it is possible to hot-patch the codes of all points in the code space, preserving their position and significantly reducing the cost of subsequent layout.

#### 8.3.7. Cluster pruning and space optimisation

As new points get added and laid out, some clusters may become impractically large. Unfortunately, the space behaviour is hard to predict to act in advance.

Clusters often lack well-defined substructure, and most of their points share one or few common features. In such cases, it makes sense to prune the code space by removing excessive and redundant points of large clusters. The same applies to a detector hierarchy.

This will help reduce the size and speed up the activation of a code space.

As the pruning occurs, the overall code space gradually shifts from storing *facts* to storing *generalisations*, so that clusters tend to become averaged representatives of their respective classes. The space itself begins to work similar to Kohonen's maps [8].

## 8.4. Efficiency

#### 8.4.1. Caching

Our models are organised as a hierarchy of relatively independent modules and layers. In many cases, a deterministic code for stimuli can be obtained.

Therefore, each output code can be matched to its stimulus. In the simplest case, this can look like a hash table where keys are the output codes and values are the stimuli that caused them.

For example, in case of a morphology model, one can activate the space with a particular word, detect the resulting activation, and cache the embedding code in a hash table.

Since the activation map and, respectively, the embedding will always be the same for a given word form, it is enough to do it only once.

#### 8.4.2. Speculativity and parallelism

Changing a single pair of points in a code space layout has little effect on the final result. This allows us to compute the energies of each point pair speculatively assuming that only one pair is changed for each step, and that all other points remain in their places. In practice, this means that many candidate pairs can be computed from the same state of the code space.

This makes it possible to run the computation in parallel on multiple cores of a single node, multiple nodes in a cluster, or over a distributed network. Even if some of the exchanges would worsen the situation, subsequent iterations will recover, since misses do not affect the overall convergence of the layout (misses are always random and inherently unstable, unlike successful exchanges).

In the limit, cluster nodes can continue speculative computation even during state synchronisation, thus avoiding downtime and fully utilising available computational resources, thereby compensating the effect of the Amdahl's Law [72], [73].

The same principles apply to parallel and distributed computation of a detector hierarchy.

## 8.5. Reliability

#### 8.5.1. Confabulation and criticality

Current language models are based on the principle of token-by-token prediction.

This works, but by its very nature, it leads to confabulation[22], where a model makes up plausible facts rather than admitting that it doesn't know something. Since generation for a network boils down to the probability of choosing the next token, a convincing but false answer may be more likely than a denial.

Our architecture is based on a different principle and does not use predictive models. We expect our models to be resistant to confabulation, aware of unknowns, and able to clearly *distinguish fact from fiction.*

In our case, we can evaluate how well the model's response matches the memory. An adequate response should be coherent with the context and memories.

In theory, the same approach could be applied to improve the criticality of the model.

[22] False memories, erroneously referred to as hallucinations in the ML literature.

#### 8.5.2. Resistance to accidental and intentional modifications

Our models tolerate accidental or intentional modifications because there is no single point of failure: code space points are redundant, detectors rely on many points, and detector codes are also redundant.

Online learning seems especially challenging, since persistent and patient attacker can exploit the ability of a model to learn and store facts, and carefully steer the model towards malicious behaviour.

On the other hand, a direct equivalent of "one pixel" attacks [74], [75], [76] seems to be improbable.

Aside from intentional attacks, there could be situations where a model's memory or its ability to process information can degrade.

When changes to the model and additional layout are necessary, detector drift will allow the activation code to be preserved even if the clusters (pinwheels) themselves have changed their position.

Thus, it is doubtful that non-systemic failures can have any noticeable effect on model performance.

In case of significant changes in the model, a gradual degradation of the model can be observed, while only the quality of the damaged sections suffers.

## 9. Applications

Let us examine research directions that, to our opinion, fit well with the strengths of the discrete approach.

The material of this chapter is highly *hypothetical in nature*, so we ask the reader to exercise caution and be understanding.

### 9.1. Vector databases and search

In our opinion, RAG-centred models are the most promising. Their editing and online learning capabilities are ideal for creating a holistic database of memories containing all the data without artificially slicing it into fragments.

It is possible to imagine a system that would retrieve data from memory *by meanings*, considering the whole context, not just by vector similarity.

### 9.2. Adaptive codecs, stream compression

The ability of models to learn from limited data can be used to implement adaptive semantic codecs.

When processing static video and audio recordings (ahead of time), it is possible to pre-build a profile over the whole recording and encode semantic information. This way, a significant compression ratio can be achieved comparing to classical entropy codecs.

### 9.3. Integration with neural networks

Normalised activation levels of a detector hierarchy can be used as an embeddings in $\mathbb{R}^n$, instead of sparse binary codes.

We can imagine a heterogeneous architecture in which the output of a discrete model is fed as an input to a neural network. This can be helpful in seamlessly integrating our models with existing neural network architectures.

### 9.4. Discrete language models

The next logical step is to build a discrete language model that combines the versatility of neural network models with the capabilities of the discrete approach.

Modern large language models cost millions of dollars to train. Therefore, any method that can reduce costs deserves attention.

Nevertheless, we believe, it is the discrete memory-based architecture would prove to be the most effective.

### 9.5. Strong artificial intelligence

Modern neural network architectures are making remarkable progress, but are fundamentally limited in their capabilities.

Classical vector databases and RAGs are suitable for capturing and extracting *facts*, but they are completely unsuitable for consolidating new *ex-* *perience*. Therefore, neural network models are fundamentally incapable of learning to the same extent as humans.

We believe, our research has found a way to overcome this barrier and, in the long run, would allow for a model capable of continuous and unlimited experience accumulation.

Over time, this should lead to the creation of a strong form of artificial intelligence.

## 9.6. Integration with animals and humans

Our models were largely inspired by the structure of the human and animal brain.

Tonotopic maps of auditory cortex [77], maps of oculomotor dominance and orientation sensitivity [3], topographic place cells [78] and grid cells [79] in the hippocampus, all resemble the structure of a spatially organised code space. We strongly believe this is not coincidental.

Of course, it is too early to say anything for certain. More research and convincing evidence are needed.

Nevertheless, we believe that the discrete approach is applicable here too, and can, in the long term, be used to integrate with biological neural networks and interpret their activity.

# 10. Conclusion

The present work investigated a discrete approach to structural coding and processing of information.

It was shown that the primary stimuli of different modalities can be represented as discrete vectors with the similarity property.

Based on the manifold hypothesis, a method for obtaining structural codes of concepts through dimensionality reduction and clustering, was presented.

The theory was tested experimentally by constructing structural embeddings of the Russian and English morphology, and evaluating code space layout of immunohistochemical markers.

It was shown that the resulting codes do reflect the structural features of stimuli domain, inherit its topology, and can be used as embeddings.

## 10.1. Further research

### 10.1.1. Implementations refinement

The solutions presented in this paper are far from product quality.

Nevertheless, with proper attention, it is possible to move towards practical results based on the discrete semantics and, in perspective, to commercial applications of the technology.

### 10.1.2. Comparative analysis

Due to the fundamental differences between neural networks and discrete models, it is difficult to assess advantages and effectiveness of the latter.

However, it is possible to imagine a setup in which the primary encoding is performed using a discrete model, and semantic interpretation and processing are performed by neural networks. For example, we can implement analogues of Word2Vec [80] and GPT-2 [81] models, where the input layer of the network is a normalised activation vector of discrete detector hierarchy, and subsequent layers are taken from the original models.

This will test the effectiveness of the heterogeneous model from a feature engineering[23] perspective, but will not reveal anything about the capabilities of discrete semantics.

At the same time, we are sure that true potential of the discrete approach can be revealed by working with discrete semantics, not neural networks.

### 10.1.3. Semantics

The embeddings presented in this article are *structural*. They reflect the structure of the stimulus domain but do not provide a semantic interpretation.

The authors are working on methods of discrete *semantic* information processing using structural embeddings, which will be described in subsequent articles.

[23] A technique of manual feature construction. It is opposed to the approach in which the model finds the optimal representation for the problem at hand.

Eventually this would allow performing an adequate comparative analysis of the models and, hopefully, demonstrate the advantages of our approach over existing neural network architectures.

#### 10.1.4. Other modalities

The discrete approach is well-suited for implementing machine vision models.

Among other things, the authors are working on MNIST [82] and HASY [83] classification. If performance and colour representation issues would be solved, testing on the ImageNet [84] dataset would become feasible.

In addition to vision, terrain positioning and navigation problems are naturally solved.

#### 10.1.5. Discrete transformer

Reconsidering the transformer architecture in discrete terms is the most promising task.

The authors already have implementations of individual parts (attention mechanism, semantic transformations), but a full-fledged model has not yet been built.

## 10.2. Contribution of participants

**Dmitriy Kashitsyn**

Early experiments in Redozubov's team.

In this project: lead researcher; memory encoding, storage and processing systems; design and implementation of encoding algorithms, color merging, fuzzy storage and code retrieval, layout, detection, distributed learning; accelerated implementation of layout and activation using Burn, graphical interfaces for code handling and layout control; Russian language fragmentation and morphology, primary coding and layout of immunohistochemistry and human speech; research log, the article writing and its English translation.

**Dmitriy Shabanov**

Early experiments in Redozubov's team.

In this project: experimentation and implementation of the English morphology, prototype implementation of distributed layout, memory store debugging, refactoring, discussion and idea generation, technical support, English translation of the article.

## 10.3. Acknowledgements

The authors are especially grateful to Ivan Avdeev for participating in the project, help with refactoring and implementation of Vulkan shaders for accelerated code space layout and GUI rendering.

The authors sincerely thank the open source community and would like to honour the following projects in particular:

| | |
|---|---|
| Rust | https://rust-lang.org |
| Rust Analyzer | https://rust-analyzer.github.io/ |
| Rayon | https://docs.rs/rayon |
| Serde | https://serde.rs |
| Ndarray | https://docs.rs/ndarray |
| Wgpu | https://wgpu.rs/ |
| Burn | https://burn.dev/ |
| CubeCL | https://github.com/tracel-ai/cubecl |
| Iced | https://iced.rs/ |
| Typst | https://typst.app/ |
| Pandoc | https://pandoc.org/ |
| Obsidian | https://obsidian.md/ |
| Tldraw | https://tldraw.com/ |

## 10.4. Funding

The project was self-funded by the authors.

# Bibliography

[1] S. Wu, S.-i. Amari, and H. Nakahara, 'Population Coding and Decoding in a Neural Field: A Computational Study', *Neural Computation*, vol. 14, no. 5, pp. 999–1026, 2002, doi: 10.1162/089976602753633367.

[2] L. McInnes, J. Healy, and J. Melville, 'UMAP: Uniform Manifold Approximation and Projection for Dimension Reduction'. [Online]. Available: https://arxiv.org/abs/1802.03426

[3] T. Bonhoeffer and A. Grinvald, 'Iso-orientation domains in cat visual cortex are arranged in pinwheel-like patterns', *Nature*, vol. 353, no. 6343, pp. 429–431, Oct. 1991, doi: 10.1038/353429a0.

[4] S. Najafian *et al.*, 'A theory of cortical map formation in the visual brain'. [Online]. Available: https://www.nature.com/articles/s41467-022-29433-y

[5] A. P. Georgopoulos, A. B. Schwartz, and R. E. Kettner, 'Neuronal Population Coding of Movement Direction', *Science*, vol. 233, no. 4771, pp. 1416–1419, 1986, doi: 10.1126/science.3749885.

[6] A. Pouget, P. Dayan, and R. S. Zemel, 'Inference and computation with population codes', *Annual Review of Neuroscience*, vol. 26, no. Volume26, 2003, pp. 381–410, 2003, doi: https://doi.org/10.1146/annurev.neuro.26.041002.131112.

[7] C. K. A. D. S. Boerlin Martin AND Machens, 'Predictive Coding of Dynamical Variables in Balanced Spiking Networks', *PLOS Computational Biology*, vol. 9, no. 11, pp. 1–16, 2013, doi: 10.1371/journal.pcbi.1003258.

[8] T. Kohonen, 'Self-organized formation of topologically correct feature maps', *Biological Cybernetics*, vol. 43, no. 1, pp. 59–69, Jan. 1982, doi: 10.1007/BF00337288.

[9] L. van der Maaten and G. Hinton, 'Visualizing Data using t-SNE', *Journal of Machine Learning Research*, vol. 9, no. 86, pp. 2579–2605, 2008, [Online]. Available: http://jmlr.org/papers/v9/vandermaaten08a.html

[10] A. Rauber, D. Merkl, and M. Dittenbach, 'The Growing Hierarchical Self-Organizing Map: Exploratory Analysis of High-Dimensional Data', *Neural Networks, IEEE Transactions on*, vol. 13, p. 1331–, 2002, doi: 10.1109/TNN.2002.804221.

[11] G. Hinton and R. Salakhutdinov, 'Reducing the Dimensionality of Data with Neural Networks', *Science (New York, N.Y.)*, vol. 313, pp. 504–507, 2006, doi: 10.1126/science.1127647.

[12] Y. Bengio, P. Lamblin, D. Popovici, and H. Larochelle, 'Greedy Layer-Wise Training of Deep Networks', in *Advances in Neural Information Processing Systems*, B. Schölkopf, J. Platt, and T. Hoffman, Eds., MIT Press, 2006, p. . [Online]. Available: https://proceedings.neurips.cc/paper_files/paper/2006/file/5da713a690c067105aeb2fae32403405-Paper.pdf

[13] G. Hinton, S. Osindero, and Y.-W. Teh, 'A Fast Learning Algorithm for Deep Belief Nets', *Neural Computation*, vol. 18, pp. 1527–1554, 2006, doi: 10.1162/neco.2006.18.7.1527.

[14] R. Salakhutdinov and G. Hinton, 'An Efficient Learning Procedure for Deep Boltzmann Machines', *Neural Computation*, vol. 24, pp. 1967–2006, 2012, doi: 10.1162/NECO_a_00311.

[15] А. Редозубов, 'Логика мышления'. [Online]. Available: https://habr.com/ru/articles/214109/

[16] А. Редозубов, 'Логика сознания'. [Online]. Available: https://habr.com/ru/articles/308268/

[17] А. Редозубов, 'Искусственный интеллект как совокупность вопросов'. [Online]. Available: https://habr.com/ru/articles/151102/

[18] A. Redozubov, 'Pattern-wave model of brain. Mechanisms of information processing, memory organization'. [Online]. Available: https://arxiv.org/abs/1406.6901

[19] A. Redozubov, 'Holographic Memory: A Novel Model of Information Processing by Neuronal Microcircuits', in *The Physics of the Mind and Brain Disorders: Integrated Neural Circuits Supporting the Emergence of Mind*, I. Opris and M. F. Casanova, Eds., Cham: Springer International Publishing, 2017, pp. 271–295. doi: 10.1007/978-3-319-29674-6_13.

[20] А. Редозубов, 'Формализация смысла. Часть 3. Формирование контекстов', *Онтология проектирования*, vol. 11, no. 4(42), pp. 437–449, 2021, doi: 10.18287/2223-9537-2021-11-4-437-449.

[21] А. Редозубов, 'Формализация смысла. Часть 2. Пространство контекстов', *Онтология проектирования*, vol. 11, no. 3(41), pp. 309–319, 2021, doi: 10.18287/2223-9537-2021-11-3-309-319.

[22] C. Bishop, 'Pattern Recognition and Machine Learning', vol. 16, 2006, pp. 140–155. doi: 10.1117/1.2819119.

[23] R. W. Hamming, 'Error-detecting and error-correcting codes', *Bell System Technical Journal*, pp. 147–160, 1950.

[24] J. Conway and N. Sloane, *Sphere Packings, Lattices and Groups*, vol. 290. 1988, p. . doi: 10.1007/978-1-4757-2016-7.

[25] D. E. Knuth, *The Art of Computer Programming, Volume 4A: Combinatorial Algorithms, Part 1*, 1st ed. Addison-Wesley Professional, 2011.

[26] B. Bloom, 'Space/Time Trade-Offs in Hash Coding With Allowable Errors', *Commun. ACM*, vol. 13, pp. 422–426, 1970, doi: 10.1145/362686.362692.

[27] A. Vaswani *et al.*, 'Attention Is All You Need'. [Online]. Available: https://arxiv.org/abs/1706.03762

[28] J. Devlin, M.-W. Chang, K. Lee, and K. Toutanova, 'BERT: Pre-training of Deep Bidirectional Transformers for Language Understanding'. p. , 2018. doi: 10.48550/arXiv.1810.04805.

[29] J. Su, Y. Lu, S. Pan, B. Wen, and Y. Liu, 'RoFormer: Enhanced Transformer with Rotary Position Embedding'. p. , 2021. doi: 10.48550/arXiv.2104.09864.

[30] T. Cormen, C. Leiserson, R. Rivest, and C. Stein, *Introduction to Algorithms (3. ed.)*. 2009, p. .

[31] W. Pugh, 'Skip Lists: A Probabilistic Alternative to Balanced Trees', 1989, pp. 437–449. doi: 10.1145/78973.78977.

[32] S. Russell and P. Norvig, 'Artificial Intelligence : A Modern Approach / S.J. Russell, P. Norvig.', p. , 2018.

[33] R. Geisberger, P. Sanders, D. Schultes, and D. Delling, 'Contraction Hierarchies: Faster and Simpler Hierarchical Routing in Road Networks', 2008, pp. 319–333. doi: 10.1007/978-3-540-68552-4_24.

[34] Z. Wan, X. Dong, L. Wang, E. Zhu, Y. Gu, and Y. Sun, 'Parallel Contraction Hierarchies Can Be Efficient and Scalable'. p. , 2024. doi: 10.48550/arXiv.2412.18008.

[35] C. Vaaga, M. Borisovska, and G. Westbrook, 'Dual-transmitter neurons: Functional implications of co-release and co-transmission', *Current opinion in neurobiology*, pp. 25–32, 2014, doi: 10.1016/j.conb.2014.04.010.

[36] N. Chuhma, S. Mingote, H. Moore, and S. Rayport, 'Dopamine Neurons Control Striatal Cholinergic Neurons via Regionally Heterogeneous Dopamine and Glutamate Signaling', *Neuron*, vol. 81, pp. 901–912, 2014, doi: 10.1016/j.neuron.2013.12.027.

[37] R. Fettiplace and C. Hackney, 'The sensory and motor roles of auditory hair cells', *Nature reviews. Neuroscience*, vol. 7, pp. 19–29, 2006, doi: 10.1038/nrn1828.

[38] A. J. Hudspeth, 'Integrating the active process of hair cells with cochlear function', *Nature Reviews Neuroscience*, vol. 15, no. 9, pp. 600–614, Sep. 2014, doi: 10.1038/nrn3786.

[39] R. Masland, 'The Neuronal Organization of the Retina', *Neuron*, vol. 76, pp. 266–280, 2012, doi: 10.1016/j.neuron.2012.10.002.

[40] T. Baden, P. Berens, K. Franke, M. Roson, M. Bethge, and T. Euler, 'The functional diversity of retinal ganglion cells in the mouse', *Nature*, vol. 529, p. , 2016, doi: 10.1038/nature16468.

[41] D. Hubel and T. Wiesel, 'Receptive fields, binocular interaction and functional architectures in cats visual cortex', *The Journal of physiology*, vol. 160, pp. 106–154, 1962, doi: 10.1113/jphysiol.1962.sp006837.

[42] J. Cooley, P. Lewis, and P. Welch, 'The Finite Fourier Transform', *Audio and Electroacoustics, IEEE Transactions on*, vol. 17, pp. 77–85, 1969, doi: 10.1109/TAU.1969.1162036.

[43] G. Strang, 'Wavelets', *American Scientist*, vol. 82, no. 3, pp. 250–255, 1994, Accessed: Jun. 01, 2025. [Online]. Available: http://www.jstor.org/stable/29775194

[44] A. Gorban and I. Tyukin, 'Blessing of dimensionality: Mathematical foundations of the statistical physics of data', *Philosophical Transactions of The Royal Society A Mathematical Physical and Engineering Sciences*, vol. 376, p. , 2018, doi: 10.1098/rsta.2017.0237.

[45] V. Mountcastle, 'Mountcastle VBThe columnar organization of the neocortex. Brain 120(Part 4):701-722', *Brain : a journal of neurology*, pp. 701–722, 1997, doi: 10.1093/brain/120.4.701.

[46] D. Buxhoeveden and M. Casanova, 'The minicolumn hypothesis in neuroscience', *Brain : a journal of neurology*, vol. 125, pp. 935–951, 2002, doi: 10.1093/brain/awf110.

[47] J. von Neumann, *Theory of Self-Reproducing Automata.* Champain, IL: University of Illionois Press, 1966.

[48] M. Gardner, 'The fantastic combinations of John Conway's new solitaire game "life" by Martin Gardner', *Scientific American*, vol. 223, pp. 120–123, 1970.

[49] B. Ghojogh, M. Crowley, F. Karray, and A. Ghodsi, 'Principal Component Analysis', 2023, pp. 123–154. doi: 10.1007/978-3-031-10602-6_5.

[50] S. Kirkpatrick, C. Gelatt, and M. Vecchi, '(1983) S. Kirkpatrick, C. D. Gelatt, Jr., and M. P. Vecchi, "Optimization by simulated annealing," Science 220: 671-680', 1988, pp. 554–568. doi: 10.7551/mitpress/4943.003.0034.

[51] P. Laarhoven van and E. Aarts, *Simulated annealing : theory and applications.* in Mathematics and its applications. Reidel, 1987.

[52] M. Fairchild, *Color Appearance Models: Fairchild/Color Appearance Models.* 2013, p. . doi: 10.1002/9781118653128.

[53] M. Ester, H. P. Kriegel, J. Sander, and X. Xu, 'A Density-Based Algorithm for Discovering Clusters in Large Spatial Databases with Noise', in *Proc. 2 nd Int. Conf. on Knowledge Discovery and Data Mining(KDD ' 96)*, Portland, OR, 1996, pp. 226–231.

[54] E. Schubert, J. Sander, M. Ester, H. Kriegel, and X. Xu, 'DBSCAN revisited, revisited: Why and how you should (still) use DBSCAN', *ACM Transactions on Database Systems*, vol. 42, pp. 1–21, 2017, doi: 10.1145/3068335.

[55] S. Lloyd, 'Least Squares Quantization in PCM's', *IEEE Transactions on Information Theory*, vol. 28, pp. 129–136, 1982, doi: 10.1109/TIT.1982.1056489.

[56] J. B. MacQueen, 'Some Methods for Classification and Analysis of Multivariate Observations', in *Proceedings of 5th Berkeley Symposium on Mathematical Statistics and Probability*, pp. 281–297. [Online]. Available: http://home.dei.polimi.it/matteucc/Clustering/tutorial_html/kmeans.html#macqueen

[57] 'Reprint of: Mahalanobis, P.C. (1936) "On the Generalised Distance in Statistics."', *Sankhya A*, vol. 80, no. 1, pp. 1–7, Dec. 2018, doi: 10.1007/s13171-019-00164-5.

[58] R. Sutton, 'UMAP: Uniform Manifold Approximation and Projection for Dimension Reduction'. [Online]. Available: http://www.incompleteideas.net/IncIdeas/BitterLesson.html

[59] M. Yousefi and J. Collins, 'Learning the Bitter Lesson: Empirical Evidence from 20 Years of CVPR Proceedings'. p. , 2024. doi: 10.48550/arXiv.2410.09649.

[60] R. Sennrich, B. Haddow, and A. Birch, 'Neural Machine Translation of Rare Words with Subword Units', p. , 2015, doi: 10.48550/arXiv.1508.07909.

[61] J. Devlin, M.-W. Chang, K. Lee, and K. Toutanova, 'BERT: Pre-training of Deep Bidirectional Transformers for Language Understanding'. [Online]. Available: https://arxiv.org/abs/1810.04805

[62] T. Kudo and J. Richardson, 'SentencePiece: A simple and language independent subword tokenizer and detokenizer for Neural Text Processing'. p. , 2018. doi: 10.48550/arXiv.1808.06226.

[63] DeepSeek-AI *et al.*, 'DeepSeek-R1: Incentivizing Reasoning Capability in LLMs via Reinforcement Learning'. [Online]. Available: https://arxiv.org/abs/2501.12948

[64] T. D. Duong, 'Tiktokenizer: online playground for OpenAPI tokenizers'. [Online]. Available: https://tiktokenizer.vercel.app/

[65] 'Incorrect count of 'r' characters in the word "strawberry"'. [Online]. Available: https://community.openai.com/t/incorrect-count-of-r-characters-in-the-word-strawberry/829618/

[66] T. Kudo, 'Subword Regularization: Improving Neural Network Translation Models with Multiple Subword Candidates'. [Online]. Available: https://arxiv.org/abs/1804.10959

[67] U. Consortium, 'The Unicode Standard'. [Online]. Available: https://unicode.org/standard/standard.html

[68] 'Tatoeba: Collection of sentences and translations'. [Online]. Available: https://tatoeba.org/

[69] N. Nagarajan and M. Pop, 'Sequence assembly demystified', *Nature reviews. Genetics*, vol. 14, p. , 2013, doi: 10.1038/nrg3367.

[70] J.-R. Lin *et al.*, 'Multiplexed 3D atlas of state transitions and immune interaction in colorectal cancer', *Cell*, vol. 186, no. 2, pp. 363–381, Jan. 2023, doi: 10.1016/j.cell.2022.12.028.

[71] 'Multiplexed 3D atlas of state transitions and immune interactions in colorectal cancer'. [Online]. Available: https://www.tissue-atlas.org/atlas-datasets/lin-wang-coy-2021/

[72] G. M. Amdahl, 'Validity of the single processor approach to achieving large scale computing capabilities', in *Proceedings of the April 18-20, 1967, spring joint computer conference*, in AFIPS '67 (Spring). Atlantic City, New Jersey: ACM, 1967, pp. 483–485. doi: 10.1145/1465482.1465560.

[73] D. P. Rodgers, ‘Improvements in Multiprocessor System Design.’, in *ISCA*, T. F. Gannon, T. Agerwala, and C. V. Freiman, Eds., IEEE Computer Society, 1985, pp. 225–231. [Online]. Available: http://dblp.uni-trier.de/db/conf/isca/isca85.html#Rodgers85

[74] J. Su, D. V. Vargas, and K. Sakurai, ‘One Pixel Attack for Fooling Deep Neural Networks’, *IEEE Transactions on Evolutionary Computation*, vol. 23, no. 5, pp. 828–841, Oct. 2019, doi: 10.1109/tevc.2019.2890858.

[75] X. Peng, D. Zhou, G. Sun, and Y. Hu, ‘Adversarial system of one-pixel attack for hyperspectral image classification’. p. , 2023. doi: 10.21203/rs.3.rs-3221027/v1.

[76] L. Clare, A. Marques, and J. Correia, ‘A Comparative Analysis of Evolutionary Adversarial One-Pixel Attacks’, 2024, pp. 147–162. doi: 10.1007/978-3-031-56855-8_9.

[77] A. M. Leaver and J. P. Rauschecker, ‘Functional Topography of Human Auditory Cortex’, *Journal of Neuroscience*, vol. 36, no. 4, pp. 1416–1428, 2016, doi: 10.1523/JNEUROSCI.0226-15.2016.

[78] E. Moser, E. Kropff, and M.-B. Moser, ‘Place Cells, Grid Cells, and the Brain's Spatial Representation System’, *Annual review of neuroscience*, vol. 31, pp. 69–89, 2008, doi: 10.1146/annurev.neuro.31.061307.090723.

[79] M. Fyhn, S. Molden, M.-B. Moser, and E. Moser, ‘Microstructure of a spatial map in the entorhinal cortex’, *Nature*, vol. 436, pp. 801–806, 2005, doi: 10.1038/nature03721.

[80] T. Mikolov, K. Chen, G. Corrado, and J. Dean, ‘Efficient Estimation of Word Representations in Vector Space’, *Proceedings of Workshop at ICLR*, vol. 2013, p. , 2013.

[81] A. Radford, J. Wu, R. Child, D. Luan, D. Amodei, and I. Sutskever, ‘Language Models are Unsupervised Multitask Learners’, *OpenAI*, 2019, [Online]. Available: https://cdn.openai.com/better-language-models/language_models_are_unsupervised_multitask_learners.pdf

[82] Y. Lecun, L. Bottou, Y. Bengio, and P. Haffner, ‘Gradient-Based Learning Applied to Document Recognition’, *Proceedings of the IEEE*, vol. 86, pp. 2278–2324, 1998, doi: 10.1109/5.726791.

[83] M. Thoma, ‘The HASYv2 dataset’. [Online]. Available: https://arxiv.org/abs/1701.08380

[84] J. Deng, W. Dong, R. Socher, L.-J. Li, K. Li, and F.-F. Li, ‘ImageNet: a Large-Scale Hierarchical Image Database’, 2009, pp. 248–255. doi: 10.1109/CVPR.2009.5206848.